%% file: main.tex
\documentclass[lettersize,journal]{IEEEtran}
\usepackage{amsmath,amsfonts}
\usepackage{algorithmic}
\usepackage{algorithm}
\usepackage{array}
\usepackage[caption=false,font=normalsize,labelfont=sf,textfont=sf]{subfig}
\usepackage{textcomp}
\usepackage{stfloats}
\usepackage{url}
\usepackage{verbatim}
\usepackage{subcaption}
\usepackage{graphicx}
\usepackage{cite}
\usepackage{multirow}
\usepackage{ctable}

\usepackage{hyperref} 
\hypersetup{
    colorlinks=true,
    linkcolor=red,
}
\usepackage{amssymb}
\usepackage{bbding}
\usepackage{makecell}
\usepackage{booktabs}

\begin{document}

\title{Vision-Language Model for Object Detection and Segmentation:
A Review and Evaluation}

\author{Yongchao~Feng, Yajie~Liu, Shuai~Yang, Wenrui~Cai, Jinqing~Zhang, Qiqi~Zhan, Ziyue~Huang, Hongxi~Yan, Qiao~Wan, Chenguang~Liu, Junzhe~Wang, Jiahui~Lv, Ziqi~Liu, Tengyuan~Shi, Qingjie~Liu, ~\IEEEmembership{Member,~IEEE}, and Yunhong~Wang,~\IEEEmembership{Fellow,~IEEE}%
\thanks{This work was supported by the National Natural Science Foundation of China	under Grant 62176017. \textit{(Corresponding author: Qingjie Liu)}}
\thanks{Yongchao~Feng, Yajie~Liu, Shuai~Yang, Wenrui~Cai, Jinqing~Zhang, Qiqi~Zhan, Ziyue~Huang, Hongxi~Yan, Qiao~Wan, Chenguang~Liu, Junzhe~Wang, Jiahui~Lv, Ziqi~Liu, Tengyuan~Shi, Qingjie~Liu, and Yunhong~Wang are with the State Key Laboratory of Virtual Reality Technology and Systems, Beihang University, Beijing 100191, China.}
}

\markboth{Under the review of Proceedings of the IEEE}%
{Shell \MakeLowercase{\textit{et al.}}: A Sample Article Using IEEEtran.cls for IEEE Journals}


\maketitle

\begin{abstract}
Vision-Language Model (VLM) have gained widespread adoption in Open-Vocabulary (OV) object detection and segmentation tasks. Despite they have shown promise on OV-related tasks, their effectiveness in conventional vision tasks has thus far been unevaluated. In this work, we present the systematic review of VLM-based detection and segmentation, view VLM as the foundational model and conduct comprehensive evaluations across multiple downstream tasks for the first time: 1) The evaluation spans eight detection scenarios (closed-set detection, domain adaptation, crowded objects, etc.) and eight segmentation scenarios (few-shot, open-world, small object, etc.), revealing distinct performance advantages and limitations of various VLM architectures across tasks. 2) As for detection tasks, we evaluate VLMs under three finetuning granularities: \textit{zero prediction}, \textit{visual fine-tuning}, and \textit{text prompt}, and further analyze how different finetuning strategies impact performance under varied task. 3) Based on empirical findings, we provide in-depth analysis of the correlations between task characteristics, model architectures, and training methodologies, offering insights for future VLM design. 4) We believe that this work shall be valuable to the pattern recognition experts working in the fields of computer vision, multimodal learning, and vision foundation models by introducing them to the problem, and familiarizing them with the current status of the progress while providing promising directions for future research. A project associated with this review and evaluation has been created at \url{https://github.com/better-chao/perceptual_abilities_evaluation}.
\end{abstract}


\section{Introduction}

\begin{figure*}
    \centering
    \includegraphics[width=1.00\linewidth]{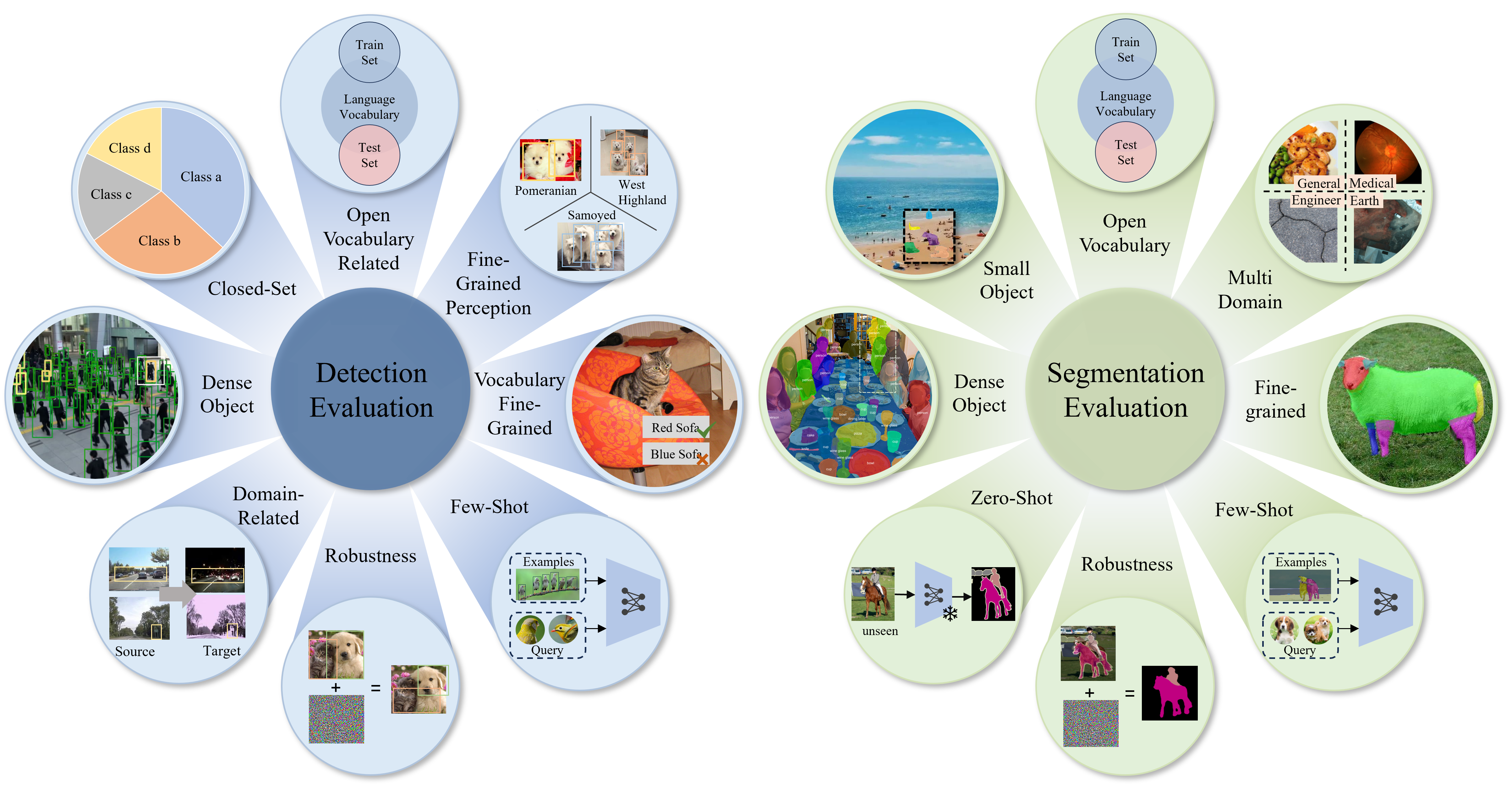}
    \caption{Illustration of Evaluation Framework for Vision-Language Models in Detection and Segmentation Tasks. For detection VLM models, we conduct comprehensive evaluations across: Traditional Closed-Set, Open Vocabulary, Fine-Grained Perception, Vocabulary Fine-Grained Perception, Few-Shot, Robustness, Domain-Related, Dense Object tasks. For segmentation VLM models, we perform systematic evaluation on Open Vocabulary, Multi Domain, Fine-Grained, Few-Shot, Robustness, Zero-Shot, Dense object, Small object tasks.}
    \label{fig:enter-label}
\end{figure*}

As artificial intelligence technology has rapidly advanced, vision-language models (VLMs) have emerged as a significant achievement in multimodal learning, becoming a focal point of research in computer vision and natural language processing. This evolution has been driven by several key factors: firstly, the iterative development of model architectures, transitioning from traditional convolutional neural networks (CNNs)~\cite{krizhevsky2012imagenet,aloysius2017review,rawat2017deep,khan2020survey} to transformer-based architectures~\cite{vaswani2017attention,khan2022transformers,han2022survey,lin2022survey} and further to large-scale pre-trained models~\cite{darcet2023vision,radford2021learningtransferablevisualmodels}, has laid a solid foundation for enhancing VLM performance. Secondly, the remarkable progress in computational power, particularly with the rapid development of GPUs and TPUs, has enabled the processing of large-scale data and complex models. Additionally, the exponential growth of data availability has facilitated VLM development, with datasets expanding from limited sizes to large-scale visual-language datasets, providing extensive image-text pairs for model training. Furthermore, the increasing demand for complex real-world tasks, especially the shift from traditional closed-set detection to open-set scenarios requiring diverse capabilities, has further propelled academic research toward multimodal models. Against this backdrop, VLMs have evolved from single-modality approaches to advanced multimodal fusion frameworks, demonstrating remarkable advantages. By aligning visual and textual features, VLMs can effectively leverage diverse data forms, enhance generalization capabilities for novel categories, and achieve outstanding performance in object detection and segmantation tasks.

Vision serves as the core perceptual channel for interpreting environmental information, which necessitates systematic evaluation of VLM's efficacy in enhancing conventional vision tasks through multimodal understanding. Object detection~\cite{ren2016faster} and segmentation~\cite{ADE20K} constitute fundamental tasks in computer vision, serving as essential components for perception and scene understanding. These technologies form the backbone of various practical applications across multiple domains, including autonomous driving~\cite{cordts2016cityscapes}, medical imaging~\cite{blumenstiel2023mess}\cite{zhang2024segment}\cite{ma2024segment}, and intelligent robotics~\cite{blumenstiel2023mess} and so on. 

Current VLMs fundamentally operate by aligning visual and textual features to achieve their broad and robust capabilities. In object detection tasks, VLM-based detection aligns visual features with text descriptions through contrastive learning approaches, as exemplified by GLIP~\cite{li2022grounded} and GroundingDINO~\cite{liu2023grounding}, achieving generalization across unseen categories through pre-training on large-scale datasets such as CC12M (Conceptual 12M \cite{changpinyo2021conceptual}), YFCC1M (a subset of YFCC100M \cite{thomee2016yfcc100m}).
In the context of segmentation tasks, recent works have focused on transferring global multi-modal alignment capabilities of VLMs to fine-grained alignment tasks, specifically region-text~\cite{cha2023learning} and pixel-text alignment~\cite{cho2024cat}. These advancements leverage diverse supervision strategies to facilitate dense prediction in pixel-wise segmentation tasks.
At their core, these models extend concepts from pre-training approaches such as CLIP~\cite{radford2021learningtransferablevisualmodels}; however, while CLIP functions as a classification model, the alignment mechanisms and principles differ across VLMs. For instance, some models leverage contrastive learning for feature alignment, while others employ cross-attention for feature fusion. Notably, current VLMs predominantly demonstrate strong performance on open-vocabulary (OV) tasks, but their ability to generalize to other specific tasks remains an area requiring further exploration.

Due to the potential and powerful capabilities of VLMs, many works have been exploring how to apply VLMs to downstream tasks, including object detection, semantic segmentation, and more. For example, DA-Pro~\cite{li2023learning} builds upon RegionCLIP~\cite{zhong2022regionclip} by dynamically generating domain-specific detection heads through domain-relevant and domain-agnostic prompt prefixes for each target category, thereby significantly improving cross-domain detection performance. COUNTGD~\cite{amini2024countgd} improves instance counting by augmenting the text prompts in GroundingDINO~\cite{liu2023grounding} with visual exemplars of corresponding categories, forming enhanced textual descriptions for detecting target objects in input images, achieving the first open-world counting model. However, existing research and related reviews have primarily focused on detection and segmentation tasks in open-vocabulary settings, often overlooking the complexities and challenges of real-world scenarios. As a result, comprehensive evaluations across a wide range of visual downstream tasks have not been conducted. 
As shown in Fig.~\ref{fig:enter-label}, to thoroughly assess the performance of VLM models in different scenarios, we have designed 8 different setting for detection tasks, covering traditional closed-set detection tasks, open-vocabulary-related tasks, as well as domain adaptation scenarios and dense-object scenarios that are more realistic. For segmentation tasks, we have set up 8 different settings, including zero-shot evaluation, open-world semantic segmentation tasks, as well as small-object and dense segmentation tasks.

\begin{figure}
	\centering
	\includegraphics[width=\columnwidth]{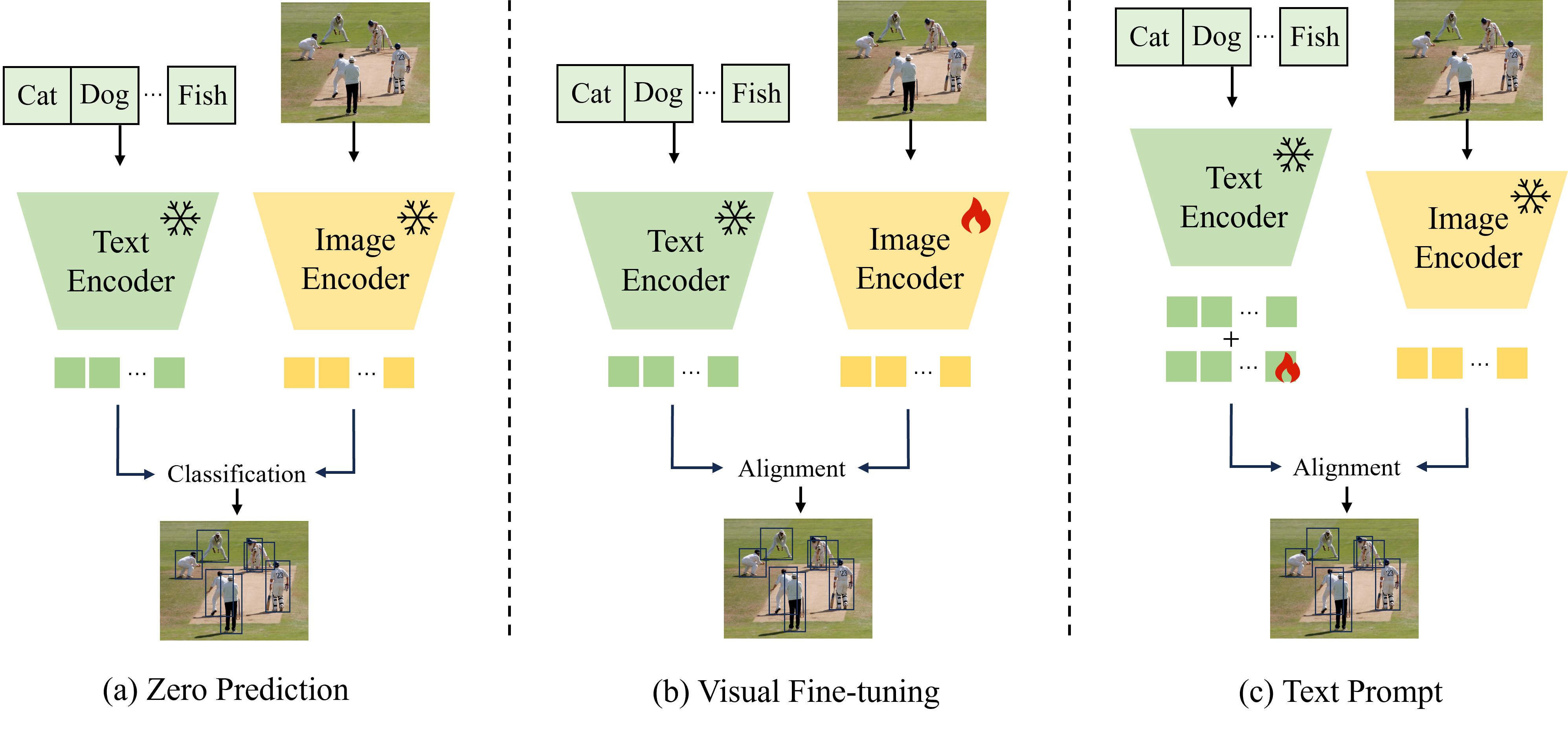} 
	\caption{Illustration of three granular fine-tuning strategies for visual-language detection models. (a) \textit{Zero Prediction} directly evaluates the VLM on downstream tasks without fine-tuning. (b) \textit{Visual Fine-tuning} adapts the VLM's visual branch on downstream data before evaluation, and (c) \textit{Text Prompt} optimizes only the text prompts with downstream data prior to evaluation.}
	\label{fig:finetune_method}
\end{figure}

In the context of VLM-based detection tasks, as shown in Fig.~\ref{fig:finetune_method}, three granularity levels of fine-tuning are employed to assess model performance: \textit{\textbf{Zero Prediction}}, \textit{\textbf{Visual Fine-tuning}}, and \textit{\textbf{Text Prompt}}. These three ways differ in their trade-offs between computational cost and performance, making them suitable for various downstream tasks.

\textit{Zero Prediction}: This approach involves directly applying the pre-trained VLM model to downstream datasets without any fine-tuning. It leverages the model's inherent generalization capabilities and is particularly suitable for scenarios requiring rapid deployment. Formally, for a pre-trained model $f_{\theta}(x, t)$, where $x$ represents the image and $t$ represents the text prompt, \textit{Zero Prediction} directly applies $f_{\theta}(x, t)$ to downstream datasets.

\textit{Visual Fine-tuning}: This approach involves fine-tuning the visual branch of the VLM on downstream visual tasks while keeping the text branch fixed. By adapting the model to the distribution of downstream data, it enables rapid alignment of the VLM to specific tasks. However, this method incurs a relatively high fine-tuning cost. Formally, if the model consists of a visual encoder $E_v$ and a text encoder $E_t$, \textit{Visual Fine-tuning} modifies $E_v$ while keeping $E_t$ fixed.

\textit{Text Prompt}: This approach focuses on fine-tuning only the text prompts, adapting them to downstream tasks through minimal adjustments. Specifically, it introduces learnable parameters to the text encoding process, enabling task-specific adjustments with low computational overhead. In some cases, this method can even surpass the performance of \textit{Visual Fine-tuning} on specific downstream tasks. Formally, for a text prompt $t = [t_1, t_2, \dots, t_n]$, \textit{Text Prompt} introduces learnable parameters $\Delta_t$, resulting in an adapted prompt $t' = t + \Delta_t$.


In contrast to conventional semantic segmentation models that are confined to a fixed set of predefined categories~\cite{chen2017deeplab}, VLM-based segmentation approaches~\cite{cho2024cat} offer the potential for open-vocabulary segmentation of arbitrary categories. However, the fundamental question remains: do current models truly achieve the promise of segmenting anything? In this work, we conduct a comprehensive evaluation of their capabilities across multiple domains using diverse benchmark datasets. Through extensive empirical studies and in-depth analysis, we systematically investigate the strengths and limitations of state-of-the-art VLM-based segmentation models~\cite{cho2024cat, jiao2025collaborative_maft+, xie2024sed}. Our findings provide valuable insights and establish concrete research directions for advancing the development of more robust and versatile VLM-based segmentation models.

In this study, we present a comprehensive survey of vision-language models (VLMs) in dense prediction visual tasks and summarize our three main contributions as follows:
\begin{itemize}
 \item Pioneering Evaluation: This paper is the first to treat VLMs as "foundation models" and conduct extensive evaluations across a wide range of downstream visual tasks. Through this unique perspective, we systematically demonstrate the performance of VLMs across different visual tasks, providing valuable benchmarks for understanding their potential and limitations.
 \item Granular Analysis of Fine-tuning Strategies: We systematically investigate the impact of three fine-tuning approaches—\textit{zero prediction}, \textit{visual fine-tuning}, and \textit{text prompt}—on downstream tasks, with a particular focus on segmentation tasks. This in-depth analysis reveals the strengths and weaknesses of various fine-tuning strategies in practical applications, offering critical insights for model optimization.
 \item In-depth Mechanism Analysis: From the perspectives of training methodologies and model architectures, we explore how these factors influence model performance on downstream tasks. This research goes beyond surface-level applications and delves into the intrinsic mechanisms of VLMs, providing support for future model design and improvement.
\end{itemize}
In summary, our study not only provides comprehensive evaluation and in-depth analyses of VLMs but also lays a solid foundation for advancing the field, promoting further breakthroughs and progress in object detection and segmentation tasks.
The remaining sections of the paper are organized as follows: Sec.~\ref{sec:Background} conducts a review of VLM-based detection and segmentation related work; Sec.~\ref{sec:vlm-based detection task} and Sec.~\ref{sec:vlm-based segmentation task} present the detection and segmentation evaluation results and corresponding analysis across various tasks; Sec.~\ref{sec:future} outlines potential future directions for VLM development. Finally, Sec.~\ref{sec:conclusion} concludes the paper and summarizes the key contributions of this work.

\section{Background}
\label{sec:Background}

\begin{figure}
    \centering
    \includegraphics[width=1\linewidth]{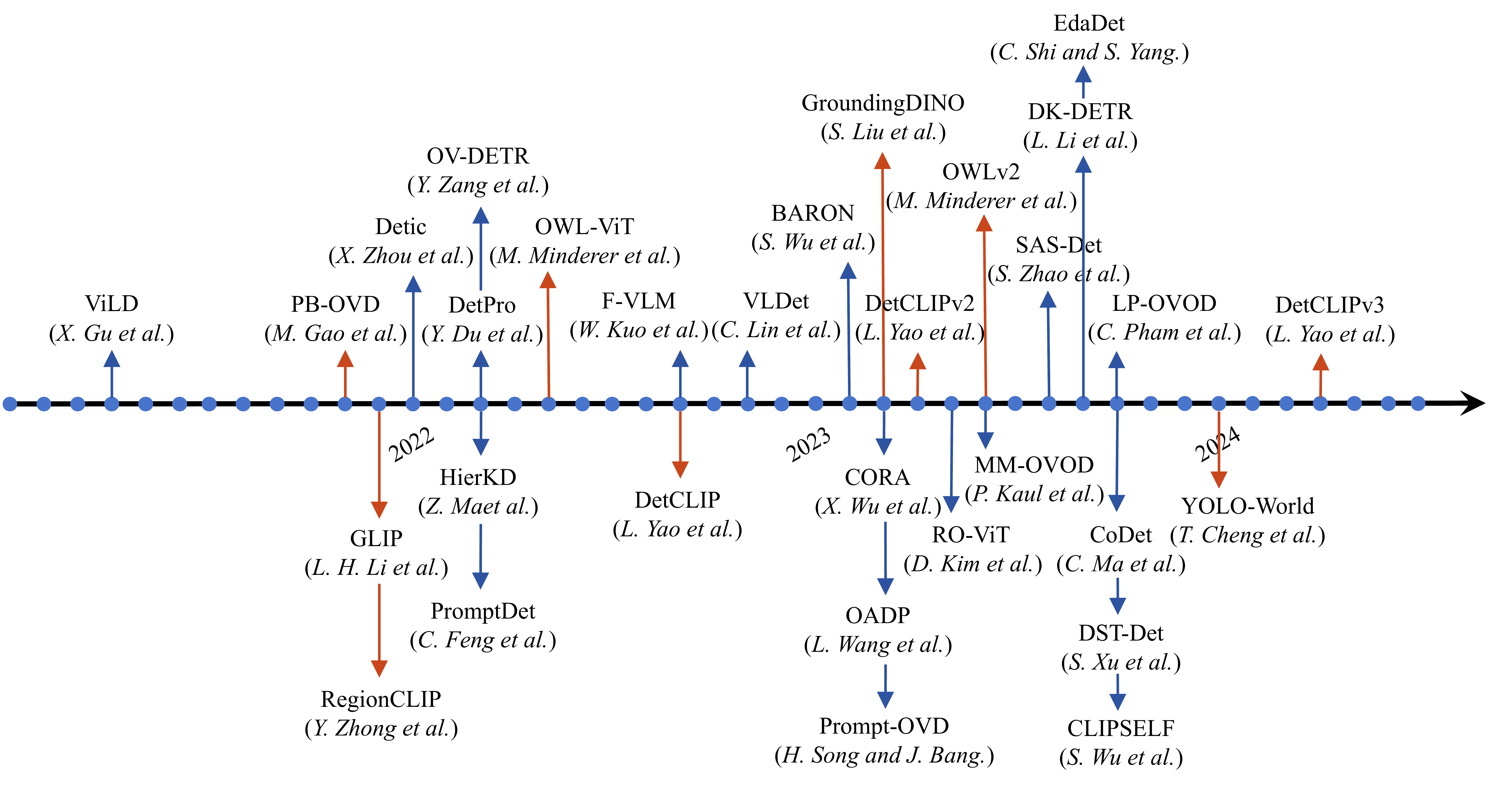}
    \caption{The timeline of VLM-based detection methods.}
    \label{fig:detector_timeline}
\end{figure}

\subsection{VLM-based Detection Method}

Although traditional object detection methods have achieved success under the supervised learning paradigm, they still encounter challenges when detecting arbitrary objects in open-world scenarios. 
Vision-language-based detection, also known as open-vocabulary object detection (OVD), offers a promising solution to this challenge. By introducing the text modality and leveraging pre-training on large-scale multimodal datasets, OVD aligns visual and textual modalities, enabling the detection of corresponding objects based on arbitrary text inputs.

\begin{figure}
    \centering
    \includegraphics[width=1\linewidth]{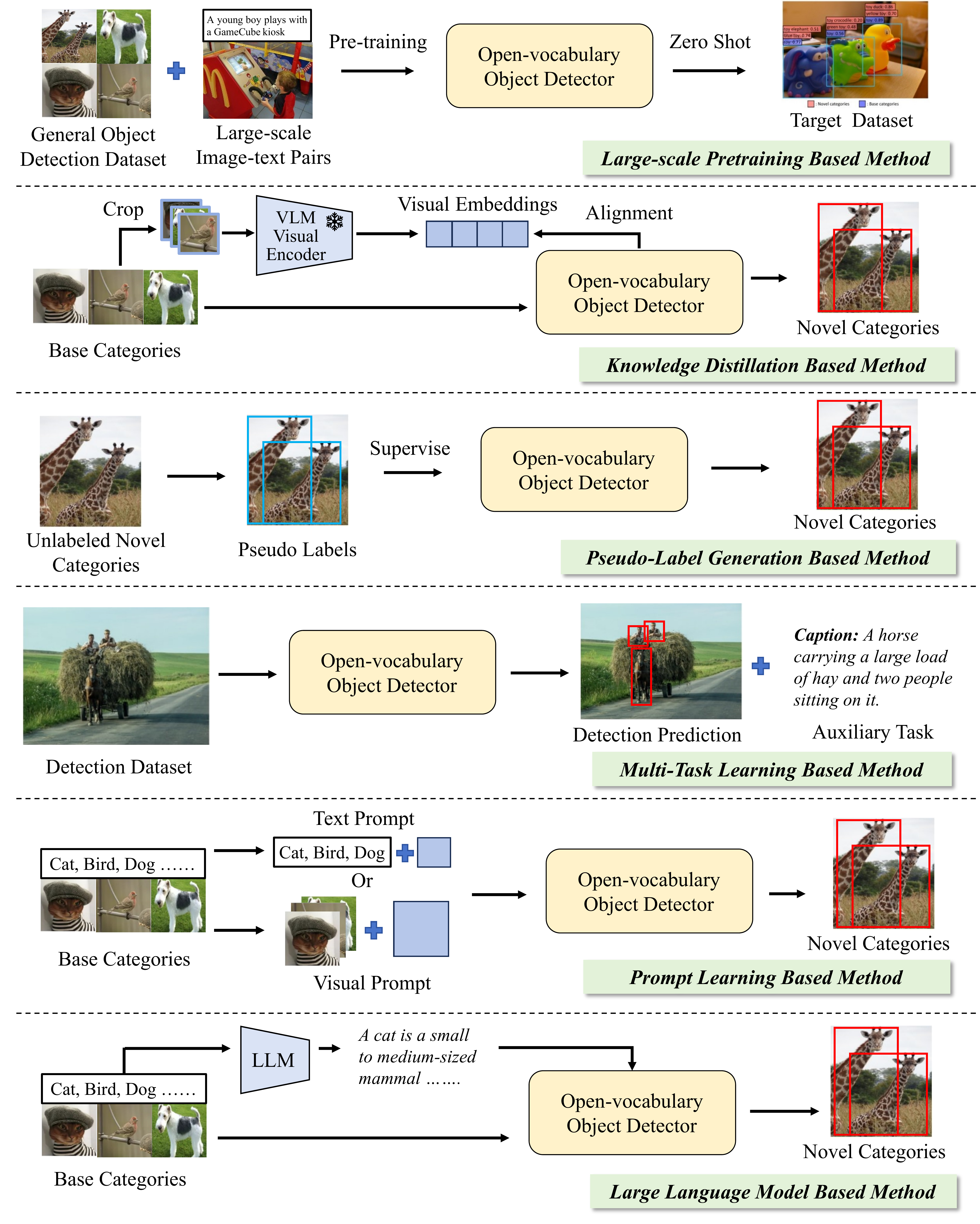}
    \caption{Different types of VLM-based Detection Methods. Large-scale Pretraining Based Methods are trained on large-scale dataset to improve zero-shot performance on rare categories. The rest types of methods utilize learning strategies for specific open vocabulary datasets and are collectively classified as Learning Strategy Based Method.}
    \label{fig:6methods}
\end{figure}

Among the VLM-based detection methods, some methods collect large-scale datasets and pre-train them to get impressive zero-shot performance. We refer to these methods as Large-scale Pretraining Based Method. Meanwhile, many methods design learning strategies for specific open vocabulary datasets, such as OV-COCO or OV-LVIS. The learning strategies include knowledge distillation, pseudo-label generation, multi-task learning, prompt learning and large language model assistance, which are collectively classified as Learning Strategy Based Method. The basic details of Large-scale Pretraining Based Method and Learning Strategy Based Method are presented in the Table \ref{tab:first_method} and \ref{tab:second_method}, respectively. The timeline of VLM-based detection methods is shown in Fig.~\ref{fig:detector_timeline}, and the illustrations of those types of methods are shown in Fig.~\ref{fig:6methods}.

\input{tables/ovd_first_model_summary}
\input{tables/ovd_second_model_summary}

\subsubsection{Large-scale Pretraining Based Method}
In recent years, the large-scale data pre-training method has shown a strong representation learning ability, which is also suitable for open vocabulary detection. By pre-training on large-scale data, the model can learn rich visual and semantic features, which helps to improve the generalization ability of unknown categories. 

CLIP~\cite{radford2021learningtransferablevisualmodels} effectively learns image-level representations by pre-training on a large number of image-text pairs and achieves excellent performance in zero-sample classification tasks. However, some fine-grained visual tasks require region-level representation. GLIP~\cite{li2022grounded} reconstructs the object detection task into a phrase location task, linking all candidate categories as text input in addition to image input. In this way, the problem of candidate region classification is transformed into the problem of alignment between candidate regions and words, thus unifying detection and phrase localization tasks. RegionCLIP~\cite{zhong2022regionclip} is designed to extend CLIP to learn region-level visual representations, enabling fine-grained alignment between image areas and text concepts. PB-OVD~\cite{gao2022open} processes the activation map of images to automatically obtain the pseudo bounding-boxes of diverse objects from large-scale image-caption pairs. DetCLIP~\cite{yao2022detclip} proposes a parallel concept representation to make better use of heterogeneous data, encoding different forms of detection data, positioning data, and graphic data to maximize the use of large data sets for pre-training. DetCLIPv2~\cite{yao2023detclipv2} optimizes the training process of DetCLIP, which utilizes 13× more image-text pairs while requiring only a similar training time. DetCLIPv3 uses Visual Large Language Model to build Auto-annotation data pipeline that refines the annotations of image text pairs to provide higher quality data for pre-training. GroundingDINO~\cite{liu2023grounding} adds cross-modal fusion to the image and text encoding phase, the query selection phase, and the final decoding phase to achieve more powerful performance. YOLO-World~\cite{cheng2024yolo} proposed a visual language path aggregation network, which uses text-guided CSPLayer to inject text information into image features and uses Image Pooling Attention mechanism to enhance the text embedding of image perception. OV-DINO~\cite{wang2024ov} introduces a Unified Data Integration pipeline to unify different data sources into a detection-centered data form to eliminate data noise caused by pseudo labels.

\subsubsection{Knowledge Distillation Based Method}
Distilling the knowledge from the visual encoder of pretrained VLMs makes the open vocabulary detection models easier to establish associations with text embeddings obtained by the text encoder of VLMs, which can effectively improve the ability to recognize unseen categories.

ViLD~\cite{gu2022openvocabularyobjectdetectionvision} first distills the knowledge of VLM to the two-stage detector Mask R-CNN \cite{he2017mask} by aligning the features of the proposal regions with the image embeddings obtained by utilizing the image encoder of VLM. HierKD~\cite{ma2022open} applies knowledge distillation on one-stage detector and also introduces the global stage distillation method, which aligns the text features of image captions with the global image features. DK-DETR~\cite{li2023distilling} chooses the Deformable DETR as the student model and treats the feature alignment between detector and VLM as a pseudo-classification problem, narrowing the feature distance belonging to the same object and stretching the feature distance belonging to different objects. OADP~\cite{wang2023object} analyzes the problems of comprehensiveness and purity in the process of cutting candidate regions and neglectfulness of global scene understanding in the process of knowledge distillation and makes up for the lack of global scene understanding through multi-scale distillation. BARON aligns the embedding of bag of regions instead of individual regions to the embeddings of words in a sentence obtained by utilizing the text encoder of a VLM.

\subsubsection{Pseudo-Label Generation Based Method}
In addition to leveraging visual-language models (VLMs) for knowledge distillation, utilizing their powerful cross-modal representation capabilities to generate pseudo-labels for images is also an effective approach in open-vocabulary detection. By automatically generating labels for unlabeled regions in the images, VLMs can enhance the training data in unsupervised or weakly supervised settings, thereby improving the model's capability to recognize unknown categories. This method not only reduces the reliance on manual annotations but also enables rapid expansion of the model's recognition scope on large-scale datasets.

Zhao et al. \cite{zhao2022exploiting} proposed a simpler pseudo-label generation approach by directly applying a class-agnostic RPN network to extract candidate regions and using the VLM to classify these regions. To ensure high-quality pseudo-labels, they applied repeated RoI operations and used a filtering process that combined the RPN scores with the predictions from the VLM. Apart from leveraging pre-trained VLMs, self-training with teacher-student architectures is another widely used approach for pseudo-label utilization. Zhao et al. \cite{zhao2024taming} proposed SAS-Det, where a teacher network generates pseudo-labels to train a student network, and the student periodically updates the teacher. In addition to the two-stage methods for pseudo-label generation, Xu et al. \cite{xu2023dst} proposed an end-to-end training framework called DST-Det, which dynamically generates pseudo-labels during the training process using VLMs. During the RPN stage, these regions are treated as foreground objects, while at the final classification stage, the corresponding novel categories are added directly to the classification targets.

\subsubsection{Multi-Task Learning Based Method}
Joint training with other tasks in open-vocabulary detection not only enriches training data but also introduces additional task constraints, enhancing the model's generalization ability. Multi-task learning enables knowledge sharing across tasks, allowing the model to leverage complementary information to improve recognition performance for unknown categories.

Given that object detection inherently involves localization and classification, combining detection and classification tasks is an intuitive approach. Zhou et al. \cite{zhou2022detecting} proposed Detic, which applies image-level supervision to the largest candidate region for classification data while following standard detection losses for detection data. By leveraging the extensive vocabulary of classification datasets, Detic significantly enhances open-vocabulary detection performance without introducing additional losses. Joint training of detection and segmentation has also been explored, though prior work, such as Mask R-CNN \cite{he2017mask}, is limited to closed-set models with aligned bounding box and mask annotations. Zhang et al. \cite{zhang2023simple} introduced OpenSeeD to address the challenges of open-vocabulary detection and segmentation. OpenSeeD divides decoder queries into foreground and background queries, enabling foreground detection and background segmentation. It also introduces conditional mask decoding to learn masks from segmentation data and generate masks for detection data. This unified framework improves performance in both open-vocabulary detection and segmentation by combining data and task supervision. In addition, Long et al. \cite{long2023capdet} proposed CapDet, which jointly trains detection with dense captioning, where detection losses and captioning losses jointly constrain the training process. This approach benefits detection from the rich language concepts in captioning data and allows the model to predict category-free labels, achieving true open-vocabulary detection.

\subsubsection{Prompt Learning Based Method}
Prompt learning is an effective technique for adapting foundation models to different domains. By incorporating learned prompts into the foundation model, the knowledge of the model can be more easily transferred to downstream tasks. This approach has also been applied to open-vocabulary detection, where prompts guide the model to achieve stronger generalization on unknown categories.

Du et al. \cite{du2022learning} proposed DetPro, which introduces a set of shared learnable parameters that are prepended to the embeddings of each category name. For a given image, a class-agnostic RPN is employed to extract candidate regions. Positive candidates are guided to align more closely with the embeddings of their corresponding ground-truth category, while negative candidates are pushed further away from all category embeddings, enabling the model to effectively learn generalized prompts. Similarly, PromptDet \cite{feng2022promptdet} introduces prompts on the text side but focuses on improving semantic clarity and flexibility. This method appends descriptive phrases to each category name to reduce ambiguity and incorporates learnable parameters into the generated text embeddings. Additionally, it leverages web-crawled image-text pairs to expand the vocabulary with new categories and allows the learned prompts to be iteratively refined for better performance. Beyond adding learnable prompts on the text side, Wu et al. \cite{wu2023cora} proposed CORA, a DETR-based detector that incorporates learnable prompts on the image side to adapt CLIP for open-vocabulary detection. It features two key modules: a Region Prompt Module, which aligns the CLIP image encoder with region-level features to address distribution mismatches, and an Anchor Pre-Matching Module, which associates object queries with dynamic anchor boxes to enable class-aware regression.

\subsubsection{Large Language Model Based Method}
With the exceptional generalization and reasoning abilities demonstrated by large language models (LLMs) across various tasks, leveraging LLMs for auxiliary training has become a key direction in open-vocabulary detection. The extensive knowledge base and cross-modal understanding capabilities of LLMs provide robust support for open-vocabulary detection, especially under limited annotations, allowing models to better recognize unseen categories and handle complex scenarios.

Kaul et al. \cite{kaul2023multi} proposed an open-vocabulary detector with a multimodal classification head that supports category descriptions through text, images, or their combination. Text descriptions are generated using GPT-3 \cite{brown2020language} to create multiple rich descriptions per category, averaged into a text feature. Image descriptions are obtained by processing category-specific images through a VLM image encoder and aggregating their features with a Transformer. Text and image features are then fused via weighted averaging to enable detection based on multimodal inputs. Similarly, Jin et al. \cite{jin2024llms} proposed DVDet, which enhances detection by generating fine-grained descriptors for each category. Candidate regions compute similarity with a fixed number of descriptors, and descriptors are dynamically optimized during training by retaining frequently used ones and discarding rarely used ones. For confusing categories, an LLM generates distinguishing descriptors that are added to refine classification. To address the limitations of CLIP's text space, which lacks detailed textual and visual information and tends to overfit base categories, Du et al. \cite{du2025lami} proposed LaMI-DETR. This method uses GPT-3.5 \cite{jin2024llms} to generate rich visual descriptions, transforming class names into comprehensive visual concepts. These concepts are grouped with T5 \cite{raffel2020exploring}, and categories from different groups are sampled during training to encourage learning generalized foreground features. During inference, visual descriptions assist in distinguishing confusing categories, enhancing performance on unseen objects.

Among the six types of VLM-based detection methods, Large-scale Pretraining Based Methods utilize a wide variety of datasets for pertaining and generally obtain better generalization ability to different detection tasks. Meanwhile, the other five types of methods, which can be collectively classified as Learning Strategy Based Methods, focus on learning specific open-vocabulary datasets, such as OV-COCO and OV-LVIS. Consequently, we evaluate both Large-scale Pretraining Based Methods and Learning Strategy Based Methods on open-vocabulary related detection tasks and additionally evaluate the performance of Large-scale Pretraining Based Methods on more detection tasks.

\subsection{VLM-based Segmentation Method}

\begin{figure}
    \centering
    \includegraphics[width=\columnwidth]{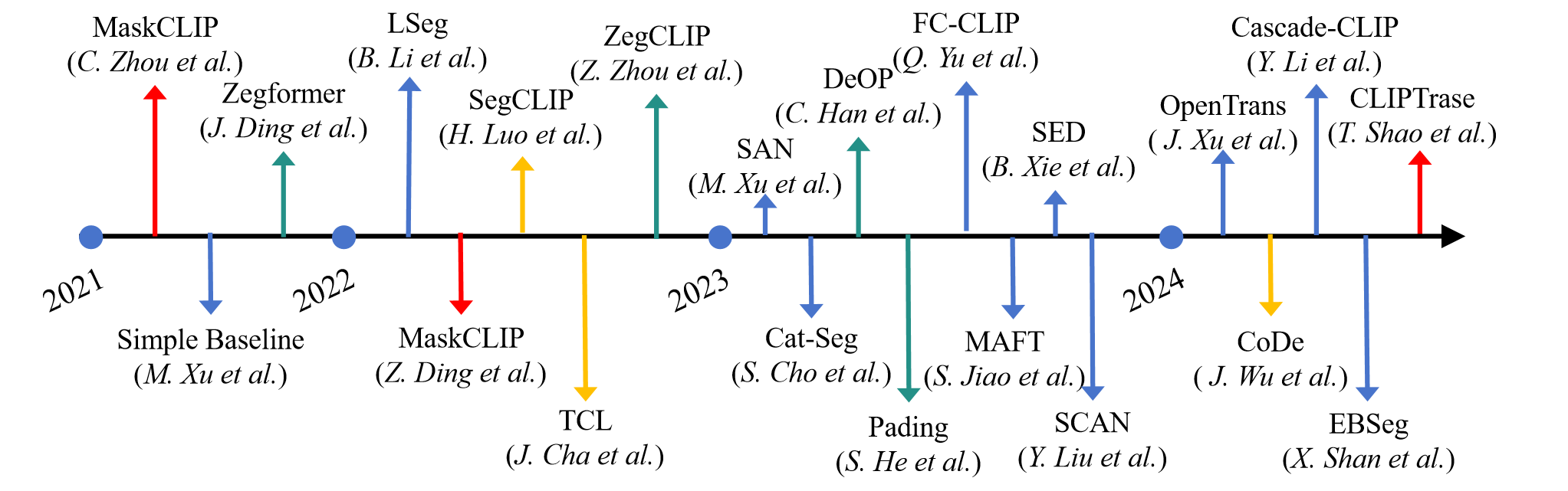}
    \caption{The timeline of VLM-based segmentation methods.}
    \label{fig:segment_timeline}
\end{figure}

\begin{figure}
    \centering
    \includegraphics[width=1\linewidth]{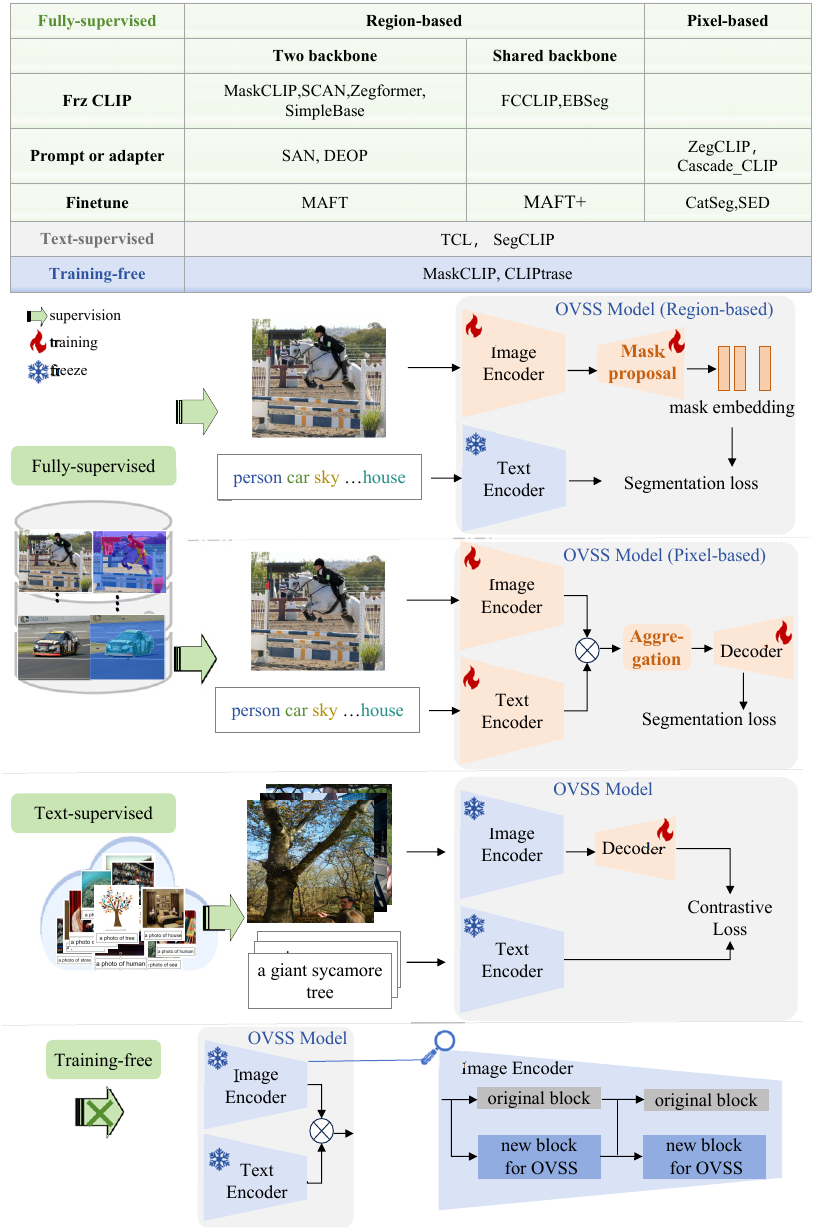}
    \caption{Different types of VLM-based Open-Vocabulary Segmentation methods. Current VLM-based open-vocabulary segmentation methods fall into three categories depending on training supervision: fully-supervised, text-supervised, and training-free approaches. The fully-supervised category is further classified by model design.}
    \label{fig:ovss_methods}
\end{figure}

To fully harness the robust open-vocabulary understanding capabilities of CLIP for dense prediction tasks, existing works have employed various types of supervision, as illustrated in Fig.~\ref{fig:segment_timeline} and Fig.~\ref{fig:ovss_methods}. These include: (1) dense annotations on limited categories, (2) large-scale image-text pairs, and (3) unsupervised methods. The following sections are organized according to these three types of supervision.
\subsubsection{Fully-supervised Open-Vocabulary Semantic Segmentation}
To enhancec the segmentation capabilities of CLIP, open-vocabulary semantic segmentation models learn from dense annotations from limited categories, exemplified by the 171 categories available in the COCO-Stuff datasets.

Two-stage methods first generate class-agnostic mask proposals and then leverage pre-trained vision-language models, e.g., CLIP, to classify masked regions. OVseg~\cite{liang2023open_ovseg} identifies the performance bottleneck of the two-stage paradigm is that the pretrained CLIP model does not perform well on masked images and proposes to finetune CLIP on a collection of masked image regions and their corresponding text descriptions by mask prompt tuning. To avoid the time-consuming operation to crop image patches and compute feature from an external CLIP image model, MaskCLIP~\cite{ding2022open} introduces the Mask Class Tokens for efficient feature extraction and each Mask Class Token learns from the corresponding mask area of the images. SAN~\cite{xu2023side} attaches a side network to a frozen CLIP model with two branches: one for predicting mask proposals, and the other for predicting attention bias which is applied in the CLIP model to recognize the class of masks. They propose the [SLS] tokens which adopt the similar design with Mask Class Token in MaskCLIP. DeOP~\cite{han2023open} introduces the Generalized Patch Severance to harmful interference between patch tokens in the encoder and Classification Anchor Learning module to find patches that to be focused in the spatial pooling for classification. SCAN~\cite{liu2024open} employs a semantic integration module designed to incorporate the global semantic perception of original CLIP into proposal embedding to mitigate domain bias caused by unnatural background and providing global context.

FCCLIP~\cite{yu2024convolutions} proposes a sing-stage framework, which builds both mask generator and CLIP classifier on top of a shared Frozen Convolutional CLIP backbone ans consists of three modules: a class-agnostic mask generator, an in-vocabulary classifier, and an out-of-vocabulary classifier. To maintain the CLIP’s zero-shot transferability, previous practices favour to freeze CLIP during training. MAFT~\cite{jiao2023learning} reveals that CLIP is insensitive to different mask proposals and tends to produce similar predictions for various mask proposals of the same image and proposes to finetune CLIP with mask-aware loss and self-distillation loss. To achieves vision-text collaborative optimization, MAFT+~\cite{jiao2025collaborative_maft+} incorporates CLIP-T into the fine-tuning process to concurrently optimize the text representation. This vision-text joint optimization alleviates the training complexity and enhances the vision and text alignment. In contrast to the two-stage paradigm that utilizing mask proposal generators, CAT-Seg~\cite{cho2024cat} investigate methods to transfer the holistic understanding capability of images to the pixel-level task of segmentation. They propose a cost aggregation-based framework which consists of spatial and class aggregation to reason the multi-modal cost volume. In light of CAT-Seg, SED~\cite{xie2024sed} comprises a hierarchical encoder-based cost map generation and a gradual fusion decoder with category early rejection.

\input{tables/seg_model_summary}

\subsubsection{Text-Supervised Open-Vocabulary Segmentation}

To address the high cost of traditional methods relying on dense mask annotations, text-supervised approaches propose region-level alignment using image-text pairs, enabling image segmentation with solely text supervision. Text-supervised open-vocabulary segmentation commonly employs contrastive loss between image and text to project image feature embeddings and text feature embeddings into a shared space, thereby enabling further classification of image segmentation proposals. Due to the lack of annotation supervision of dense region masks, all text-supervised segmenters adopt pixel-based perception methods.
TCL \cite{cha2023learning} employs a Text-Grounded Decoder to perform upsampling and convolutional processing on image patches, generating pixel-level feature maps. TCL then conducts contrastive learning \cite{he2020momentum} between pixel-level features and text features to achieve object segmentation. SegCLIP \cite{luo2023segclip} introduces a Semantic Group Module to aggregate image patches into arbitrary-shaped semantic regions, it dynamically aggregates image patches using learnable central queries and cross-attention, then aligns the aggregated patches with text for segmentation. Additionally, SegCLIP incorporates the MAE \cite{he2022masked} image reconstruction loss and superpixel KL loss \cite{felzenszwalb2004efficient} to assist the learning process.

\subsubsection{Training-free Open-Vocabulary Segmentation}

Training-free open-vocabulary segmentation models typically generate mask proposals using methods such as clustering method and class-agnostic mask proposal network, while refining these mask proposals using attention weights and pixel-level similarity scores. CLIPtrase \cite{shao2024explore} uses DBSCAN \cite{khan2014dbscan} to directly cluster the image to obtain the object mask. To refine the mask obtained from direct clustering, CLIPtrase enhances the attention of different image patches to other image patches within the same semantic region through the Semantic relevance restoration module, and selectively discards some noisy clusters based on the attention, thus obtaining a refined mask. Instead of using a clustering method, MaskCLIP \cite{ding2022open} firstly trains a category-agnostic mask proposal network. When migrating to open-vocabulary segmentation tasks, MaskCLIP integrates RMA module into the backbone network. RMA module refines the proposed mask based on the attention weights between the mask and the image patches.

Our taxonomy of VLM-based semantic segmentation methods is presented in Tab~\ref{tab:seg_methods}.

\section{VLM-based detection task}
\label{sec:vlm-based detection task}


\subsection{General Closed-Set Evaluation}

Closed-set object detection remains the most widely adopted evaluation paradigm in object detection, wherein both training and testing are conducted on the same predefined set of categories, allowing for an effective assessment of a model’s fundamental detection capabilities.
Although VLMs, trained on large-scale datasets, demonstrate strong zero-shot performance on common object detection benchmarks \cite{lin2014microsoft, Pascal_VOC, gupta2019lvis}, their closed-set performance is highly dependent on the composition of the pretraining data, raising concerns about the fairness of direct comparisons. 
Therefore, we investigate the detection capabilities of VLMs after finetuning (visual and text prompt finetuning) to evaluate their potential as foundational detection models. 
For comparison, we also assess the performance of traditional detection models as reference baselines. 
As shown in Table \ref{tab:det_general_closed}, we draw conclusion as following:

\input{tables/det_general_closed}


(1) The performance of traditional methods improves progressively as architectures evolve. 
Faster R-CNN \cite{ren2016faster}, as a representative of the traditional two-stage detection paradigm, established a foundational object detection framework.
However, its dependence on region-based feature extraction and proposal generation limits its performance on challenging datasets (e.g., LVIS \cite{gupta2019lvis}). 
The YOLO-v8 \cite{yolov8_ultralytics}, following the single-stage detection paradigm, has undergone continuous iterations, consistently outperforms Faster R-CNN. 
Dynamic Head \cite{dai2021dynamic}, on the other hand, introduces dynamic attention mechanisms, demonstrating superior performance compared to YOLO-v8. 
DINO \cite{zhang2022dino} fundamentally disrupts traditional paradigms by fully embracing a Transformer-based end-to-end architecture, achieving the highest performance across all datasets, highlighting the pivotal role of global feature expression and Transformer-based adaptive modeling in advancing detection capabilities.

(2) The performance of OVD methods heavily also depends on the underlying detector architecture. 
RegionCLIP \cite{zhong2022regionclip} and PB-OVD \cite{gao2022open}, based on the traditional Faster R-CNN  \cite{ren2016faster} architecture, encounter limitations due to their relatively outdated feature extraction frameworks, leading to suboptimal performance on complex datasets.
GLIP \cite{li2022grounded}, built upon the Dynamic Head \cite{dai2021dynamic}, integrates visual-text alignment through unified training, demonstrating robust closed-set performance.
YOLO-World \cite{cheng2024yolo}, built on YOLO-v8 \cite{yolov8_ultralytics}, retains the computational efficiency of single-stage detectors, though its performance remains slightly inferior to that of GLIP. 
Grounding-DINO \cite{liu2023grounding} and OV-DINO \cite{wang2024ov} introduce the deep visual-text interaction mechanism based on the Transformer-based DINO \cite{zhang2022dino} architecture, significantly enhancing feature alignment and multi-modal semantic modeling. 
These models achieve best closed-set performance across complex datasets, validating the importance of underlying architecture. 

(3) Visual finetuning outperforms text prompt fine-tuning, particularly in more intricate datasets like COCO \cite{lin2014microsoft} and LVIS \cite{gupta2019lvis}. 
The effectiveness of visual finetuning lies in its direct optimization of visual representation, enabling better capture of object shapes, textures, and local details. 
The enhancement in visual representation discriminability leads to a more pronounced improvement on the long-tail dataset LVIS \cite{gupta2019lvis}. 
Text fine-tuning primarily improves semantic alignment and generalization, providing limited benefits in simpler datasets (e.g., VOC \cite{Pascal_VOC}). 
These observations underscore that visual feature modeling remains the primary driver of performance improvements, with text optimization serving as a complementary tool to visual refinement.


\subsection{General Open Vocabulary Evaluation}
The general open Vocabulary detection task aims to evaluate the model's ability to detect uncommon categories, which is important in practical applications. COCO~\cite{lin2014microsoft} and LVIS~\cite{gupta2019lvis} detection datasets are commonly used benchmarks of open-vocabulary detection. During the evaluation, the categories of COCO are split into ``base'' and ``novel'', which means the base categories are easier to encounter than the novel categories. Meanwhile, base categories of LVIS are marked as ``common'' and ``frequent'' and its novel categories are marked as ``rare''. LVIS \textit{minival} shares the same categories with LVIS, but uses a subset of LVIS's test set as its test set. For Large-scale Pretraining Based Methods, they are first evaluated with zero prediction setting to show their original open-vocabulary performance. After that, these methods are trained on the training set of datasets with only the base categories in the setting of visual fine-tuning. For Learning Strategy Based Methods, they have access the images of the target dataset during training and we directly report the performance of official models.
\input{tables/det_ovd}
\input{tables/det_ovd2}

The results of the Large-scale Pretraining Based Methods and Learning Strategy Based Methods are shown in Tab.~\ref{tab:det_ovd} and Tab.~\ref{tab:det_ovd2}. From the experimental results in the table, we draw conclusion as following:

(1) OV-DINO and Grouding-DINO achieve cutting-edge open vocabulary performance of the Large-scale Pretraining Based Methods, which indicates that the DINO detection framework also shows significant advantages for open vocabulary detection tasks. On the other hand, YOLO-World also shows competitive performance while keeping real-time inference speed, demonstrating the potential of the YOLO framework in open vocabulary detection tasks. Among the Learning Strategy Based Methods, LAMI-DETR has the best performance in open vocabulary detection accuracy, which is attributed to the use of large language model to cluster the potentially confusing categories and the design of a special loss to distinguish the easily confused categories.

(2) Comparing the performance on OV-COCO benchmark, Large-scale Pretraining Based Methods have obvious advantages over the Learning Strategy Based Methods. For instance, the $AP_{novel}$ of OV-DINO achieves 76.2\%, much higher than the 46.7\% obtained by DST-Det. However, there is no large gap between the performance of those two types of methods on OV-LVIS benchmark. We think this is due to the fact that the novel categories of COCO dataset are relatively common and frequently appear in the pre-training dataset of the first type of methods. On the contrary, the novel categories of LVIS are much rarer, which is more beneficial for the Learning Strategy Based Methods that are able to exploit LVIS data.

(3) Compare the Zero-prediction and visual finetuning performance of the Large-scale Pretraining Based Methods in Table~\ref{tab:det_ovd}, it can be found that visual finetuning improves the accuracy of the Large-scale Pretraining Based Methods in base categories significantly, while the performance of several methods in novel categories is likely to be decreased. It indicates that simply applying visual finetuning on base categories can lead to catastrophic forgetting and affect the generalization performance of the model. 

(4) The amount of pre-training datasets is another factor that affects the performance of the Large-scale Pretraining Based Methods in open vocabulary detection. A larger pre-training dataset will provide the model with more samples containing rare semantics categories, so that the model can obtain better open vocabulary detection capabilities. For instance, the OV-DINO (A) is pretrained on Object365 dataset, while OV-DINO (B) is pretrained on both Object365 and GoldG dataset and get higher metrics on the $AP_r$ of LVIS \textit{minival} dataset. 

\subsection{Open Vocabulary Generalization Evaluation}
Open vocabulary generalization evaluation is generally adopted by the second type of models. It measures the model's generalization on other data sets after fine-tuning open vocabulary on a single data set, which includes not only the generalization of domain but also the generalization of categories. The generalization ability is evaluated under several settings, such as testing on the PASCAL VOC datasets~\cite{everingham2015pascal}, LVIS datasets~\cite{gupta2019lvis} and Objects365 dataset~\cite{shao2019objects365} after being fine-tuned on the base categories of COCO. It should be noted that the datasets selected for testing should not be used to pretrain or fine-tune the detectors. The detection accuracy in the unseen datasets reflects the usability of Open Vocabulary detectors in practice.
\input{tables/det_generalization}

The results of the Large-scale Pretraining Based Methods and Learning Strategy Based Methods are shown in Tab.~\ref{tab:det_generalization}. From the experimental results in the table, the following conclusions can be drawn:

(1) Open vocabulary generalization of Large-scale Pretraining Based Methods is generally higher than that of Learning Strategy Based Methods. For instance, OV-DINO, a representative of the Large-scale Pretraining Based Methods, achieves 47.9\% AP on LVIS$\rightarrow$COCO setting, while LAMI-DETR, the second-type method with the highest accuracy, only achieves 42.8\%. It indicates that a larger pre-training dataset will contain more data from different domains that carry more semantic categories, which can effectively help improve both the domain and category generalization capability of the model, making it easier to achieve better generalization performance. 

(2) When the model generalizes from a dataset with more categories to a simple dataset with less categories (e.g. COCO$\rightarrow$VOC), the VLM-based detectors can generally achieve high accuracy. Conversely, when it generalizes from a dataset with less categories to a dataset with more categories (e.g. COCO$\rightarrow$Object365), a low accuracy performance is likely to be obtained. This comparison indicates that using a semantically rich dataset for pre-training can significantly improve the generalization ability of the model.

\subsection{Domain-Related Evaluation}
Domain-related detection is a classic evaluation task, typically categorized into domain adaptive object detection (DAOD) and domain generalization object detection (DGOD). DAOD consists of a single source domain and a target domain, while DGOD involves a single source and multiple target domains.
Under \textit{zero prediction}, models are directly evaluated on the target domain(s). In \textit{visual fine-tuning} and \textit{text prompt}, models are fine-tuned on the source domain and then tested on one or multiple target domains. This evaluation assesses VLM robustness and adaptation capability to out-of-distribution data, while providing optimization guidance for domain-aware VLM detection algorithms.
\subsubsection{Domain Adaptation Settings}
The DAOD settings includes two scenarios: autonomous driving and natural images. For autonomous driving, it involves three domain adaptation settings: Cityscapes$ \rightarrow $FoggyCityscapes, Sim10k$ \rightarrow $Cityscapes, and KITTI$ \leftrightarrows $Cityscapes. In Cityscapes$ \rightarrow $FoggyCityscapes, Cityscapes~\cite{cordts2016cityscapes} (2,975 road images with 8 object classes) serves as the source domain, while FoggyCityscapes~\cite{sakaridis2018semantic} (500 foggy images) acts as the target domain, testing adaptation under weather changes. Sim10k$ \rightarrow $Cityscapes evaluates adaptation from synthetic (Sim10k~\cite{johnson2016driving}, 10K images) to real (Cityscapes) data, focusing solely on the car class. KITTI$ \leftrightarrows $Cityscapes (KITTI, 7,481 images) assesses cross-camera viewpoint adaptation due to differing sensor configurations and shooting angle. The natural image adaptation scenario examines domain adaptation under large stylistic shifts for general object detection, covering two sub-tasks: Pascal VOC$ \rightarrow $ Watercolor and Pascal VOC$ \rightarrow $ Comic. The Pascal VOC ~\cite{everingham2015pascal} dataset contains 16,551 real-world training images across 20 object categories. For the Watercolor~\cite{inoue2018cross} and Comic~\cite{inoue2018cross} target domains, each consists of 1,000 stylized images sharing 6 categories with Pascal VOC (bike, bird, car, cat, dog, person) for testing. This setup evaluates model generalization from realistic to artistic image styles.

\input{tables/voc2clip}

\subsubsection{Domain Generalization Settings}
To evaluate VLMs, we adopt the same datasets used in ~\cite{wu2022single}, comprising five distinct weather-condition sets: Day Clear, Night Clear, Dusk Rainy, Night Rainy, and Day Foggy. These images are sourced from three main datasets: Berkeley Deep Drive 100K (BBD-100K) ~\cite{yu2020bdd100k}, Cityscapes ~\cite{cordts2016cityscapes}, and Adverse-Weather ~\cite{hassaballah2020vehicle}, supplemented by synthetically rendered rainy images from ~\cite{wu2021vector} and artificially generated foggy images from ~\cite{sakaridis2018semantic}. Training is conducted exclusively on 19,395 day clear images, with an additional 8,313 sunny images reserved for validation and model selection. Testing employs the remaining four weather conditions: 26,158 night clear images, 3,501 dusk rainy images, 2,494 night rainy images, and 3,775 day foggy images. All datasets provide bounding box annotations for seven object categories: bus, bike, car, motorbike, person, rider, and truck. 

\input{tables/domain_generalization}

As shown in Tab.~\ref{tab:daod} and Tab.~\ref{tab:dgod}, we draw conclusion as following:
(1) Compared with traditional methods, VLMs demonstrate superior capabilities in domain adaptation and generalization tasks, primarily due to their exposure to diverse cross-domain scenarios during pre-training. This inherent advantage enables VLMs to outperform traditional models pre-trained on ImageNet in cross-domain adaptation tasks. However, despite the extensive cross-domain knowledge accumulated during pre-training, task-specific adaptations remain crucial. Empirical results indicate that \textit{Visual Fine-tuning} significantly enhances model performance on target domains, underscoring the necessity of task-specific adaptation. However, it is important to note that while VLMs exhibit strong cross-domain adaptation abilities, their performance still falls short of specialized domain adaptation methods, indicating potential for improvement.

(2) \textit{Text Prompt} has been shown to effectively improve model performance in most domain adaptation scenarios. However, their effectiveness is setting-dependent, as evidenced by performance drops in certain domain adaptation settings and datasets (e.g., GLIP and GroundingDINO models on the Cityscapes$ \rightarrow $Foggy Cityscapes adaptation). Conversely, text prompts exhibit notable advantages in scenarios where domain shifts are primarily due to viewpoint transformations and the number of target categories is limited. This highlights the importance of aligning prompt strategies with specific domain adaptation.

(3) The performance variations among different VLMs in domain adaptation tasks can be attributed to their cross-modal feature fusion capabilities. By integrating visual and textual features across different levels (e.g., texture, color, shape, global features), VLMs enhance their ability to capture domain-related representations, thereby improving domain generalization. For instance, models like YOLO-World, GLIP, and GroundingDINO demonstrate progressively stronger domain adaptation capabilities as the intensity of feature fusion increases. 

(4) Further analysis reveals that models with larger parameter sizes and more extensive pre-training data demonstrate superior performance in domain adaptation tasks. This is because larger models can better approximate complex data distributions, while models with diverse pre-training data are more likely to adapt to unseen domain shifts. Consequently, these models exhibit stronger advantages in domain adaptation scenarios.

In summary, VLMs leverage cross-modal alignment to enhance the openness of traditional closed-set detectors while benefiting from pre-training on diverse, textually aligned data. Despite their advantages in domain adaptation, VLMs still require optimization in feature fusion mechanisms, prompt strategies, and cross-modal alignment to fully realize their potential in domain adaptation and generalization tasks.

\subsection{Few-Shot Evaluation}
Although open-vocabulary detectors have been trained on large-scale datasets, they still face challenges in handling significant scene variations and infrequent object categories. 
In edge cases, supplying a limited set of samples for few-shot learning remains essential for adapting to novel scenarios and objects \cite{xu2024multi, madan2023revisiting}.
To rigorously evaluate the few-shot learning capabilities of the detector, we employ the ODinw \cite{li2022grounded} benchmark for evaluation.
The ODinw (Object Detection in the Wild) \cite{li2022grounded} benchmark consists of 35 sub-datasets, encompassing images and objects from diverse domains, including remote sensing, medical, and biological fields. 
The ODinw benchmark includes two widely used versions, namely ODinw-13 and ODinw-35, which contain 13 categories and 35 categories, respectively.
Compared to common datasets such as COCO and LVIS, the objects in ODinW are rarer and more diverse. 
Thus, despite VLMs exhibiting a certain level of perception for a broad spectrum of objects, effectively detecting these edge-case targets remains a challenge without few-shot fine-tuning or prompt-based adaptation. 
Considering the significant structural differences among detectors, we evaluate with two common finetuning strategies: visual finetuning and text prompt finetuning.
We follow the conventional few-shot object detection setup, training the model with $k$ support samples (i.e. $k$-shot), where $k$ belongs to the set $\{0, 1, 3, 5, 10\}$, and evaluating it across whole test test, where $k = 0$ represents the case of no fine-tuning, serving as the baseline performance. 
As shown in Table \ref{tab:det_few_shot}, we draw conclusion as following:

\input{tables/det_few_shot}

(1) As the number of support samples (shot number) increases, the overall performance of most methods demonstrates an upward trend.
This can be attributed to the fact that additional support samples enhance the models' ability to capture enriched object features, thereby facilitating the learning of category-level semantic information and improving detection accuracy in few-shot scenarios.
However, in certain cases, a performance decline is observed between the 1-shot and 3-shot settings, which may be due to instability in feature alignment and semantic modeling when the number of samples is limited.
For instance, with very few support samples, models may overfit to isolated data points, leading to short-term performance fluctuations.
As the shot number increases further, the model's performance stabilizes, indicating that the inclusion of additional samples effectively mitigates these instabilities.

(2) The model architecture and training data have a significant impact on the performance in few-shot scenarios. 
Compare different models, Grounding DINO \cite{liu2023grounding} consistently outperforms other methods in both zero-shot and few-shot scenarios. 
This significant performance advantage is attributed to its Transformer-based architecture and the utilization of large-scale pretraining data. 
The extensive pretraining imbues the model with rich visual and semantic knowledge, enabling Grounding-DINO to effectively comprehend category semantics and detect objects under zero-shot and few-shot conditions. 
For OV-DINO \cite{wang2024ov}, while its zero-shot performance is not as strong as Grounding-DINO, its performance improves significantly as the number of support samples increases. 
This improvement can be attributed to its strong foundational architecture, derived from DINO. 
In contrast, RegionCLIP \cite{zhong2022regionclip} and PB-OVD \cite{gao2022open} exhibit consistently lower performance, with only limited improvement as the shot number increases. 
This is mainly due to their reliance on relatively traditional detection architectures and simplified visual-text alignment mechanisms. 
These architectures struggle with capturing strong feature representations in few-shot learning tasks, particularly in complex scenarios involving rare categories.
Furthermore, these methods are constrained by the limited scale of pretraining data and the absence of advanced multi-modal interaction mechanism, which restricts their performance and adaptability in few-shot tasks.

(3) Visual finetuning consistently outperforms text prompt finetuning on few-shot tasks. 
Visual finetuning directly optimizes image feature extraction, enabling the model to capture detailed target features such as shape, texture, and spatial information more effectively, which is particularly critical for few-shot object detection.
In contrast, text prompt finetuning primarily focuses on visual-text semantic alignment. 
However, in few-shot settings, where textual cues are limited, this alignment often fails to compensate for the model's deficiencies in feature representation, resulting in overall lower performance.


\subsection{Robustness and Noise Resistance}
Most existing open-vocabulary detection methods focus on generality and are trained on extensive, clean, and high-quality image datasets. 
However, in the real world, image quality is often affected by weather conditions, camera imaging conditions, and other factors \cite{michaelis2019benchmarking}, which poses a significant challenge to the robustness and noise resistance. 

To evaluate the robustness and noise resistance of existing open-vocabulary detectors in real-world scenarios, we conduct a comprehensive benchmark.
The robust benchmark includes 15 types of corruptions across 5 severity levels, designed to assess the impact of a broad range of corruption types on object detection models, including Gaussian noise, shot noise, fog, snow, and others. 
Corruption is implemented through image data augmentation, so theoretically, it can be applied to any dataset. 
In this study, we follow the previous benchmark \cite{michaelis2019benchmarking}, selecting VOC, COCO, and Cityscapes as the three robust evaluation datasets. 
The corrupted versions of these datasets are denoted as VOC-C, COCO-C, and Cityscapes-C, respectively. 
We report clean performance (P$_{clean}$), mean performance under corruption (mPC), and relative performance under corruption (rPC) to measure robustness. 
As shown in Table \ref{tab:det_robust}, we draw conclusion as following:

\input{tables/det_robust}

(1) The dataset complexity has a significant impact on the rPC. 
Simpler datasets, such as VOC-C and Cityscapes-C, which have fewer categories and relatively uniform sample distributions, are easier to detect, resulting in higher rPC scores across most models. 
In contrast, in the more complex dataset COCO-C, the rPC metrics of the models decline compared to simpler datasets.

(2) Observe the P$_{clean}$ metric, it is evident that architecture and dataset scale play a critical role on detection performance. 
The experimental results indicate that Transformer-based models, such as Grounding-DINO \cite{liu2023grounding} deliver the best performance across all datasets, owing to their global modeling and contextual reasoning capabilities. 
In addition, models like YOLO-World \cite{cheng2024yolo} and GLIP leverage their large-scale training datasets to achieve strong performance in simpler scenarios, with YOLO-World even surpassing the Transformer-based Grounding DINO in terms of P$_{clean}$ on VOC-C. 
However, models like RegionCLIP \cite{zhong2022regionclip} and PB-OVD \cite{gao2022open} deliver the weakest performance, both in simple and complex scenarios, highlighting the limitations of their outdated architectures and training frameworks.

(3) Observe the rPC metric, there is no straightforward relationship between model size and robustness. 
For instance, different versions of YOLO-World, GLIP, and OV-DINO \cite{wang2024ov} exhibit only marginal variations in their rPC across datasets.
This suggests that architectural design plays a more pivotal role in enhancing robustness than merely increasing the number of parameters.
Furthermore, while rPC fluctuates substantially across datasets, P$_{clean}$ and rPC exhibit a consistently positive correlation. 
For example, in less challenging conditions (i.e., VOC-C and Cityscapes-C), YOLO-World surpasses other models due to its stable structure.
In contrast, under more complex scenarios (i.e., COCO-C), Transformer-based models capitalize on their capacity to capture high-level global representations, providing significantly better robustness compared to other methods.

(4) Dataset scale has a significant impact on the robustness (rPC) and performance (P$_{clean}$). 
Despite the distinct differences in model architectures, Grounding DINO, YOLO-World, and GLIP all demonstrate remarkable robustness. 
In contrast, RegionCLIP and PB-OVD exhibit poor robustness across all scenarios due to their outdated architectures. 
Overall, P$_{clean}$ and rPC exhibit a general positive correlation, with YOLO-World performing best in simple scenarios and Grounding-DINO demonstrating stronger robustness in complex scenarios.

\subsection{Fine-Grained Perception Capability}
Distinguishing fine-grained semantic information is also an important capability of VLM based detectors. Several fine-grained datasets are widely used to evaluate traditional visual perception methods. Stanford Dogs dataset~\cite{khosla2011novel} collects the images of 120 dog breeds and labels the category and bounding boxes of each sample. Caltech-UCSD Birds-200-2011 dataset~\cite{WahCUB_200_2011} collects the images of 200 species of birds and provides correct annotations too. When using these fine-grained perception datasets for evaluation, the VLM detection models are fine-tuned on the training set and evaluated on the test set. Accordingly, the fine-grained categories are processed by the text encoder to obtain the corresponding embeddings, which are aligned with the image features. The detection accuracy reflects the fine-grain semantic comprehension capability of VLM detection methods.
\input{tables/det_fine_grained}

The results are shown in Tab.~\ref{tab:det_fine_grained}. From the experimental results in the table, we draw conclusion as following:

(1) The zero prediction evaluation of the VLM detectors has poor results. The best performance on Standford Dogs dataset is only 3.5\% AP, which means that these VLM-based detectors can hardly distinguish the fine-grained semantics of a certain category. The reason may be that the semantic granularity during pre-training is not aligned with that during inference, which indicates that the power of zero-shot performance cannot be exercised if the relevant fine-grained semantics are not exposed during pre-training. This reveals that VLM-based methods have not yet established hierarchical multi-granular semantic understanding capabilities.

(2) In stark contrast to zero-pretrain, there is a large accuracy improvement in all of the evaluated methods after visual fine-tuning on the training set of the fine-grained datasets. For instance, the AP of YOLO-World is increased by more than 60\% after visual finetuning. It indicates that the model can quickly learn how to extract key visual features that are crucial for distinguishing fine-grained semantics and align them with text embeddings. This shows that with supervision of semantic granularity alignment, it is not difficult to achieve strong fine-grained perception capabilities.

(3) Compared with visual finetune that upgrades all of the parameters in the visual encoder of the models, the performance of VLM detectors is not largely improved by using Text Prompt. It indicates that only adjusting the category embeddings cannot enhance the fine-grained recognition ability. The reason may be that visual cues play an important role in distinguishing fine-grained semantics. For instance, the color and texture are key criteria for deciding the breed of dogs. However, the visual encoder of VLM detectors is frozen during text prompt, which prevents the updating of visual cues.

\subsection{Open Vocabulary Fine-Grained Perception Capability Evaluation}
As of now, most works evaluate the effectiveness of open-vocabulary detectors using established benchmarks like COCO \cite{lin2014microsoft} and LVIS \cite{gupta2019lvis}, which are designed for closed-set object detection. 
These benchmarks primarily focus on generic class labels and do not explore the capabilities of these detectors when the input text is more elaborate and includes fine-grained characteristics of the object. 
However, merely recognizing object categories is insufficient, particularly in complex environments that necessitate an understanding of detailed attributes of objects and their parts, such as color, texture, and material. 
Therefore, assessing the performance of open-vocabulary detectors at a fine-grained level has become particularly crucial. 
Recently, Bianchi et al. \cite{fgovd} introduced an evaluation benchmark named FG-OVD. 
This benchmark suite is constructed based on the PACO dataset \cite{paco} and employs a Large Language Model (LLM) to generate positive captions from semi-structured object descriptions, while negative captions of varying difficulty levels and distinct attributes are crafted through attribute substitution. 
Specifically, the benchmark provides a comprehensive evaluation across eight distinct scenarios, categorized into Difficulty-based and Attribute-based benchmarks. 
Difficulty-based benchmarks enable the assessment of detector performance across different difficulty levels by altering the hardness of negative captions. 
On the other hand, Attribute-based benchmarks allow for the precise selection of attribute types to facilitate the evaluation of detectors' capabilities in recognizing specific attributes.
We follow \cite{fgovd} to evaluate the performance of several models on this benchmark.
As shown in Table \ref{tab:fgovd}, we draw conclusion as following:

\textbf{(1)} In the difficulty-based evaluation, a significant performance drop was observed when transitioning from “trivial” to “easy” difficulty levels, indicating that detectors are more reliable in distinguishing category-level differences but less effective in handling fine-grained attribute variations. 
From “easy” to “medium,” the performance decline was relatively moderate, suggesting that replacing two attributes versus three attributes had limited impact on the models’ discrimination capabilities.  
However, from “medium” to “hard,” the performance degradation became more pronounced.  In scenarios where negative classes only had one attribute replaced, the high similarity between positive and negative classes posed a significant challenge for detectors. 
This further highlights the complexity of fine-grained detection tasks and their strong dependence on the number of attribute substitutions. Particularly in open-vocabulary detection, distinguishing subtle attribute differences remains a critical bottleneck. 
In terms of model comparison, OWL series \cite{minderer2022simple}\cite{minderer2024scaling} models performed the best overall, benefiting from large-scale image-text pretraining, which endowed them with strong capabilities for fine-grained attribute recognition. 
ViLD \cite{gu2022openvocabularyobjectdetectionvision} followed closely, leveraging knowledge distilled from CLIP \cite{radford2021learningtransferablevisualmodels} and thus indirectly learning from image-text data, demonstrating good generalization abilities.
However, at the “trivial” difficulty level, Detic \cite{zhou2022detecting} exhibited the best performance, yet its performance dropped significantly as task difficulty increased. This indicates that Detic’s training strategy is more focused on classification and detection tasks, lacking the ability to effectively capture attribute-level details, which limits its performance in fine-grained detection tasks.

\textbf{(2)} In the attribute-based evaluation, model performance varied significantly depending on the attribute type and the distribution of training data.  
Overall, models demonstrated superior detection performance for high-frequency attributes such as color and material, compared to low-frequency attributes like pattern and transparency.
This disparity is primarily due to the widespread presence of high-frequency attributes in image-text training data, which enables models to better learn and capture relevant features. 
Additionally, attributes like color and material are often more intuitive, further enhancing detectors’ performance on these attributes. 
In terms of model comparisons, OWL series \cite{minderer2022simple}\cite{minderer2024scaling} and Grounding-DINO \cite{liu2023grounding} consistently exhibited superior performance across most attribute categories. 
The OWL series, in particular, benefited from training strategies involving large-scale image-text data, allowing it to excel in high-frequency attributes while also demonstrating strong generalization capabilities for certain low-frequency attributes, such as pattern. 
Grounding-DINO achieved performance comparable to the OWL series across multiple attribute categories, supported by its diverse training data strategy, which provided the model with broader attribute feature learning capabilities.


\input{tables/fg_ovd}

\subsection{Dense Object Perception Capability}
Dense Object detection propose a challenging task in scenarios where objects are closely packed or overlapping. Unlike traditional object detection, dense objects are frequently adjacent or occluded, such as in traffic scenes, crowded public spaces, or aerial imagery. This makes detecting such objects more difficult and requires specialized techniques. Therefore, we evaluate dense object perception capability for object detection.

To evaluate the dense object detection capability, we select three datasets: CrowdHuman~\cite{shao2018crowdhuman}, OCHuman~\cite{Zhang_2019_CVPR}, and WiderPerson~\cite{zhang2019widerperson}.
CrowdHuman~\cite{shao2018crowdhuman} (15,000 training images and 4,370 validation images) contains approximately 22.6 pedestrians in average per image. 
OCHuman~\cite{Zhang_2019_CVPR} (2,500 training images and 2,231 validation images) emphasizes heavy occlusion. With an average 0.67 Max IoU for each person, OCHuman is the most complex and challenging dataset related to human. 
WiderPerson~\cite{zhang2019widerperson} (8,000 training images and 1,000 validation images) contains dense pedestrians with various kinds of occlusions. As shown in Tab.~\ref{tab:dense_det}, we draw conclusion as following:

\begin{figure*}[!t] 
  \centering
  \begin{minipage}[t]{0.32\textwidth} 
    \centering
    \includegraphics[width=\linewidth]{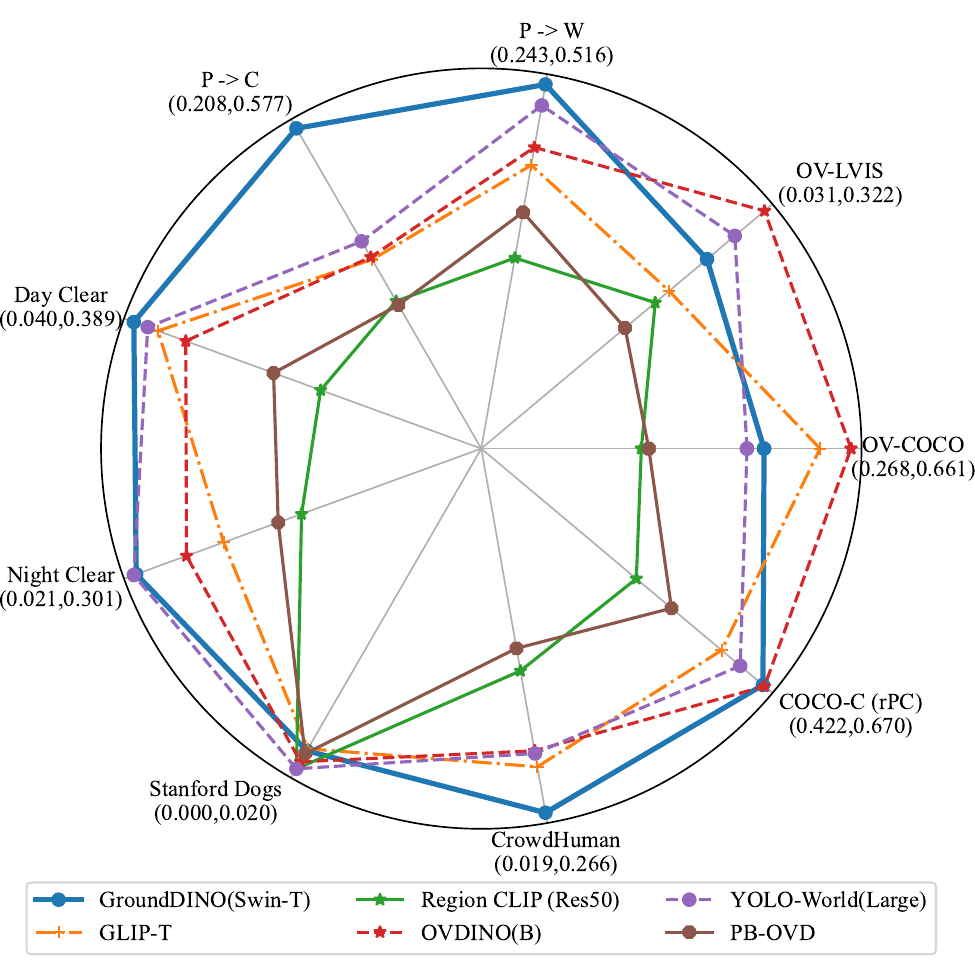}
  \end{minipage}
  \hfill 
  \begin{minipage}[t]{0.32\textwidth}
    \centering
    \includegraphics[width=\linewidth]{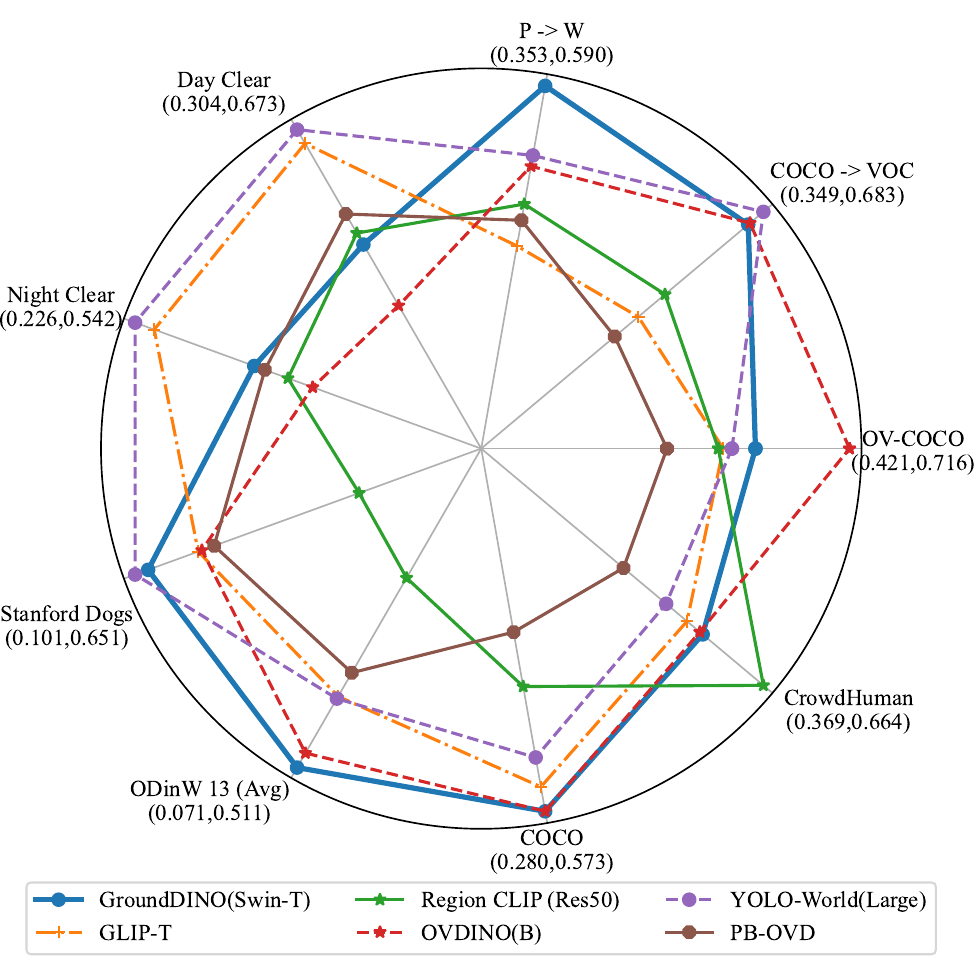}
  \end{minipage}
  \hfill
  \begin{minipage}[t]{0.32\textwidth}
    \centering
    \includegraphics[width=\linewidth]{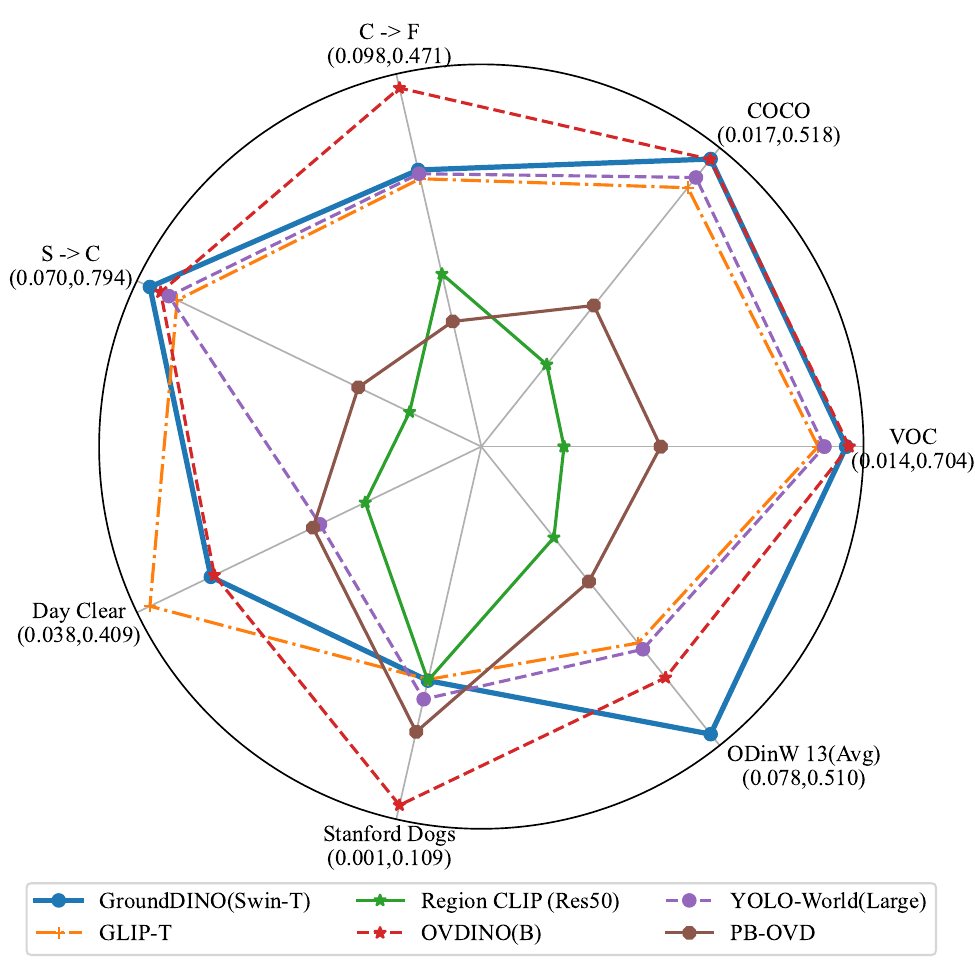}
  \end{minipage}
  
  \vspace{1.0 mm} 
  \caption{The performance comparison of different detectors on various tasks when making predictions using different paradigms. The three subgraphs respectively show the comparisons of predictions made using the Zero-Prediction paradigm, the Visual-Finetuning paradigm, and the Text-prompt paradigm.}
  \label{fig:radar_all}
\end{figure*}

(1) Zero-shot prediction results exhibit relatively lower performance. For example, Grounding DINO~\cite{liu2023grounding} achieves the best performance on both CrowdHuman and WiderPerson datasets, whereas other approaches such as RegionClip~\cite{zhong2022regionclip}, OVDINO (A)~\cite{wang2024ov}, and PB-OVD~\cite{gao2022open} perform significantly worse in zero prediction. During pretraining, methods such as Grounding DINO incorporate GoldG dataset, while RegionClip, OVDINO (A), and PB-OVD do not. It indicates that GoldG plays a critical role in enhancing the perception of dense objects.

(2) Visual fine-tuning consistently enhances model performance across all datasets. For instance, GLIP-T (C)~\cite{li2022grounded} gains 30.6\% AP on CrowdHuman and 45.5\% AP on WiderPerson after visual fine-tuning, demonstrating its effectiveness in dense object detection scenarios.

(3) The impact of text prompt fine-tuning varies across methods. While models such as Grounding DINO~\cite{liu2023grounding}, RegionCLIP~\cite{zhong2022regionclip}, and OV-DINO~\cite{wang2024ov} benefit considerably from this technique across datasets, others like YOLO-World~\cite{cheng2024yolo} and PB-OVD~\cite{gao2022open} exhibit relatively modest improvements. A key distinction lies in the pretraining strategy: YOLO-World and PB-OVD employ a frozen CLIP text encoder, whereas the others allow for end-to-end optimization. Given that CLIP primarily captures global semantic information, we hypothesize that freezing the text encoder limits the model's ability to effectively represent occluded or densely distributed objects, thereby constraining its capacity to leverage text prompt fine-tuning in dense scenes.

\input{tables/dense_det}

\subsection{Discussion}
The performance of detection methods across three granularity tuning approaches is summarized in Fig.~\ref{fig:radar_all}, we draw conclusion as following: 1) From an overall perspective, due to their pre-training on massive datasets, VLM models inherently possess strong generalization capabilities. Under \textit{zero prediction} settings, most VLMs perform well across all tasks except fine-grained tasks. With \textit{visual fine-tuning}, nearly all VLM models demonstrate good performance across all tasks, confirming that VLMs have the potential to serve as foundational models for various downstream applications. Under \textit{text prompt}, certain models (e.g. GroundingDINO and YOLO-World) achieve better performance than \textit{visual fine-tuning} in some domain adaptation scenarios at lower computational cost, which fully demonstrates the potential and effectiveness of \textit{text prompt} ways.
2) VLM models based on DETR frameworks with visual-text feature fusion (such as GroundingDINO and OVDINO) exhibit superior performance compared to traditional Faster R-CNN architecture VLMs (e.g., RegionCLIP) in most tasks. This clearly illustrates the critical importance of thorough visual-text feature fusion in VLM architectures, as it enables comprehensive information exchange between modalities and facilitates better visual-text feature alignment.
3) The model performance shows strong consistency across different tasks. For instance, GroundingDINO consistently achieves the best performance across all tasks, indicating that VLM models can essentially function as "feature extractors." More powerful VLMs can provide more robust features for various tasks, thereby maintaining leading performance across the board. The performance of individual models demonstrates strong consistency across different visual tasks.

\section{VLM-based segmentation task}
\label{sec:vlm-based segmentation task}




\input{tables/mess_full}

\subsection{Open-World Semantic Segmentation Evaluation}
\subsubsection{Open-Vocabulary Semantic Segmentation}
Open-vocabulary semantic segmentation (OVSS) aims to segment objects for arbitrary categories. Unlike conventional semantic segmentation methods that are limited to predefined categories, OVSS can handle a wider range of diverse real-world scenarios where object categories are dynamic and uncertain. By enabling the model to segment unknown categories, OVSS demonstrates greater adaptability and generalization in practical applications. Additionally, OVSS can address the challenges posed by the multi-grained and diverse nature of semantics, allowing for more flexible class name inputs, making it more user-friendly.

The task is commonly evaluated on ADE20K, PASCAL VOC and PASCAL-Context dataset. ADE20K has 20k training and 2k validation images, with two sets of categories: A-150 with 150 frequent classes and A-847 with 847 classes. PASCAL-Context contains 5k training and validation images, with 459 classes in the full version (PC-459) and the most frequent 59 classes in the PC-59 version. PASCAL VOC has 20 object classes and a background class, with 1.5k training and validation images. We adopt mean Intersection over Union (mIoU) as evaluation metric for all experiments.
\input{tables/ovss}

The experimental results for OVSS are presented in Tab.~\ref{tab:ovss}. The quantitative results reveal several key findings:

(1) Methods that utilize dense annotations on limited categories during training demonstrate higher performance compared to those using the supervision of large-scale image-text pairs (CC12M) and training-free methods. For instance, the Cat-Seg model, trained on 118,000 images across 171 categories, achieves a 13.1\% improvement in average mIoU compared to the TCL model, which is trained on 15 million image-text pairs. Additionally, Cat-Seg shows a 13.0\% improvement in average mIoU over the training-free method CLIPtrase.

(2) Methods that leverage region-level information achieve higher performance than those performing pixel-wise classification. For example, the two-stage method SAN achieves a 1\% improvement in average mIoU compared to ZegCLIP. This may be attributed to the smaller feature gaps in image-region-level representations compared to image-pixel-level representations.

(3) The training-free method CLIPtrase achieves relatively lower performance on the PC-59 dataset compared to ZegCLIP and DeOP, which were trained on COCO-Stuff-156 that includes many categories overlapping with PC-59. However, CLIPtrase outperforms ZegCLIP, DeOP, and the weak-supervised methods on the PC-459 dataset, which includes numerous novel categories. This difference in performance may indicate that the limited fine-tuning data could impact the generalization ability of CLIP and result in underwhelming performance on novel categories.

\subsubsection{Multi-domain Semantic Segmentation}

The experimental results presented in Tab.~\ref{tab:ovss} may not fully capture the behavior of open-vocabulary semantic segmentation models in real-world scenarios that involve more complex and domain-specific datasets, particularly in the field of medical sciences. In order to address this limitation, we conducted a comprehensive multi-domain evaluation on the MESS (Multi-domain Evaluation of Semantic Segmentation)~\cite{mess} benchmark. This benchmark is specifically designed to evaluate the real-world applicability of open-vocabulary models and consists of 22 datasets.

The MESS benchmark includes a diverse range of domain-specific datasets from various fields such as earth monitoring, medical sciences, engineering, agriculture, and biology. Additionally, it encompasses a wide variety of general domains, including driving scenes, maritime scenes, paintings, and body parts.   In Tab.~\ref{tab:mess}, we report the individual results for each domain-specific dataset, as well as the average scores for each field. This comprehensive evaluation provides insights into the performance of open-vocabulary semantic segmentation models across different domains and helps assess their generalization capabilities.

The quantitative results indicate the following:

(1)The open-vocabulary semantic segmentation models exhibit poor performance on various domain-specific datasets, exhibiting significantly lower mIoU compared to the top-performing supervised methods. For instance, the highest-performing OVSS method SED~\cite{xie2024sed},  lags behind the best supervised methods by a margin of 47.6 mIoU. This highlights the need for ongoing research and development efforts to address the challenges and improve the performance of the OVSS model in diverse domains.

(2) None of the methods consistently demonstrate high performance across datasets with different domains and categories. For instance, while the Cat-Seg method, which utilizes dense annotations during training, achieves the best overall performance in the general field, it exhibits relatively lower performance in the domains of engineering and medical sciences. In these specific domains, it even falls behind the training-free methods MaskCLIP and CLIPtrase. From this perspective, it is challenging to simply determine the superiority or inferiority of different supervision techniques for learning OVSS. Each method has its own unique advantages and applicability. Therefore, it is worthwhile to continue exploring and researching the impact of different supervision signals on OVSS.

(3) Pixel-based methods, including Cat-Seg and SED, exhibit superior generalization capabilities compared to their region-based counterparts, such as MAFT+. For instance, while SED and MAFT+ achieve similar performance in the OVSS benchmarks, with mIoU scores of 42.7 \% and 42.2 \% respectively, SED significantly outperforms MAFT+ by a substantial margin of 1.8\% in average mIoU, as detailed in Tab.~\ref{tab:mess}. This disparity may be attributed to the limited generalization of mask proposals within region-based methods, which are confined to training on a restricted dataset.

(4) The OVSS models particularly struggle with datasets that contain fine-grained categories, such as CUB-200. In such contexts, the models' performance is suboptimal, which may due to the pronounced inter-class similarity that complicates the distinction between categories. Consequently, further advancements are essential to refine these models and bolster their efficacy in fine-grained category segmentation tasks.

\subsection{Fine-grained Segmentation Evaluation}
\subsubsection{Fine-grained Semantic Segmentation Settings}
The goal of multi-granularity semantic segmentation is to segment different parts of the same semantic target in a more detailed manner. For example, for a crowd, multi-granularity semantic segmentation needs to segment areas such as human trunk and limbs. Our evaluation uses PASCAL-Part \cite{pascal-part} dataset and ADE20k-Part-234\cite{ade20k-part-234} dataset. 

PASCAL-Part is an annotated add-on set to PASCAL VOC 2010~\cite{Pascal_VOC}.Animals used for training and evaluation in PASCAL are highly metamorphic and occur at different scales and with different levels of occlusion. It can also be used as an ensemble for semantic part segmentation of the human body. Each image contains multiple human bodies with unrestricted poses and occlusions, of which 1716 are used for training and 1817 for testing. It can provide detailed pixel annotations for six parts of the human body: head, torso, upper/lower arms and upper/lower legs. 

ADE20k-Part \cite{ade20k-part} provides open-ended annotations of 847 objects and
1000+ parts, following the WordNet hierarchy. It covers a broad range of scenes, including indoor
spaces such as “bedrooms”, and outdoor spaces like “streetscapes”. However, the part annotations in
ADE20K are extremely sparse and incomplete (less than 15\% object instances have part annotations),
which poses significant challenges for both training models and evaluating their performance. To avoid this effect, we used the ADE20k-Part-234 \cite{ade20k-part-234} dataset, a subset of ADE20k-Part that consists of 44 objects
and 234 parts, providing a cleaner dataset for improved analysis and evaluation.

In our evaluation, we train our model on the seen categories of two datasets and evaluate it on both the seen and unseen categories to segment different numbers of the part class. This task aims to assess the model’s analogical reasoning ability, which is designed by selecting
novel objects that possess related parts to the base objects, rather than being completely irrelevant. To split the object classes in each dataset, we group the object classes into higher-level
categories (e.g. Animals, Vehicles) based on their shared attributes. Within each hyper-category, we
split the objects into seen and unseen classes. The unseen objects in the training set are set to the background. In this way, an unseen object part
class may be novel at the object level (e.g., “dog’s head” is an unseen class while “cat’s head” is a seen
class) or both at the object level and the part level (e.g., “bird’s beak”).

In terms of evaluation metrics, we first calculate the mean class-wise Intersection over Union (mIoU) on both base and novel classes. 

As shown in Table \ref{finegrained-seg}, we draw conclusion as following:

\textbf{(1)} In fine-grained segmentation, the methods that shared backbone, such as FC-CLIP \cite{NEURIPS2023_661caac7_FC_CLIP}, EBSeg \cite{shan2024open} still significantly outperform the two-backbone approaches such as MaskCLIP \cite{ding2022open}. This demonstrates that the visual features of a single backbone are well-equipped to handle downstream segmentation tasks. Additionally, stable visual features during the training process facilitate convergence.
\textbf{(2)} In fine-grained segmentation, for the seen categories, due to relatively sufficient training, the accuracy of mask proposal is guaranteed, and region-based methods such as FC-CLIP \cite{NEURIPS2023_661caac7_FC_CLIP,liu2024open} still have an advantage. However, for the unseen categories, pixel-based recognition methods such as SED \cite{xie2024sed} and CatSeg \cite{cho2024cat} demonstrate comparable or even better performance. Pixel-based recognition offers a more refined understanding of image regions and is more suitable for fine-grained local segmentation tasks.

\input{tables/fine_grained_segmentation}

\subsection{Few-Shot Evaluation}

\subsubsection{Few-Shot Semantic Segmentation Settings}

\input{tables/fewshot-semantic-segmentation-pascal}

Few-shot semantic segmentation aims to segment novel images of a specific category given only a limited number of annotated images of that category. Given that Vision-Language Models (VLMs) are pre-trained on extensive datasets and possess strong visual-text semantic alignment capabilities, it is worthwhile to evaluate the performance of VLM-based semantic segmentation models in the few-shot setting.

Commonly used datasets for evaluating few-shot semantic segmentation include PASCAL-$5^i$ \cite{shaban2017one}, COCO-$20^i$ \cite{Nguyen_2019_ICCV_coco_20i}, and FSS-1000 \cite{Li_2020_CVPR_fss1000}. PASCAL-$5^i$  is composed of the PASCAL VOC 2012 \cite{everingham2012pascal} dataset and the SBD \cite{SBD} dataset, encompassing a total of 20 categories.  PASCAL$5^i$ evenly divides the 20 categories into 4 subsets, with 5 categories in each subset.COCO-$20^i$ is based on the COCO2014 \cite{lin2014microsoft} dataset and contains 80 categories. COCO-$20^i$ also evenly divides all categories into 4 subsets, with 20 categories in each subset. 

During the evaluation, one of the four subsets can be selected evaluation, while the remaining three subsets serve as base classes that can be used for pre-training or fine-tuning. Common evaluation benchmarks are 1-shot and 5-shot, meaning that only one image or five images, respectively, are available for training per novel class. 

FSS-1000 comprises a total of 1000 categories, including numerous objects that are either unseen or unlabeled in commonly used datasets like PASCAL VOC \cite{everingham2015pascal} and COCO \cite{lin2014microsoft}. Examples of these novel categories include tiny everyday objects, merchandise, cartoon characters, and logos. The 1-shot and 5-shot settings are also commonly used in evaluation.

The commonly used metrics for Few-shot semantic segmentation are mIoU and FB-IoU. The calculation of mIoU is consistent with the previous evaluation method. FB-IoU focuses on the IoU calculation in the segmentation of foreground and background categories. Its basic principle still follows the general calculation logic of IoU, but it particularly pays attention to the accuracy of the foreground area in the segmentation result. FB-IoU regards the foreground class and the background class as one category respectively.

As shown in Table \ref{fewshot-seg1}, Table \ref{fewshot-seg2} and Table \ref{fewshot-3}, we draw conclusion that:
\textbf{(1)} Compared with the methods that freeze CLIP, such as FC-CLIP \cite{NEURIPS2023_661caac7_FC_CLIP, xie2024sed,liu2024open}, in the Few Shot tasks that require training, the Prompt-based method \cite{zhou2023zegclip,li2024cascade} without changing the forward process of the original backbone network can maintain the stability of features and thus achieve comparable or better performance. However, the Adapter-based method such as DeOP \cite{han2023open} modifies the features of the backbone network and may also introduce many dataset-related or task-related parameters, resulting in poorer performance in Few Shot tasks.
\textbf{(2)} Compared with the methods that freeze CLIP, such as MaskCLIP~\cite{ding2022open}, when using prompt to finetune the model \cite{zhou2023zegclip} or finetuning the model directly \cite{jiao2023learning}, the performance improvement will be more significant in few-shot tasks when the number of visible samples increases. This result is intuitive because more visible samples in training further optimize the model parameters, making them more suitable for task requirements. However, the method that freeze CLIP lacks such an adaptive adjustment mechanism for specific tasks and samples.
\textbf{(3)} The methods that share backbone \cite{NEURIPS2023_661caac7_FC_CLIP,shan2024open,liu2024open} are generally better than the methods that use two backbone \cite{xu2022simple,ding2022open} in Few-shot tasks. The methods that share backbone can utilize limited training data more effectively by sharing the backbone network. In the Few-shot task where data is scarce, the methods that use two backbones require more parameters for learning and adjustment due to its two independent backbone networks, which easily leads to overfitting. However, using one single backbone can share the feature extraction capability among different tasks or classes, reducing the total number of parameters and enhancing the generalization ability of the model. In addition, sharing backbone has a more unified representation during the training process. In the case where the number of training iterations is not large, maintaining a unified representation is more conducive to training stability and convergence.
\textbf{(4)} Methods based on region recognition \cite{NEURIPS2023_661caac7_FC_CLIP,liu2024open,shan2024open} outperform the methods based on pixel recognition \cite{zhou2023zegclip,xie2024sed}. In fine-grained segmentation, the segmentation quality is not good enough. Region-based methods can obtain more holistic features of the target. In contrast, pixel-based recognition is easily misled by noise and local information. In fine-grained segmentation, the fineness advantage of pixel-based recognition is not yet a performance bottleneck. Therefore, region-based recognition methods are often superior. However, it can be observed that SED \cite{xie2024sed} with a relatively large number of training steps can achieve performance similar to that of region-based recognition methods.

\subsection{Robust Segmentation Evaluation}

\input{tables/robust_seg}

The human visual system exhibits a remarkable robustness that current computer vision systems struggle to match. Humans can effortlessly interpret scenes despite various visual distortions such as snow, blur, pixelation, and even novel combinations of these corruptions while computer vision models often fail to maintain accuracy under similar conditions. Developing machine learning and computer vision systems that can achieve this level of robustness, withstanding a wide range of corruptions and variations in real-world environments, is a crucial challenge and an important goal for advancing artificial intelligence. Here we will test the robustness of the model by corrupting the image.

According to ~\cite{hendrycks2018benchmarking}, we use 15 different corruptions to adjust the images to simulate the image interference that may occur in actual situations. The 15 corruptions can be divided into 4 categories, Noise, Blur, Weather and Digital. Noise simulates the situation where the image is polluted by noise. Blur simulates the disturbance of the image when the object and the camera move. Weather simulates images in different weather conditions. Digital simulates the degradation of image quality that may be caused by transforming the image. Each corruption can be divided into five levels from weak to strong. We generate a corrupted image for each level of each corruption on five commonly used datasets of open vocabulary, and generate new datasets to test the robustness of the model. The test results are shown in the table.

The results are shown in Tab.~\ref{tab:seg_robust}. From the experimental results in the table, the following conclusions can be drawn:

(1) Overall, the existing models exhibit similar behavior when facing perturbations, with none demonstrating significantly superior robustness compared to the others. Specifically, on more challenging datasets such as ADE150K, ADE849, PC59, and PC459, there is a strong correlation between segmentation performance on corrupted data and the original performance, with a Pearson correlation coefficient of 0.91. This suggests that the model's performance under perturbation can be roughly predicted based on its original performance.

(2) Methods that adopt a frozen CLIP approach (e.g., EBSeg, FC-CLIP) demonstrate stronger robustness compared to those that fine-tune CLIP (e.g., MAFT-Plus). Specifically, under the cross-dataset setting, EBSeg and MAFT-Plus achieve comparable performance (EBSeg: 35.4\% vs. MAFT-Plus: 36.0\% mIoU), but EBSeg shows significantly less performance degradation under corruption scenarios, with its rPC improving by 3.46 over MAFT-Plus. Additionally, prompt-based methods (e.g., SAN and Cascade-CLIP), which do not alter the network's forward process, also maintain strong robustness. In contrast, adapter-based methods (e.g., DeOP) tend to perform the worst, as the adapter modules are typically lightweight and struggle to capture complex features, while also compromising the generalization ability of the original CLIP model.

(3) Pixel-based models (including Cascade-CLIP and Cat-Seg) exhibit stronger robustness compared to region-based approaches (such as DeOP). In region-based methods, noise can affect the generation of the entire region mask, whereas in pixel-based methods, noise only impacts pixel-level features, resulting in a more localized and limited effect. Specifically, although Cascade-CLIP and DeOP achieve similar performance in clean settings with average mIoUs of 35.4\% and 36\% respectively, Cascade-CLIP significantly outperforms DeOP under perturbation scenarios, with an average rPC that is 7.16\% higher.

\subsection{Zero-Shot Evaluation}

\input{tables/zeroshot_seg}

Zero-shot evaluation refers to training the model on the seen classes of the dataset and testing it on unseen categories. This task aims to segment classes that were not encountered during training, evaluating the model's zero-shot segmentation ability. It also assesses the model's capacity for semantic mapping and generalization from seen to unseen categories. We evaluate all the models on COCO Stuff and Pascal VOC 2012. Following the standard setup, we divide the COCO Stuff dataset into 156 seen classes and 15 unseen classes and the Pascal VOC 2012 dataset into 15 seen classes and 5 unseen classes. We provide the mean Intersection-over-Union (mIoU) on both seen and unseen classes and harmonic mean IoU (hIoU) among the seen classes and unseen classes.

The results are shown in Tab.~\ref{tab:zeroshotseg}. From the experimental results in the table, the following conclusions can be drawn: 

(1) CLIP-based models exhibit better generalization capabilities. For instance, ZegFormer, based on CLIP, attains mIoU scores of 33.2\% and 63.6\% on unseen categories. Through large-scale image-text pair pretraining, CLIP acquires an understanding of objects at a conceptual level rather than relying exclusively on predefined category labels. Consequently, even for unseen categories, CLIP-based models can leverage global contextual information for reasoning, thereby enhancing their performance on novel categories.

(2) The performance of pixel-based models tends to be slightly superior to that of region-based models. For example, CascadeCLIP achieves superior mIoU scores of 43.4\% (COCO-Stuff) and 83.1\% (Pascal VOC) on novel classes, outperforming  the best-performing region-based methods by 3.3\% and 6.5\%, respectively. The performance discrepancy may stem from region-based methods' reliance on mask proposal networks, where poor-quality proposals degrade final performance.

(3) Multi-level image features contribute to improving the model's generalization ability. CascadeCLIP, which exploits multi-level image features, attains mIoU scores of 43.4\% and 83.1\% on the novel classes of the COCO-Stuff and VOC datasets, respectively. One possible explanation is that CascadeCLIP harnesses multi-level visual features from CLIP's vision encoder while integrating distinct text embeddings to facilitate multi-level vision-language alignment. Since CLIP's pretraining objective prioritizes global image understanding, the final-layer image features may emphasize holistic image representations while discarding fine-grained details essential for segmentation tasks. By incorporating intermediate-layer image features, the model can more effectively capture object details, thereby enhancing its capability to segment unseen categories.

\subsection{Dense Object Segmentation Evaluation}

Segmenting highly-overlapping objects is challenging and the segmentation errors of dense objects account for a large proportion of the total segmentation errors  ~\cite{Ke_2021_CVPR}. Compared with conventional scenarios, the segmentation of dense objects places higher demands on the model, and the results in this scenario can better reflect the accuracy of model segmentation.

\subsubsection{Dense Object Segmentation Settings}

This scenario focuses on highly dense scenes where objects are closely adjacent to or even occluded by each other. This poses greater challenges to the segmentation model, thereby better reflecting its accuracy and understanding of the overall object. 

For dense object segmentation in natural scenes, we evaluate it on the COCO-OCC dataset ~\cite{Ke_2021_CVPR}. COCO-OCC dataset is a subset of the COCO validation set, containing 1,005 images. Each image in the COCO-OCC exhibits an overlap rate greater than 0.2 between the bounding boxes of its objects. 

Crowded scenes with dense human are a common type of dense object detection scenarios. Under this scenario, we utilize the OCHuman ~\cite{Zhang_2019_CVPR} and CIS ~\cite{10222709} datasets for evaluation. In these dataset, each human instance is significantly occluded by one or multiple other individuals, posing a significant challenge for instance segmentation. The OCHuman dataset comprises 8110 meticulously annotated human instances across 4731 images, with an average MaxIoU of 0.67 per Image. The CIS dataset encompasses the labeling of 463 images sourced from the CrowdHuman ~\cite{shao2018crowdhuman} validation set, each image depicting 3 to 10 humans experiencing occlusion, resulting in a comprehensive collection of 3,453 meticulously annotated human instances.

The experimental results for Robust Segmentation are presented in Tab.~\ref{tab:seg_dense}. FC-CLIP and ODISE models perform well in dense object segmentation. Based on the experimental results, we draw the following two conclusions.

(1) Diffusion-based mask generation demonstrates outstanding performance. During training, diffusion models leverage cross-attention between text embeddings and image features to learn rich semantic representations aligned with linguistic descriptions. These representations capture not only low-level visual cues but also high-level semantic concepts such as object categories, attributes, and relationships. This enables diffusion models to better understand image content and achieve more accurate target segmentation.

ODISE employs a conditional diffusion model for mask generation. Compared to another mask-based method, MAFT+, ODISE achieves higher segmentation accuracy across three datasets, despite not explicitly optimizing text features for visual alignment as MAFT+ does. Furthermore, on the more challenging COCO-OCC dataset, ODISE surpasses the best-performing FC-CLIP, showcasing strong competitiveness.

(2) Compared to fine-tuned CLIP methods, frozen CLIP approaches perform better in dense object segmentation. Fine-tuning on small-scale datasets may compromise CLIP's original ability to distinguish instance-level features. While MAFT+ achieves better results than FC-CLIP on standard open-vocabulary segmentation datasets such as ADE150 (31.1 vs. 33.6 mIoU) and slightly higher panoptic quality (PQ) in panoptic segmentation tasks (26.8 vs. 27.1), it falls short on dense object datasets.

This performance gap may be attributed to FC-CLIP’s strategy of incorporating a weighted fusion of features from the original CLIP visual encoder during mask category prediction, which helps preserve CLIP’s strong generalization ability in complex, crowded scenes.

\input{tables/seg_dense_obj}

\subsection{Small Object Segmentation Evaluation}

In segmentation tasks, small objects are often difficult for models to accurately recognize and segment due to their limited pixel representation, susceptibility to image noise, resolution constraints, and interference from larger objects. Small object semantic segmentation aims to assign semantic labels to each pixel of small-scale objects, such as cars, pedestrians, cyclists, traffic signs, and traffic lights. This task is particularly essential in domains like autonomous driving, where precise segmentation of small objects ensures safer navigation and better decision-making. Similarly, in UAV-based remote sensing, the dense prediction of small-scale entities such as buildings, vegetation, and roads supports applications in city planning and land-use monitoring. Pixel-level categorization of small objects like pedestrians and cars further aids traffic monitoring and crowd estimation, enabling more efficient and intelligent urban management.

In Vision-Language Models (VLMs), evaluating the capability for small object segmentation is key to understanding their potential in high-resolution, multi-scale perception, and fine-grained feature extraction. Performance analysis of small object segmentation provides insights into a model's overall visual capability on multiple levels, including detail processing, multi-scale perception, complex scene understanding, and generalization ability. This task holds significant value in various applications, such as autonomous driving, smart city planning, and advanced remote sensing analysis. Evaluating small object segmentation highlights the robustness, adaptability, and scalability of VLMs, further underscoring their importance in diverse visual tasks.

\subsubsection{Small Object Segmentation Settings}

\input{tables/seg_small_object_city_camvid}

\input{tables/seg_small_object_uavid_udd6}

For small object segmentation in natural scenes,we evaluate it on Cityscapes dataset~\cite{cordts2016cityscapes},CamVid dataset~\cite{BROSTOW200988-camvid},UAVid dataset~\cite{lyu2020uavid} and UDD6 dataset~\cite{chen2018large}.In the Cityscapes dataset, small objects mainly include pedestrians, cyclists, traffic signs, street signs, and so on. Compared with large objects such as vehicles and buildings, these small objects occupy relatively few pixels, especially when they are at a long distance and become even smaller. 

The CamVid dataset has a low image resolution (720×960). We define sign symbol, pedestrian, pole, and bicyclist as small-object classes. The remaining seven object classes are all denoted as large-object classes. Although the CamVid dataset is small in size, it provides unique challenges for small object segmentation tasks, especially because of its low resolution and complex lighting conditions.

The image samples in the UAVid dataset were captured by the UAV platform at approximately 50 metres with a 4096×2160 or 3840×2160 resolution. The dataset contains eight categories of objects and backgrounds in urban scenes (building, tree, background, road, low vegetation, static car, moving car, and human).Since the images were captured from a UAV perspective, all of these categories are defined as small objects.

The UDD6 dataset contains image samples captured by a UAV (DJI-Phantom 4) range from 60 to 100 m with a 4096X2160 or 4000x3000 resolution. Six categories of objects and backgrounds in urban scenes (other, facade, road, vegetation, vehicle, and roof) are contained in the UDD6 dataset.Since the images were captured from a UAV perspective, all of these categories are defined as small objects.

Therefore evaluating these datasets can reflect the capability of VLMs in small object segmentation.

\subsubsection{Analysis of experimental results}

Table~\ref{tab:seg_small_1} and Table~\ref{tab:seg_small_2} report the individual results of 12 different training methods and strategies for VLM segmentation models across two different scenarios and four different datasets, including the corresponding evaluation metrics (such as mIoU) and the performance of the best supervised method for small object segmentation tasks. This comprehensive evaluation provides deep insights into the performance of semantic segmentation models in the field of small object segmentation and helps assess their generalization capabilities.

The quantitative results indicate that the VLM segmentation models generally perform mediocrely on datasets with a large number of small objects, with mIoU significantly lower than that of the best supervised methods. For example, the best-performing model, SCAN~\cite{liu2024open}, lags behind the optimal supervised method by a margin of 24 in mIoU. This highlights the need for continued research and development to address the challenges and improve the performance of VLMs in small object segmentation.

The qualitative analysis reveals the following key observations:

(1) Freeze CLIP models outperform other fine-tuning strategies.Among supervised methods, models leveraging frozen CLIP features generally exhibit superior performance. This advantage stems from the preservation of CLIP's general and transferable vision-language representations, which are learned from large-scale datasets. These models maintain strong generalization and robustness, especially in small-object and open-domain scenarios. In contrast, fine-tuned models, although more adaptive to specific tasks, are prone to overfitting or catastrophic forgetting when trained on small or heterogeneous datasets. Prompt- or adapter-based approaches, despite being lightweight, often introduce task-specific biases that degrade the generalization ability. Representative examples include SCAN~\cite{liu2024open} achieving the best performance on CamVid, FC-CLIP~\cite{yu2024convolutions} on Cityscapes, and the MAFT series~\cite{jiao2023learning}~\cite{jiao2025collaborative} on UAVid and UDD6 datasets.

(2) Region-based methods outperform pixel-based methods in small object segmentation.Region-level approaches enhance the feature resolution and semantic consistency of small objects by focusing on localized regions and suppressing background noise. They are more effective in capturing boundary and detail information of small-scale targets. In contrast, pixel-level models, which rely heavily on local features and lack contextual understanding, are more susceptible to misclassification and omission of small objects. This performance gap is reflected in the superiority of region-based models such as SCAN~\cite{liu2024open}, FC-CLIP~\cite{yu2024convolutions}, and MAFTseries~\cite{jiao2023learning}~\cite{jiao2025collaborative} , compared to pixel-based models like Cascade-CLIP~\cite{li2024cascade} and Cat-Seg.

(3) Training-free models outperform text-supervised models.Training-free models benefit from retaining the rich vision-language priors of large-scale pretrained models (e.g., CLIP) by modifying the inference process without updating weights. This strategy enhances generalization to unseen categories and fine-grained structures. In contrast, text-supervised methods often suffer from noisy or imprecise labels, limiting the accuracy of feature learning. This issue is exacerbated in complex scenes or small object scenarios. For instance, although CLIPTrase~\cite{shao2024explore} outperforms SegCLIP~\cite{luo2023segclip}, both still lag significantly behind fully supervised approaches. 

\subsection{Discussion}
\begin{figure}[!t]
    \centering
    \includegraphics[width=0.90\columnwidth]{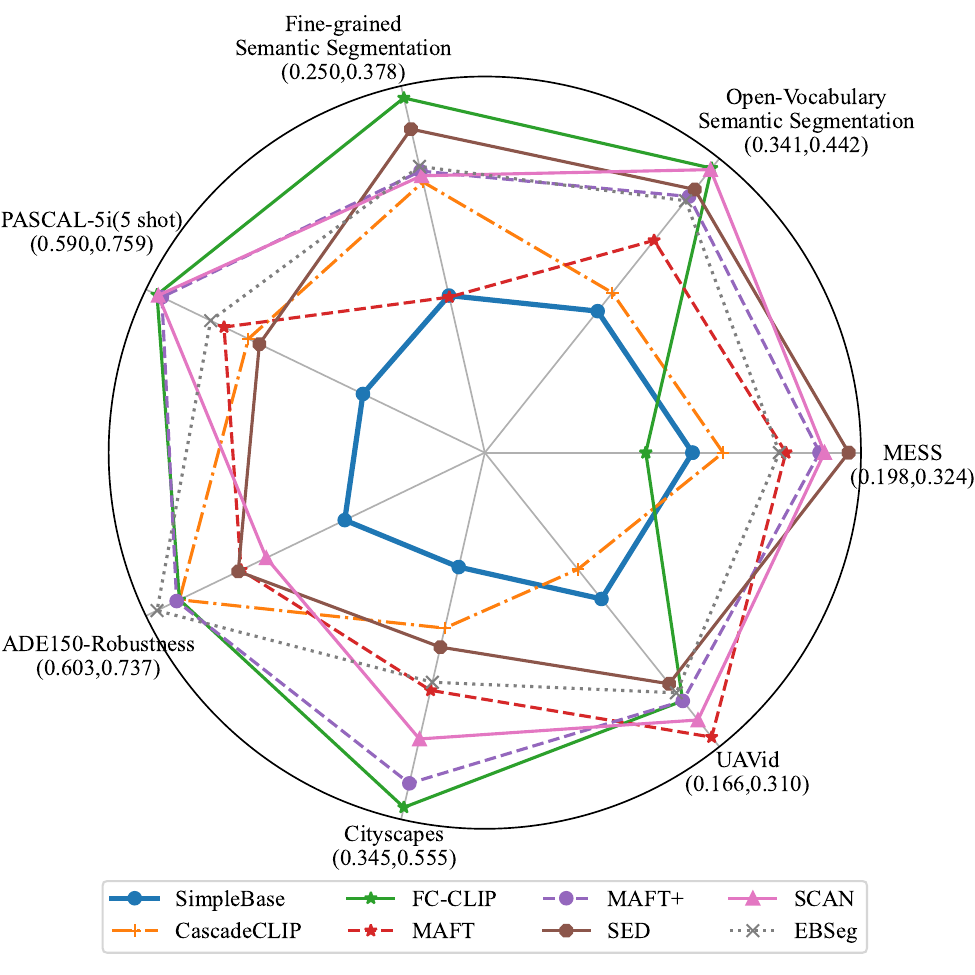}
    \caption{Comparisons of different segmentation methods on various tasks.}
    \label{fig:radar_text_prompt}
\end{figure}

   The performance of various segmentation methods on common benchmarks is summarized in Fig.~\ref{fig:radar_text_prompt}. While VLM-based approaches have demonstrated significant progress, their effectiveness in specialized domains, particularly in domain-specific tasks (MESS) and fine-grained semantic segmentation, remains limited and warrants further investigation. Our comprehensive evaluation reveals that no single model consistently achieves superior performance across all benchmarks, emphasizing the importance of systematic comparative analysis to better understand the strengths and limitations of different approaches.

\section{FUTURE DIRECTIONS}
\label{sec:future}


Visual-Language Models (VLMs) demonstrate effective utilization of vision-text paired data, exhibiting strong performance on downstream datasets under three granular fine-tuning paradigms: \textit{zero prediction}, \textit{visual fine-tuning}, and \textit{text prompt}. Segmentation VLMs achieve zero prediction without task-specific fine-tuning. This technology has achieved remarkable success due to its exceptional visual recognition capabilities. In this section, we discuss potential research challenges and future directions for VLMs across various visual recognition tasks.

\textbf{(1) Optimization of Pre-training Paradigms and Integration with Downstream Tasks} 
While VLM models currently rely heavily on large-scale pre-training data, designing pre-training paradigms that explicitly consider downstream tasks remains a critical research direction. Future work can explore incorporating task-specific multi-modal alignment mechanisms, cross-modal retrieval strategies, or constraints (e.g., spatial relationship modeling for dense prediction tasks) into the pre-training process~\cite{li2021align,lin2024vila}. Additionally, enhancing the model's adaptability to small-sample tasks or improving data utilization efficiency (e.g., pseudo-label generation, data augmentation) during pre-training deserves further investigation~\cite{zhou2022conditional,kim2024vlm}. Furthermore, integrating domain-specific requirements of downstream tasks (e.g., geometric constraints or semantic associations for detection or segmentation) into the pre-training process could lead to more efficient model training and better task transfer~\cite{huo2023geovln,ma2022open,rao2023retrieval,gan2022vision}.

\textbf{(2) Optimization of Network Architectures and Innovation in Visual-Text Feature Fusion} 
Although compact network structures and deep integration of visual and text features have proven effective in improving VLM performance, achieving early-stage fusion of visual and text features in the backbone remains an open challenge~\cite{singh2022flava,zhai2022lit}. Future research can focus on designing lightweight, efficient architectures that enable early interaction between visual and text features, enhancing the model's global semantic understanding. Additionally, developing innovative cross-modal fusion mechanisms that leverage multi-scale visual features and multi-level semantic information from text to improve spatial awareness for dense prediction tasks (e.g., detection, segmentation) is worth exploring~\cite{min2022grounding,qiu2024mining,liu2024multi,jiao2019geometry}. Furthermore, achieving model compression through pruning or knowledge distillation while maintaining performance could be an important direction for practical applications~\cite{fang2021compressing,wang2020minivlm,wang2022efficientvlm}.

\textbf{(3) Efficient Utilization of Pre-trained Visual Foundation Models to Enhance VLM Performance} 
Current VLM designs often rely on complex operations for region-level feature extraction and alignment, while pre-trained visual foundation models (e.g., CLIP or DINOv2) already demonstrate strong capabilities in visual-text alignment. Leveraging these models to further enhance VLM performance is a valuable research direction. Future work can explore transferring learning or knowledge distillation methods to transfer semantic understanding from visual foundation models to VLMs, improving their performance in dense prediction tasks~\cite{zhang2021tip,fufrozen,ramos2023smallcap,deng2024simclip}. Additionally, combining the feature extraction capabilities of visual foundation models with VLM's multi-modal alignment capabilities within a multi-task learning framework could lead to more efficient model optimization~\cite{kuo2022f,wu2024vila,chen2024frozenseg,zhao2023prompting,awais2025foundation}. For example, simplifying complex region-level feature extraction processes in VLM while enhancing fine-grained visual understanding through visual foundation models is a promising avenue for investigation.

\textbf{(4) Optimizing the trade-off between segmentation accuracy and computational efficiency.}
Current OVSS methods exhibit a dichotomy in their approach: two-stage paradigms~\cite{jiao2025collaborative_maft+, NEURIPS2023_661caac7_FC_CLIP} achieve high-quality segmentation masks through large input sizes (1024×1024), but at the expense of substantial computational resources. Conversely, one-stage methods~\cite{cho2024cat}, constrained by pretrained model limitations, operate with smaller input sizes (384×384), resulting in compromised edge accuracy despite their superior inference speed. A crucial future research direction lies in developing innovative architectures or optimization strategies that can bridge this gap, potentially through adaptive resolution mechanisms, efficient feature extraction techniques, or hybrid approaches that intelligently balance computational demands with segmentation precision.

\textbf{(5) Knowledge Distillation for Efficient Vision-Language Segmentation} Extensive experimental results demonstrate that Vision-Language Models (VLMs) with large-scale backbones (e.g., ViT-L) achieve superior segmentation performance compared to their smaller counterparts (e.g., ViT-B)~\cite{xie2024sed, cho2024cat, jiao2025collaborative_maft+, liu2024open}. However, this performance gain comes at the cost of significantly increased computational complexity and inference latency, making these large models impractical for real-world deployment scenarios. To address this limitation, we propose leveraging knowledge distillation techniques to transfer the enhanced representational capabilities of large VLMs to more compact architectures. This approach enables an optimal balance between model accuracy and computational efficiency, facilitating the development of practical segmentation systems for real-world applications.

 \textbf{(6) More versatile task heads or more efficient training paradigms}. Given the rapid evolution of Vision-Language Models (VLMs)~\cite{radford2021learningtransferablevisualmodels, li2021align}, the process of retraining or adapting distinct VLM models entails substantial computational resources and increases the complexity of deploying these models in downstream applications. The creation of universally compatible task heads or the establishment of efficient adaptation mechanisms~\cite{han2024parameter} would greatly enhance the practicality and scalability of VLMs in downstream tasks, thereby promoting their widespread deployment in real-world applications.

\section{Conclusion}
\label{sec:conclusion}


In this study, we present a comprehensive empirical evaluation of Vision-Language Model (VLM)-based methodologies, spanning a diverse array of detection and segmentation tasks to rigorously assess their capabilities in visual perception. Through systematic experimentation and multifaceted analysis, we benchmark the performance of state-of-the-art VLM approaches across eight pivotal dimensions, including but not limited to: open-vocabulary learning, cross-domain generalization, robustness to distribution shifts, and dense object recognition.
Our evaluation framework incorporates both established benchmarks and challenging real-world scenarios, encompassing over 50 distinct datasets to ensure thorough and representative assessment. For each evaluation dimension, we derive at least three evidence-based conclusions through: (1) quantitative analysis of experimental results, (2) comparative study of competing methods, and (3) in-depth examination of failure patterns and success criteria.

This large-scale investigation yields three primary contributions: First, it establishes comprehensive empirical baselines across the VLM landscape. Second, it reveals previously undocumented limitations and emergent capabilities through systematic analysis. Third, it provides promising yet underexplored research directions to accelerate progress in this dynamic field.


\bibliographystyle{IEEEtran}
\bibliography{ref.bib}

\end{document}

%% file: tables/ovd_first_model_summary.tex
\begin{table*}[htbp]
    \centering
    \caption{Summary of Large-scale Pretraining Methods for Open Vocabulary Object Detection Models. Abbreviations: O365 (Objects365 \cite{shao2019objects365}), OI (OpenImages \cite{OpenImages}), VG (Visual Genome \cite{krishna2017visual}), CC3M (Conceptual Captions \cite{sharma2018conceptual}), CC12M (Conceptual 12M \cite{changpinyo2021conceptual}), YFCC1M (a subset of YFCC100M \cite{thomee2016yfcc100m}), RefC (RefCOCO, RefCOCO+ and RefCOCOg \cite{yu2016modeling})}
    \scriptsize
    \label{tab:first_method}
    \begin{tabular}{m{2.9cm}|m{2cm}|m{1.2cm}|m{2.8cm}|m{5.5cm}|m{1cm}}
        \toprule
        Method & Image Encoder & Text Encoder & Training Datasets & Contribution & Published\\
        \midrule \midrule
        GLIP\cite{li2022grounded} \href{https://github.com/microsoft/GLIP}{[code]} & Swin Transformer & BERT & O365, OI, VG, ImageNetBoxes \cite{krizhevsky2012imagenet}, GoldG \cite{li2022grounded}, CC12M, SBU Caption & Propose a unified framework for object detection and phrase grounding in pre-training, enabling deep fusion between image and language encoders.  & CVPR'22\\
        RegionCLIP\cite{zhong2022regionclip} \href{https://github.com/microsoft/RegionCLIP}{[code]} & CLIP-ResNet50 & CLIP-text & CC3M & Extend CLIP to learn region-level visual representations for fine-grained alignment between image regions and textual concepts. & CVPR'22 \\
        PB-OVD\cite{gao2022open} \href{https://github.com/salesforce/PB-OVD}{[code]} & ResNet50 & CLIP-text & COCO Caption \cite{chen2015microsoft}, VG, SBU Caption & Propose to generate pseudo labels from large-scale image-text pairs using vision-language models for training object detectors. & ECCV'22 \\
        DetCLIP\cite{yao2022detclip} & Swin Transformer & FILIP-text & O365, GoldG, YFCC1M) & Propose a paralleled visual-concept pre-training method for open-world object detection that leverages a concept dictionary to enhance knowledge representation. & NeurIPS'22 \\
        OWL-ViT\cite{minderer2022simple} \href{https://github.com/google-research/scenic/tree/main/scenic/projects/owl_vit}{[code]} & Modified CLIP-ViT & Transformer & O365, VG & Perform image-text pretraining and end-to-end detection fine-tuning using the modified Vision Transformer for open-vocabulary object detection. & ECCV'22 \\
        OWLv2\cite{minderer2024scaling} \href{https://github.com/google-research/scenic/tree/main/scenic/projects/owl_vit}{[code]} & Modified CLIP-ViT & Transformer & WebLI \cite{chen2022pali} & Scale up detection data with self-training, which uses an existing detector to generate pseudo-box annotations on image-text pairs.  & NeurIPS'23 \\
        DetCLIPv2\cite{yao2023detclipv2} & Swin Transformer & FILIP-text & O365, GoldG, CC3M, CC12M & Propose an efficient and scalable framework for open-vocabulary object detection that learns fine-grained word-region alignment. & CVPR'23 \\
        DetCLIPv3\cite{yao2024detclipv3} & Swin Transformer & FILIP-text & O365, V3Det \cite{wang2023v3det}, GoldG, GranuCap50M \cite{yao2024detclipv3} & Propose integrating a caption generation head and utilizing an auto-annotation pipeline to provide multi-granular object labels. & CVPR'24 \\
        Grounding DINO\cite{liu2023grounding} \href{https://github.com/IDEA-Research/GroundingDINO}{[code]} & Swin Transformer & BERT & O365, OpenImage, GoldG, Cap4M \cite{li2022grounded}, COCO, RefC & Integrate a Transformer-based detector with grounded pre-training through a tight fusion of language and vision. & ECCV'24 \\
        YOLO-World\cite{cheng2024yolo} \href{https://github.com/AILab-CVC/YOLO-World}{[code]} & CSPDarkNet & CLIP-text & O365, GoldG, CC3M & Propose an enhanced YOLO detector with open-vocabulary capabilities through vision-language modeling and pre-training on large-scale datasets.& CVPR'24 \\
        OV-DINO \cite{wang2024ov} \href{https://github.com/wanghao9610/OV-DINO}{[code]} & Swin Transformer & BERT & O365, GoldG, CC1M & Propose a unified method that integrates diverse data for end-to-end pre-training, and enhances region-level cross-modality fusion and alignment. & Arxiv'24 \\
        \bottomrule
    \end{tabular}
\end{table*}

%% file: tables/ovd_second_model_summary.tex
\begin{table*}[htbp]
    \centering
    \caption{Summary of Learning Strategy Based Methods for Open Vocabulary Object Detection Models. The numbers in the 'Training Datasets' column indicate different experimental settings.}
    \label{tab:second_method}
    \scriptsize
    \renewcommand{\arraystretch}{1.5}
    \begin{tabular}{>{\raggedright}m{2.7cm}|m{1.8cm}|m{1.3cm}|m{2.2cm}|m{6.5cm}|m{1cm}}
    \toprule
        Method & Image Encoder & Text Encoder & Datasets & Contribution & Published \\
        \midrule \midrule
        Detic~\cite{zhou2022detecting} \href{https://github.com/facebookresearch/Detic}{[code]} & ResNet50 & CLIP-text & LVIS-base, IN-L, CC & Propose training detector classifiers on image classification data to enable detection of a wide range of concepts. & ECCV'22 \\
        DetPro~\cite{du2022learning} \href{https://github.com/dyabel/detpro}{[code]} & ResNet50 & CLIP-text & LVIS-base & Propose a novel method for learning continuous prompt representations for open-vocabulary object detection. & CVPR'22 \\
        OV-DETR~\cite{zang2022open} \href{https://github.com/yuhangzang/OV-DETR}{[code]} & ResNet50 & CLIP-text & \makecell[l]{1. COCO-base \\ 2. LVIS-base} & Propose an OV DETR-based detector that performs object detection using class names via CLIP-based binary matching. & ECCV'22 \\
        ViLD\cite{gu2021open} \href{https://github.com/tensorflow/tpu/tree/master/models/official/detection/projects/vild}{[code]} & ResNet50 & CLIP-text & \makecell[l]{1. COCO-base \\ 2. LVIS-base} & Propose to distill knowledge from a pre-trained VLM into a object detector by aligning the student's and teacher's embedding. & ICLR'22 \\
        HierKD\cite{ma2022open} \href{https://github.com/mengqiDyangge/HierKD}{[code]} & ResNet50 & CLIP-text & COCO-base & Propose a hierarchical visual-language knowledge distillation method for open-vocabulary one-stage detectors, combining global and instance-level distillation. &  CVPR'22 \\
        VL-PLM \cite{zhao2022exploiting} \href{https://github.com/xiaofeng94/VL-PLM}{[code]}&  ResNet50 & CLIP-text & \makecell[l]{1. COCO-base \\ 2. LVIS-base} & Propose the VL-PLM framework that leverages VLM to generate pseudo labels for novel categories to train open-vocabulary detector. & ECCV'22 \\
        PromptDet\cite{feng2022promptdet} \href{https://github.com/fcjian/PromptDet}{[code]} & ResNet50 & CLIP-text & LVIS-base, LAION-novel & Propose regional prompt learning to align textual embeddings with visual object features and a self-training framework to scale detection without manual annotations. & ECCV'22 \\
        VLDet\cite{lin2022learning} \href{https://github.com/clin1223/VLDet}{[code]} & ResNet50 & CLIP-text & \makecell[l]{1. COCO-base, \\COCO Caption \\ 2. LVIS-base, \\CC3M} & Learn from image-text pairs by formulating object-language alignment as a set matching problem between image region features and word embeddings. & ICLR'23 \\
        BARON\cite{wu2023aligning} \href{https://github.com/wusize/ovdet}{[code]} & ResNet50 & CLIP-text & \makecell[l]{1. COCO-base \\ 2. LVIS-base} & Propose a method to enhance open-vocabulary object detection by aligning the embedding of a bag of regions. & CVPR'23 \\
        CoDet\cite{ma2024codet} \href{https://github.com/CVMI-Lab/CoDet}{[code]} & ResNet50 & CLIP-text & \makecell[l]{1. COCO-base,\\ COCO Caption \\ 2. LVIS-base, CC3M} & Reformulate region-word alignment as a co-occurring object discovery problem, leveraging visual similarities to discover and align objects with shared concepts. & NeurIPS'23 \\
        CORA\cite{wu2023cora} \href{https://github.com/tgxs002/CORA}{[code]} & CLIP-ResNet50 & CLIP-text & \makecell[l]{1. COCO-base \\ 2. LVIS-base} & Propose a DETR-style framework that adapts CLIP using Region prompting to address the whole-to-region distribution gap and Anchor pre-matching for improved object localization. & CVPR'23 \\
        DK-DETR\cite{li2023distilling} \href{https://github.com/hikvision-research/opera/tree/main/configs/dk-detr}{[code]} & ResNet50 & CLIP-text & \makecell[l]{1. COCO-base \\ 2. LVIS-base} & Propose a framework that distills semantic and relational knowledge from VLM into a DETR-like detector & ICCV'23 \\
        DST-Det\cite{xu2023dst} \href{https://github.com/xushilin1/dst-det}{[code]} & ResNet50 & CLIP-text & \makecell[l]{1. COCO-base \\ 2. LVIS-base \\ 3. V3-Det-base} & Propose a strategy that leverages the zero-shot classification ability of pre-trained VLM to generate pseudo-labels for novel classes. & Arxiv'23 \\
        EdaDet\cite{shi2023edadet} & ResNet50 & CLIP-text & \makecell[l]{1. COCO-base \\ 2. LVIS-base} & Propose Early Dense Alignment (EDA) to improve base-to-novel generalization by learning dense-level alignment with object-level supervision. & ICCV'23 \\
        F-VLM\cite{kuo2022f} \href{https://github.com/google-research/google-research/tree/master/fvlm}{[code]} & CLIP-ResNet50 & CLIP-text & \makecell[l]{1. COCO-base \\ 2. LVIS-base} & Simplify training by using a frozen vision-language model and fine-tuning only the detector head. & ICLR'23 \\
        MM-OVOD\cite{kaul2023multi} \href{https://github.com/prannaykaul/mm-ovod}{[code]} & ResNet50 & CLIP-text & LVIS-base, IN-L  & Propose generating text-based classifiers with a llm, employing a visual aggregator for image exemplars, and fusing both to create a multi-modal classifier. & ICML'23 \\
        OADP\cite{wang2023object} \href{https://github.com/LutingWang/OADP}{[code]} & ResNet50 & CLIP-text & \makecell[l]{1. COCO-base \\ 2. LVIS-base} & Propose an Object-Aware Knowledge Extraction module for precise object knowledge extraction and a Distillation Pyramid mechanism for comprehensive global and block distillation. & CVPR'23 \\
        Prompt-OVD\cite{song2023prompt} & ResNet50 & CLIP-text & \makecell[l]{1. COCO-base \\ 2. LVIS-base} & Propose to use CLIP class embeddings as prompts, along with RoI-based masked attention and RoI pruning to enhance detection performance with minimal computational cost. & Arxiv'23 \\
        RO-ViT\cite{kim2023region} \href{https://github.com/google-research/google-research/tree/master/fvlm/rovit}{[code]} &  ViT & Transformer & \makecell[l]{Pretraining: ALIGN \\ 1. COCO-base \\ 2. LVIS-base} & Propose randomly cropping and resizing regions of positional embeddings to align with region-level detection, replacing softmax cross entropy with focal loss. & CVPR'23 \\
        SAS-Det\cite{zhao2024taming} \href{https://github.com/xiaofeng94/SAS-Det}{[code]} & CLIP-ResNet50 & CLIP-text & \makecell[l]{1. COCO-base \\ 2. LVIS-base} & Propose a split-and-fusion head to separate open and closed branches for complementary learning and reduces noisy supervision. & CVPR'24 \\
        CLIPSELF\cite{wu2023clipself} \href{https://github.com/wusize/CLIPSelf}{[code]} &  CLIP-ViT & CLIP-text & \makecell[l]{1. COCO base \\ 2. LVIS base} & Adapt CLIP ViT's image-level recognition to local regions by self-distilling region representations from its dense feature map. & ICLR'24 \\
        LP-OVOD\cite{pham2024lp} \href{https://github.com/VinAIResearch/LP-OVOD}{[code]} & ResNet50 & CLIP-text & \makecell[l]{1. COCO-base \\ 2. LVIS-base} & Discard low-quality boxes by training a sigmoid linear classifier on pseudo labels retrieved from the top relevant region proposals to the novel text. & WACV'24 \\
        LAMI-DETR \cite{du2025lami} \href{https://github.com/eternaldolphin/LaMI-DETR}{[code]} &  CLIP-ConvNext & CLIP-text & LVIS base & Propose a method to leverage the relationships between visual concepts, sample negative categories during training, and resolve confusing categories during inference. & ECCV'24 \\
        
        \bottomrule
    \end{tabular}
\end{table*}

%% file: tables/seg_model_summary.tex
\begin{table*}[htbp]
    \centering
    \caption{Summary of Open Vocabulary Object Segmentation Models. FT denotes full-parameter fine-tuning of CLIP.; PA denotes fine-tuning CLIP using prompts or adapters; Pix denotes pixel-based; Two denotes two backbone; Sre denotes shared backbone; TS denotes Text-supervised; TF denotes training-free.}
    \label{tab:seg_methods}
    \scriptsize
    \resizebox{1.0\textwidth}{!}{%
    \begin{tabular}{l|>{\centering\arraybackslash}p{1cm}|>{\raggedright}p{3cm}|p{2.5cm}|p{2cm}|p{2.5cm}|c}
        \toprule
        Method & Category & Additional Segmentor & Image Encoder & Text Encoder & Training Datasets & Published \\
        \midrule
        \midrule
        LSeg~\cite{li2022language}\href{https://github.com/isl-org/lang-seg}{[code]}&FT\&Pix  & - & ViT-L/16 & CLIP VIT-B/32 & PASCAL-5i/COCO-20i & ICLR'22 \\
        Cat-Seg~\cite{cho2024cat}\href{https://github.com/KU-CVLAB/CAT-Seg}{[code]} & FT\&Pix & - & CLIP VIT-B/L + Swin Transformer & CLIP VIT-B/L & COCO-Stuff & CVPR'24 \\
        SAN~\cite{xu2023side}\href{https://github.com/MendelXu/SAN}{[code]} & PA\&Two& - & CLIP VIT-B/L & CLIP VIT-B/L & COCO-Stuff & CVPR'23 \\
        Simple Baseline~\cite{xu2022simple}\href{https://github.com/MendelXu/zsseg.baseline}{[code]} & Frz\&Two & Maskformer & CLIP VIT-B & CLIP VIT-B & COCO-Stuff & ECCV'22 \\
        MaskCLIP~\cite{ding2022open}\href{https://github.com/mlpc-ucsd/MaskCLIP}{[code]} & Frz\&Two & MaskRCNN/ Mask2former & CLIP VIT-L/336 & CLIP VIT-L/336 & COCO-133 & ICML'23 \\
        DeOP~\cite{han2023open}\href{https://github.com/CongHan0808/DeOP}{[code]} &PA\&Two & Resnet101-MaskFormer & CLIP VIT-B/16 & CLIP VIT-B/16 & COCO-Stuff-156 & ICCV'23 \\
        FC-CLIP~\cite{yu2024convolutions}\href{https://github.com/bytedance/fc-clip}{[code]} &Frz\&Sre & Mask2former & ConvNeXt-Large CLIP & ConvNeXt-Large CLIP & COCO Panoptic  & NeurIPS'23 \\
        MAFT~\cite{jiao2023learning}\href{https://github.com/jiaosiyu1999/MAFT}{[code]}&FT\&Two & Maskformer & CLIP-B/16 & CLIP-B/16 & COCO-Stuff & NeurIPS'23 \\
        SED~\cite{xie2024sed}\href{https://github.com/xb534/SED}{[code]} & FT\&Pix& - & ConvNeXt-L & CLIP-B/16 & COCO-Stuff & CVPR'24 \\
        SCAN~\cite{liu2024open}\href{https://github.com/yongliu20/SCAN}{[code]} & Frz\&Two & Swin-Mask2former & CLIP VIT-B/L & CLIP VIT-B/L & COCO-Stuff & CVPR'24 \\
        EBSeg~\cite{shan2024open}\href{https://github.com/slonetime/EBSeg}{[code]} &Frz\&Sre & SAM & CLIP VIT-B/L & CLIP VIT-B/L & COCO-Stuff & CVPR'24 \\
        Zegformer~\cite{ding2022decoupling}\href{https://github.com/dingjiansw101/ZegFormer}{[code]} &Frz\&Two & Resnet50-FPN & CLIP VIT-B & CLIP VIT-B & COCO-Stuff & CVPR'22 \\
        ZegCLIP~\cite{zhou2023zegclip}\href{https://github.com/ZiqinZhou66/ZegCLIP}{[code]} &PA\&Pix &- & CLIP VIT-B & CLIP VIT-B & COCO-Stuff & CVPR'23 \\
        Pading~\cite{he2023primitive}\href{https://github.com/heshuting555/PADing}{[code]} &Frz\&Two & Resnet50-Mask2former & None & CLIP VIT-B & COCO-Stuff & CVPR'23 \\
        Cascade-CLIP~\cite{li2024cascade}\href{https://github.com/HVision-NKU/Cascade-CLIP}{[code]} &PA\&Pix & SegViT & CLIP VIT-B/16 & CLIP VIT-B/16 & COCO-Stuff & ICML'24 \\
        SegCLIP~\cite{luo2023segclip}\href{https://github.com/ArrowLuo/SegCLIP}{[code]} &TS  & - & CLIP VIT-B/16 & CLIP VIT-B/16 & CC3M+COCO Caption & ICML'24 \\
        TCL~\cite{cha2023learning}\href{https://github.com/kakaobrain/tcl}{[code]} & TS & - & CLIP VIT-B/16 & CLIP VIT-B/16 & CC3M+CC12M & CVPR'23 \\
        MaskCLIP~\cite{zhou2022extract}\href{https://github.com/chongzhou96/MaskCLIP}{[code]} &TF & - & CLIP VIT-B/16 & CLIP VIT-B/16 & - & ECCV'22 \\
        CLIPTrase~\cite{shao2024explore}\href{https://github.com/leaves162/CLIPtrase}{[code]} & TF& - & CLIP VIT-B/16 & CLIP VIT-B/16 & - & CVPR'24 \\
        \bottomrule
    \end{tabular}
    }
\end{table*}

%% file: tables/det_general_closed.tex
\begin{table*}[htbp]

\centering
\caption{General closed detection performance (\%) on VOC~\cite{Pascal_VOC}, COCO~\cite{lin2014microsoft}, and LVIS~\cite{gupta2019lvis}.}
\label{tab:det_general_closed}
\resizebox{1.0\textwidth}{!}{%
    \begin{tabular}{l|l|ccc|ccc|cccc|l}
    \toprule
    \multirow{2}{*}{Method} & \multirow{2}{*}{Finetuning Ways} & \multicolumn{3}{c|}{VOC} & \multicolumn{3}{c|}{COCO} & \multicolumn{4}{c|}{LVIS}     & \multirow{2}{*}{Published} \\
\cline{3-12}          &       & AP$_{50}$ & AP$_{75}$ & AP    & AP$_{50}$ & AP$_{75}$ & AP    & AP$_{r}$ & AP$_{c}$ & AP$_{f}$ & AP    &  \\
    \midrule
    \midrule
    Faster R-CNN ~\cite{ren2016faster} &  \multirow{4}[2]{*}{Finetuning}     &    74.6   &    47.2   &  44.3 &    55.2   &    37.2   &    34.8&6.2&15.5&24.2&17.3& NeurIPS’15 \\
    Dynamic Head ~\cite{li2022grounded} &       &   85.7    &    70.0   &    64.3   &  75.3     &   61.5    &   56.2    &    0.0   &    1.7   &   18.5    &   11.0    & CVPR’21 \\
    YOLO-v8 ~\cite{yolov8_ultralytics} &       &       83.0&       70.3&       65.0&       62.8&       50.7&       46.5&       8.4&       21.3&       32.4&       28.4& Online'23 \\
    DINO (Swin-L) ~\cite{zhang2022dino} &  &   88.1    &   74.6    &    68.9   &   75.9    &   62.9    &    57.4   &  4.1     &26.1       &41.1       &  28.2     & ICLR'2023 \\
    \hline
    PB-OVD ~\cite{gao2022open} &   \multirow{7}[2]{*}{Visual Finetuning}    &   62.0    &   34.2    &   34.7    &    47.2    &   28.7   &   28.0 &   1.1    &    4.6   &   14.7    &   8.0    & ECCV’22 \\
    GLIP-T (A) ~\cite{li2022grounded} &       &    90.4   &   78.2    &    70.9   &    71.5   &    58.5   &    53.3   &    27.1   &    37.2   &     45.3  &   38.7    & CVPR’22 \\
    GLIP-T (B) ~\cite{li2022grounded} &       &    90.5   &    78.8   &    71.5   &    72.4   &    59.2   &    54.1   &     26.3   &   40.9    &   50.4    &    42.1  & CVPR’22 \\
    Region CLIP (Res50) ~\cite{zhong2022regionclip}  &       &   78.3    &   52.5    &   48.8   &   57.7    &   39.3    &    36.9   &   18.6     &   27.8   &     34.8    &    29.0    & CVPR’22 \\
    GroundingDino (Swin-T) ~\cite{liu2023grounding} &  &  92.6     &  83.3    &75.7&   74.8    &   62.8    &   57.3   &33.5&44.7&52.7&58.1& ECCV’24 \\
    YOLO-World (Mid) ~\cite{cheng2024yolo} &       & 74.1      &58.9       &53.6       &60.9       &  49.2   &44.8       & 11.8      &21.3       &41.3       &27.9       & CVPR’24 \\
    YOLO-World (Large) ~\cite{cheng2024yolo} &       &       77.6&       63.0&       57.6&       65.0&       52.9&       48.5&       14.2&       26.4&       45.7&       31.9& CVPR’24 \\
    OVDINO (B) ~\cite{wang2024ov} &       &93.3       &83.7       &75.9    &74.7       &63.2       &   57.3    &39.5       & 45.6       &51.6       &46.9       & Arxiv'24 \\
    \hline
    PB-OVD ~\cite{gao2022open} &  \multirow{7}[2]{*}{Text Prompt}   & 44.4 & 24.1 & 24.8 &   27.6    &   16.4    &    16.1          &   1.2    &   2.5    &   4.5    &    3.1   & ECCV’22 \\
    GLIP-T (A) ~\cite{li2022grounded} &       &    82.8   &    69.5   &    62.9   &   62.0    &    49.1   &    44.8   &    6.6   &   13.0   &    29.1   &   18.2    & CVPR’22 \\
    GLIP-T (B) ~\cite{li2022grounded} &       &    82.2   &    69.9   &    63.5   &    63.4   &    50.5   &    46.3   &    4.9   &   10.6    &    25.0   &    15.3   & CVPR’22 \\
    Region CLIP (Res50) ~\cite{zhong2022regionclip}  &       &   16.1    &   5.1    &   1.4    &   5.8   &   0.4    &    1.7   &   0.4    &    0.1   &   0.0    &   0.1    & CVPR’22 \\
    GroundingDino (Swin-T) ~\cite{liu2023grounding} &   &   86.6    &    76.8  &69.7&   68.6    &   56.9    &   51.8    &10.0&15.3&29.9&20.1& ECCV’24 \\
    YOLO-World (Mid) ~\cite{cheng2024yolo} &       & 82.0      &71.9       &65.1       & 59.0      & 46.8      & 43.2      &       12.4&       16.6&       27.5&       20.2& CVPR’24 \\
    YOLO-World (Large) ~\cite{cheng2024yolo} &       &       81.9&       71.2&       64.4&       63.3&       51.9&       47.3&       15.2&       19.8&       30.2&       23.0& CVPR’24 \\
     OVDINO (B) ~\cite{wang2024ov}&       &88.0       &77.5       &70.4       & 68.5      & 56.8      &51.7       &  26.6     &  37.9     &41.7       &  37.4     & Arxiv'24 \\
    \bottomrule
    \end{tabular}%
}
\end{table*}

%% file: tables/det_ovd.tex
\begin{table*}[htbp]

\centering
\caption{General Open Vocabulary performance of Large-Scale Pretraining Based Methods on COCO~\cite{lin2014microsoft}, and LVIS~\cite{gupta2019lvis}.} 
\label{tab:det_ovd}
\resizebox{1.0\textwidth}{!}{%
    \begin{tabular}{l|l|ccc|cccc|cccc|l}
    \toprule
    \multirow{2}{*}{Method} & \multirow{2}{*}{Fine-tuning Ways} & \multicolumn{3}{c|}{OV-COCO} & \multicolumn{4}{c|}{OV-LVIS \textit{minival}}  & \multicolumn{4}{c|}{OV-LVIS}   & \multirow{2}{*}{Published} \\
\cline{3-13}          &  & AP$_{novel}$ & AP$_{base}$ & AP    & AP$_{r}$ & AP$_{c}$ & AP$_{f}$ & AP  & AP$_{r}$ & AP$_{c}$ & AP$_{f}$ & AP   &  \\
    \midrule
    \midrule
    GroundingDino (swin-T) ~\cite{liu2023grounding} & \multirow{12}[2]{*}{Zero Prediction} &57.1&47.5&49.8&  18.1  &  23.3 & 32.7  &    27.4   &10.4&15.3&29.9&20.2&ECCV’24 \\
    GroundingDino (swin-B) ~\cite{liu2023grounding} &  &62.7&55.4&57.3&27.6&33.1&37.2&34.6&19.0&24.1&32.9&26.7& ECCV’24 \\
    GLIP-T (A) ~\cite{li2022grounded} &       &   65.9    &    58.2   &    60.3   &   14.2 &13.9 &23.4 &18.5   & 6.0 &8.0 &19.4 &12.3&CVPR'22 \\
    GLIP-T (B) ~\cite{li2022grounded} &       &   69.8  &    59.3  &  62.0  &  13.5 &12.8 &22.2 &17.8   & 4.2 &7.6 &18.6 &11.3&CVPR'22 \\
    GLIP-T (C) ~\cite{li2022grounded} &       &   69.3  &  60.5   &   62.8   &  17.7 &19.5 &31.0 &24.9   & 7.5 &11.6 &26.1 &16.5&CVPR'22 \\
    Region CLIP (Res50) ~\cite{zhong2022regionclip}  &  &  25.2    &   31.4   &    26.8       &    12.1    &     14.7   &    9.0   &   11.7    &11.4  &  10.1 & 7.7 &  9.4 &CVPR’22 \\
    Region CLIP (Res50 x 4) ~\cite{zhong2022regionclip}  &  & 27.9    &   34.6  &  29.6     &       15.5&       16.9&       11.1&       14.0&  13.8&12.1&9.4&11.3&CVPR’22 \\
    OVDINO (A) ~\cite{wang2024ov} &       &   75.0   &    60.9   &    64.6   &   15.6    &   20.4    &   29.4  &  24.3  & 9.3 & 14.5 & 27.4 & 18.7 &Arxiv'24 \\ 
    OVDINO (B) ~\cite{wang2024ov}&       &   76.2    &    62.5    &    66.1   &    31.6   &    38.7   &   41.1    &   39.2    &  26.2& 30.1 & 37.3 & 32.2 &Arxiv'24 \\
    YOLO-World (Mid) ~\cite{cheng2024yolo} &       &50.9       &   42.1    &     44.4    &    24.5  &   29.0    &   35.1    &    31.6   &19.3  &22.0&31.7&25.3&CVPR’24 \\
    YOLO-World (Large) ~\cite{cheng2024yolo} &       &53.4       &44.2       & 46.6        &   22.6    &   32.0    &    35.8   &   33.0    &18.6  &23.0&32.6&26.0&CVPR’24 \\
    
    PB-OVD ~\cite{gao2022open} &       &   25.8    &   29.1    &    28.2   &    2.8   &   3.4      &    4.9   &    4.1   & 1.2 &2.5&4.5&3.1&ECCV’22 \\
    \hline
    GroundingDino (swin-T) ~\cite{liu2023grounding} & \multirow{6}[2]{*}{Visual Fine-tuing} &56.7&56.4&56.4&   35.4  &   51.3 &  55.7  &  52.1 &17.7&44.6&54.7&43.9&ECCV’24 \\
    GroundingDino (swin-B) ~\cite{liu2023grounding} &  &61.4&58.3&59.1&34.7&59.7&62.4&58.8&24.3&53.1&58.0&50.1&ECCV’24 \\
    GLIP-T (A) ~\cite{li2022grounded} &       &  3.2  &  68.2 &  51.2  &   0.0    &   46.3  &   50.1  &   44.1  & 0.4  & 36.8 & 45.4 & 33.9 &CVPR’22 \\
    GLIP-T (B) ~\cite{li2022grounded} &       &   16.7 &  69.6  &   55.8   &  1.5    &  51.1  &  55.4  &   48.9   & 2.0 & 41.7 & 50.4 & 38.3 &CVPR’22 \\
    Region CLIP (Res50) ~\cite{zhong2022regionclip}  & & 31.4 &  57.1  &  50.4        &     21.3    &   33.1     &    36.4   &    33.7    &  16.4 & 25.7 & 31.3 & 26.3 & CVPR’22 \\
    YOLO-World (Large) ~\cite{cheng2024yolo} &       &54.0       &52.0          &52.6       & 15.7      &  20.5     &   36.1    &     27.6  &11.5  &16.4 &33.4 &22.2 &CVPR’24 \\
    OVDINO (B) ~\cite{wang2024ov} &       &    68.2   &     72.8     &    71.6   &   20.0    &   55.9    &   56.2    &    52.9   & 13.1& 45.6 & 51.7 & 42.4 &Arxiv'24 \\
    PB-OVD ~\cite{gao2022open} &       &  30.7     &   46.1    &   42.1    &   1.7    &   6.7    &   16.9    &    11.2   & 0.3 &4.7&15.0&7.8&ECCV’22 \\
    \bottomrule
    \end{tabular}%
}
\end{table*}

%% file: tables/det_ovd2.tex
\begin{table}[htbp]

\centering
\caption{General Open Vocabulary performance of Learning Strategy Based Methods on COCO~\cite{lin2014microsoft}, and LVIS~\cite{gupta2019lvis}.} 
\label{tab:det_ovd2}
\scriptsize
\setlength\tabcolsep{2pt}
    \begin{tabular}{l|ccc|cccc|c}
    \toprule
    \multirow{2}{*}{Method} &  \multicolumn{3}{c|}{OV-COCO} & \multicolumn{4}{c|}{OV-LVIS}     & \multirow{2}{*}{Published} \\
\cline{2-8}     & AP$_{novel}$ & AP$_{base}$ & AP    & AP$_{r}$ & AP$_{c}$ & AP$_{f}$ & AP    &  \\
    \midrule
    \midrule
    OADP ~\cite{wang2023object} & 30.0 & 53.3 & 47.2 & 21.9 & 28.4 & 32.0 & 28.7 & CVPR'23 \\
    BARON ~\cite{wu2023aligning} & 42.7 & 54.9& 51.7 & 23.2 & 29.3& 32.5& 29.5 &  CVPR'23 \\
    VL-PLM ~\cite{zhao2022exploiting} & 34.4& 60.2 & 53.5& - & - & - & - & ECCV'22 \\
    DST-Det ~\cite{xu2023dst}& 46.7 & - & - & 34.5 & - & - & - & Arxiv’23 \\
    PromptDet ~\cite{feng2022promptdet} & 26.6&- &50.6 & 21.4 & 23.3 & 29.3 & 25.3 & ECCV'22 \\
    DetPro\cite{du2022learning}  & - & - & 34.9 & 20.8 & 27.8 & 32.4 & 28.4& CVPR'22 \\
    Detic ~\cite{zhou2022detecting}  & 27.8& 45.0& 47.1& 17.8 & 26.3 & 31.6 &  26.8& ECCV'22 \\
    MM-OVOD ~\cite{kaul2023multi}  & -& - & -& 27.3& - & - & 33.1 & ICML'23 \\
    LAMI-DETR ~\cite{du2025lami}   &-  &- & -& 43.4 & - & - & 41.3 & ECCV'24 \\
    OV-DETR\cite{zang2022open}  & 29.4 &52.7 & 61.0& 17.4 & 25.0 & 32.5 & 26.6 &  ECCV'22 \\
    ViLD &  27.6 & 59.5& 51.3& 16.7 & 26.5 & 34.2 & 27.8 &ICLR'22 \\
    CoDet&  30.6 &52.3 &46.6 & - & - & -  & - &  NeurIPS'23 \\
    CORA \cite{wu2023cora}&  43.1 &60.9 &56.2 & 28.1 & - & - & - &CVPR'23 \\
    DK-DETR\cite{li2023distilling} & - & -&- & 22.4  & 31.9  & 40.1 & 33.5 & ICCV'23 \\
    EdaDet\cite{shi2023edadet}&  37.8 & 57.7& 52.5& - & - & - & - & ICCV'23 \\
    F-VLM\cite{kuo2022f} &  28.0 & - & 39.6& 32.8 & - & - & 34.9 &  ICLR'23 \\
    Prompt-OVD\cite{song2023prompt} &  30.6& 63.5& 54.9& 29.4 & 33.0 & 23.1 & 24.2 & Arxiv'23 \\
    RO-ViT\cite{kim2023region}& 33.0 & -&47.7 & 32.1 & - & - & 34.0 & CVPR'23 \\
    SAS-Det\cite{zhao2023improving} & 37.4 & 58.5& 53.0& 29.0 & 32.3 & 36.8 & 33.5 & CVPR'24 \\
    CLIPSELF\cite{wu2023clipself} &  44.3 & -& -& 34.9 & - & - & -& ICLR'24 \\
    LP-OVOD\cite{pham2024lp} &  40.5 & 60.5& 55.2&  19.3& 26.1 &29.4  & 26.2 & WACV'24 \\
    \bottomrule
    \end{tabular}%
\end{table}

%% file: tables/det_generalization.tex
\begin{table*}[htbp]

\centering
\caption{Open Vocabulary Generalization performance. For each setting, models are trained by visual fine-tuning on the base categories of the dataset to the left of the arrow and tested on the full test set of the dataset to the right of the arrow.} 
\label{tab:det_generalization}
    \begin{tabular}{l|ccc|ccc|ccc|ccc|l}
    \toprule
    \multirow{2}{*}{Method} &  \multicolumn{3}{c|}{COCO$\rightarrow$ VOC} & \multicolumn{3}{c|}{COCO$\rightarrow$LVIS} & \multicolumn{3}{c|}{COCO$\rightarrow$Object365} & \multicolumn{3}{c|}{LVIS$\rightarrow$COCO}   & \multirow{2}{*}{Published} \\
\cline{2-13}       & AP$_{50}$ & AP$_{75}$ & AP    & AP$_{50}$ & AP$_{75}$ & AP & AP$_{50}$ & AP$_{75}$ & AP & AP$_{50}$ & AP$_{75}$ & AP  &  \\
    \midrule
    \midrule
    GroundingDino (swin-T) ~\cite{liu2023grounding}  &81.6&71.2&64.9&29.4&25.1&23.5&33.1&27.4&25.1&65.6&54.3&49.5& ECCV’24 \\
    GLIP-T (A) ~\cite{li2022grounded} &     54.4   &    44.6   &    40.2   &  11.6   & 9.3 &  8.8  &   7.6  &  6.2  &  5.8 & 59.2  &    47.4   & 43.2 & CVPR’22 \\
    GLIP-T (B) ~\cite{li2022grounded} &    57.0   &     47.2  &   42.6  &  13.7   & 11.3 &   10.7   &   9.9 &  8.1  & 7.5    &     62.0  &    50.7   & 46.3 & CVPR’22 \\
    Region CLIP (Res50) ~\cite{zhong2022regionclip}  &     75.7   &   48.2  & 46.2 &  19.0   &   12.1   &    11.6   &   12.4   &    7.8    &    7.6  &    53.3   &   35.3    & 33.5 & CVPR’22 \\
    YOLO-World (Large) ~\cite{cheng2024yolo} &     86.7       &74.8          &68.3    & 20.8&15.8      &   14.1    &30.4       &24.7       & 22.7      &52.9 &41.3       & 38.0      &CVPR’24 \\
    OVDINO (B) ~\cite{wang2024ov} &    83.2    &     71.8     & 65.3  &   52.3  &   44.0  &   41.4   &   36.3   &  29.5  &    27.2  & 63.0  &  52.7  & 47.9  & CVPR’24 \\
    PB-OVD ~\cite{gao2022open} &   59.3   &    36.1   &   34.9    &    6.5   &  3.7  &3.8&   7.7   &   4.7    & 4.6   &    41.8   &   24.2    & 23.9  & ECCV’22 \\
    OADP ~\cite{wang2023object} &      63.5   & 40.3      & 38.3       &  -     & -   & - &   -   &     -  &     - &  -     &    -   &- & ECCV'24 \\
    BARON ~\cite{wu2023aligning} &     -   &    -   &    -  &    -   &  -  &-&    -  &    -   &   -  &    55.7  &     39.1  & 36.2 & CVPR'23 \\
    VL-PLM ~\cite{zhao2022exploiting}  &   46.3    &   69.7    &  50.3     &   3.1    &    2.3   &  2.1  &  7.5 &    5.4    &     5.0 &     -  &   -    &- & CVPR'22 \\
    Detic ~\cite{zhou2022detecting} &      62.1    &    40.2   &   38.1    &   10.9    &  7.6  & 7.1 &  6.8    &   4.6    &   4.3  &   57.7    &   38.7    &  36.2 &  CVPR'22 \\
    PromptDet ~\cite{feng2022promptdet} &     -    &    -   &    -   &     -  &  -  &-&    -  &    -   & -   &  48.7     &    32.0   &  30.3 & CVPR'22 \\
    DetPro\cite{du2022learning}  &     -  &   -    &  -     &   -    &  -  &-&     - &   -    &   -  &   53.8   &      37.4  &34.9  & CVPR'22 \\
    LAMI-DETR ~\cite{du2025lami} &    -    &  -     &   -    &    -   &  -  &-&  -    &  -     &  - &   57.6    &   46.9     &  42.8  & CVPR'24 \\
    \bottomrule
    \end{tabular}%
\end{table*}

%% file: tables/voc2clip.tex
\begin{table*}[htbp]
\centering
\caption{Domain adaptation results (mAP$_{50}$/AP$_{50}$\%) on six adaptation scenarios, including Pascal VOC$ \rightarrow $WaterColor (P$ \rightarrow $ W), Pascal VOC$ \rightarrow $Comic (P$ \rightarrow $ C), Cityscapes$ \rightarrow $ FoggyCityscapes (C$ \rightarrow $ F), Sim10k$ \rightarrow $ Cityscapes (S$ \rightarrow $ C), Kitti$ \rightarrow $ Cityscapes (K$ \rightarrow $ C), and Cityscapes$ \rightarrow $ Kitti (C$ \rightarrow $ K). VLMs are finetuned by source dataset (the left of the arrow) and tested on the target dataset (the right of the arrow).}
\label{tab:daod}
\scriptsize
\resizebox{\textwidth}{!}{%
\begin{tabular}{p{3.5cm}|>{\centering\arraybackslash}p{2cm}|>{\centering\arraybackslash}p{1.cm}|>{\centering\arraybackslash}p{1.cm}|>{\centering\arraybackslash}p{1.cm}|>{\centering\arraybackslash}p{1.cm}|>{\centering\arraybackslash}p{1.cm}|>{\centering\arraybackslash}p{1.cm}|>{\centering\arraybackslash}p{1cm}}
\toprule
\multirow{2}{*}{Method}& \multirow{2}{*}{Finetuning Ways}&P→W&P→C&C→F& S→C &  K→C &  C→K  &  \multirow{2}{*}{Published} \\ \cline{3-8}
 & & mAP$_{50}$  & mAP$_{\scriptscriptstyle 50}$ & \tiny AP$_{\scriptscriptstyle50}$    & AP$_{50}$  & AP$_{50}$ & AP$_{50}$   &  \\ 
 \hline
 \hline
UMT~\cite{deng2021unbiased} & \multirow{3}{*}{Traditional Methods} &  58.1&-&41.7 & 43.1 &- & -&  CVPR'21 \\
DSD-DA~\cite{feng2024dsd} & &  - & - & 52.3 & 37.1 & 49.3  & - &  ICML'24 \\
SIGMA++~\cite{li2023sigma++} & &  57.4&57.7&44.5 & 57.7 &49.5 & 76.9&  TPAMI'23 \\
\hline
GroundingDino (swin-T) ~\cite{liu2023grounding} &  \multirow{12}{*}{Zero Prediction}&  51.6&57.7&34.4 & 45.0 &45.0 & 80.0&  ECCV'24 \\
GroundingDino (swin-B) ~\cite{liu2023grounding} &   &64.5&63.3  &42.4  &52.0 & 52.0 &74.9 &  ECCV'24 \\
GLIP-T (A) ~\cite{li2022grounded}   &   & 38.9&30.3  & 31.8& 44.3& 44.3& 82.3 &  CVPR'22 \\
GLIP-T (B) ~\cite{li2022grounded}   &   &  40.0&34.8 &27.4 &37.8 &37.8 & 80.6 &  CVPR'22 \\
GLIP-T (C) ~\cite{li2022grounded}   &   &  42.7&35.5& 29.6 &40.6 & 40.6& 81.2 &  CVPR'22 \\
RegionClip (Res50) ~\cite{zhong2022regionclip}  &   &24.3&21.6  & 13.5 & 36.3 & 36.3 & 7.09 &  CVPR'22 \\
RegionClip (Res50 $\times$ 4) ~\cite{zhong2022regionclip}   &   &28.1&28.2  &  14.3& 36.3& 36.3&  53.7&  CVPR'22 \\
OVDINO (A) ~\cite{wang2024ov}  &   & 41.8&30.7 & 36.6 &76.4 & 76.4& 67.7 &  Arxiv'24 \\
OVDINO (B) ~\cite{wang2024ov}   &   & 41.7&30.8 & 40.7 &76.6 & 76.6& 71.5 &  Arxiv'24 \\
YOLO-World (Large) ~\cite{cheng2024yolo}  &   &48.3&34.1 & 28.9 & 40.6 & 40.6 & 79.7&   CVPR'24 \\
YOLO-World (Mid) ~\cite{cheng2024yolo}   &   & 48.1&33.5&25.8  &36.1 &36.1 &78.1   &   CVPR'24 \\
PB-OVD ~\cite{gao2022open}   &   &  31.5&20.8 &9.8 &13.3  & 13.3 & 39.9  & ECCV'22 \\
\hline
GroundingDino (swin-T) ~\cite{liu2023grounding} &  \multirow{7}{*}{Visual Fine-tuning}  & 59.0&51.8&52.2&68.9 &52.1 &81.3 &  ECCV'24 \\
GLIP-T (A) ~\cite{li2022grounded}   &  & 35.3&16.8&46.4 &68.7 & 52.7& 84.3 &  CVPR'22 \\
GLIP-T (B) ~\cite{li2022grounded}   &  & 33.7&16.1&46.6 &70.1 &54.4 & 83.5 &  CVPR'22 \\
RegionClip (Res50) ~\cite{zhong2022regionclip}  &  &41.5&31.6& 33.9 & 42.3 & 49.4 & 72.6 &  CVPR'22 \\
YOLO-World (Large) ~\cite{cheng2024yolo} & & 48.7&32.5&47.5 & 69.1 &  54.8 & 81.7 &     CVPR'24 \\
OVDINO (B) ~\cite{wang2024ov}   &   & 47.1&31.4&52.3 &78.0 & 75.1& 82.1 & Arxiv'24 \\
PB-OVD ~\cite{gao2022open}   &   & 39.1&21.4&26.2 & 40.7&  41.5 & 76.5  & ECCV'22 \\
\hline
GroundingDino (swin-T) ~\cite{liu2023grounding} &  \multirow{7}{*}{Text Prompt}  &57.2 &56.5&34.0 &79.4 &77.9  &80.7 &  ECCV'24 \\
GLIP-T (A) ~\cite{li2022grounded}   &  & 40.0&30.0&32.6 &71.8 &71.2 & 82.1 &  CVPR'22 \\
GLIP-T (B) ~\cite{li2022grounded}   &  & 39.8&34.4&27.6 &73.3 &73.2 & 80.6 &  CVPR'22 \\
RegionClip (Res50) ~\cite{zhong2022regionclip}  &  & - & -&17.4 & 7.0 & 6.7 & 8.2 &  CVPR'22 \\
YOLO-World (Large) ~\cite{cheng2024yolo} & & 44.1&31.7&33.4 & 74.1 & 74.1  & 81.1 &     CVPR'24 \\
OVDINO (B) ~\cite{wang2024ov}   &  &41.8&31.1&47.1&76.4&61.3&79.9& Arxiv'24 \\
PB-OVD ~\cite{gao2022open}   &  & 31.0&11.0&9.8 &  21.3 &  21.3 & 39.9  & ECCV'22 \\
\bottomrule
\end{tabular}
}
\end{table*}

%% file: tables/domain_generalization.tex
\begin{table*}[htbp]
\centering
\caption{Domain generalization results (AP$_{50}$\%) for different weather conditions. VLMs are finetuned by source dataset (Day Clear) and tested on the others source and target dataset (Day Clear, Night Clear, Dusk Rainy, Night Rainy, and Day Foggy). }
\label{tab:dgod}
\resizebox{1.0\textwidth}{!}{%
\begin{tabular}{l|l|ccccc|l}
\toprule
Method & Finetuning Ways & Day Clear & Night Clear & Dusk Rainy & Night Rainy & Day Foggy & Published \\ 
\midrule 
\midrule
S-DGOD~\cite{wu2022single} & \multirow{3}{*}{Traditional Methods} &56.1  & 36.6 & 28.2 & 16.6 & 33.5 & CVPR'22 \\
Diversification~\cite{danish2024improving} & &52.8  &42.5 & 38.1 & 24.1 & 37.2 & CVPR'24 \\
UFR~\cite{liu2024unbiased} & &58.6  &40.8 & 33.2 & 19.2 & 39.6 & CVPR'24 \\
\hline
GroundingDino (swin-T) ~\cite{liu2023grounding} & \multirow{12}{*}{Zero Prediction} &38.9  & 29.7 & 27.8 & 13.8 & 30.2 & ECCV'24 \\
GroundingDino (swin-B) ~\cite{liu2023grounding} & &28.5  &22.2 & 22.4 & 13.3 & 23.0 & ECCV'24 \\
GLIP-T (A) ~\cite{li2022grounded} & & 34.4 & 15.1 & 24 & 11.1 & 27.2 & CVPR'22 \\
GLIP-T (B) ~\cite{li2022grounded}&  & 31.6 & 23.5 & 23.8 & 12.4 & 23.4 & CVPR'22 \\
GLIP-T (C) ~\cite{li2022grounded}&  & 33.2 & 25.5 & 24.7 & 12.7 & 26.1 & CVPR'22 \\
RegionClip (Res50) ~\cite{zhong2022regionclip} &  & 4.0 & 2.1 & 1.8 & 0.9 & 6.1 & CVPR'22 \\
RegionClip (Res50 $\times$ 4) ~\cite{zhong2022regionclip} &  & 5.2&  2.7&  2.7&  1.3&  7.3& CVPR'22 \\
OVDINO (A) ~\cite{wang2024ov}&  &26.6&20.5&18.9&9.1&22.3& Arxiv'24 \\
OVDINO (B) ~\cite{wang2024ov} &  & 29.2&  21.3&  20.0&  9.6&  26.0& Arxiv'24 \\
YOLO-World (Large) ~\cite{cheng2024yolo}  &  & 36.3 & 30.1 & 25.8 & 13.6 & 29.2 & CVPR'24 \\
YOLO-World (Mid) ~\cite{cheng2024yolo}  &  &20.1 &16.0  &14.0  &6.6  &16.2  & CVPR'24 \\
PB-OVD ~\cite{gao2022open} &  &12.8 &6.0  & 4.7 & 1.9 & 12.6 & ECCV'22 \\
\hline
GroundingDino (swin-T) ~\cite{liu2023grounding} & \multirow{6}{*}{Visual Fine-tuning} & 70.8 & 56.6 & 52.1 & 32.1 & 48.8 & ECCV'24 \\
GLIP-T (A) ~\cite{li2022grounded} &  & 64.4 & 50.8 & 44.1 & 26.1 & 43.9 & CVPR'22 \\
GLIP-T (B) ~\cite{li2022grounded}&  & 66.4 & 52.2 & 46.8 & 27.7 & 44.2 & CVPR'22 \\
RegionClip (Res50) ~\cite{zhong2022regionclip} &  & 45.6 & 27.0 & 21.4 & 8.2 & 30.7 & CVPR'22 \\
YOLO-World (Large) ~\cite{cheng2024yolo}  &  & 67.3 & 54.2 & 45.1 & 54.2  & 47.0 & CVPR'24 \\
OVDINO (B) ~\cite{wang2024ov} &  &30.4&22.6&21.4&10.6&25.2& Arxiv'24 \\
PB-OVD ~\cite{gao2022open} &  & 49.6& 31.1 & 20.3 & 7.2 & 28.3 & ECCV'22 \\
\hline
GroundingDino (swin-T) ~\cite{liu2023grounding} & \multirow{5}{*}{Text Prompt} & 39.5 & 31.2 & 29.7 & 15.6 & 32.5 & ECCV'24 \\
GLIP-T (A) ~\cite{li2022grounded} &  & 40.9 & 30.0 & 27.8 & 14.1 & 31.1 & CVPR'22 \\
GLIP-T (B) ~\cite{li2022grounded}&  & 33.5 & 25.3 & 25.1 & 13.5 & 25.2 & CVPR'22 \\
RegionClip (Res50) ~\cite{zhong2022regionclip} &  & 3.8 & 1.4 & 1.0 & 0.5 & 1.3 & CVPR'22 \\
OVDINO (B) ~\cite{wang2024ov} &  &29.8&22.0&20.5&9.9&25.8& Arxiv'24 \\
PB-OVD ~\cite{gao2022open} &  & 12.8& 6.0 & 4.7 & 1.9 & 12.6 & ECCV'22 \\
\bottomrule
\end{tabular}
}
\end{table*}

%% file: tables/det_few_shot.tex
\begin{table*}[htbp]
\centering
\caption{Few-shot object detection results (\%) on ODinW-13 and ODwinW-35 \cite{li2022grounded}.} 
\label{tab:det_few_shot}
\resizebox{1.0\textwidth}{!}{%
    \begin{tabular}{l|l|ccccc|ccccc|l}
    \toprule
    \multirow{2}{*}{Method} & \multirow{2}{*}{Finetuning Ways} & \multicolumn{5}{c|}{ODinW 13}         & \multicolumn{5}{c|}{ODinW 35}         & \multirow{2}{*}{Published} \\
\cline{3-12}          &       & 0     & 1     & 3     & 5     & 10    & 0     & 1     & 3     & 5     & 10    &  \\
    \midrule \midrule
    PB-OVD~\cite{gao2022open} &   \multirow{6}[2]{*}{Visual Fine-tuning}    &    14.7   &   24.3       &  32.3     &   35.1    &   39.3   &  5.9 &   14.5    &   24.4    &   28.6    &   34.5    & ECCV’22 \\
    GLIP-T (A)~\cite{li2022grounded} &       &  32.5  &  31.0  &   32.9   &   35.7  &   41.1  &   13.4  &   13.7   &   16.2  &   19.2  &   25.7   &CVPR’22 \\
    GLIP-T (B)~\cite{li2022grounded} &       &   32.0  &   30.4  &   31.8  &   33.6  &   39.6  &   13.8  &   13.1  &   14.7  &  17.1    &   23.6    & CVPR’22 \\
    Region CLIP (Res50) ~\cite{zhong2022regionclip}&       &   13.0    &    6.2   &    6.2   &   5.4   &    4.9   &    5.7   &    3.4   &    3.5   &    3.6   &   4.1    & CVPR’22 \\
    YOLO-World (Mid)~\cite{cheng2024yolo} &       &33.2       &25.1       &31.2       & 30.7      &28.0       &14.1       &10.6       &25.2       &25.2       &25.4       & CVPR’24 \\
    YOLO-World (Large)~\cite{cheng2024yolo} &       &   33.3    &     29.8  &  35.0     &   37.1    &   40.5    &  14.5     &     15.1  &   18.6    &   21.5    &    26.5   & CVPR’24 \\
    GroundingDino (Swin-T)~\cite{liu2023grounding} &  &  51.4     &   51.8    &   53.7    &   55.3    &   58.5    &  22.7     &   25.8    &   28.4    &   30.6    &   37.4    & ECCV’24 \\
    OVDINO (B) ~\cite{wang2024ov} & &34.2&47.0&51.9&51.7&54.1&15.9&24.9&27.8&28.1&29.0& Arxiv'24 \\
    \hline
    PB-OVD~\cite{gao2022open} &  \multirow{6}[2]{*}{Text Prompt}     &    14.7   &   15.0    &  18.2     &  19.2     &   20.2    &    5.9   &   6.2    &  7.5     &   8.2    &   9.1    & ECCV’22 \\
    GLIP-T (A)~\cite{li2022grounded} &       &    32.5   &    30.6   &   30.6    &   30.6    &   30.6    &    13.4   &    13.1   &    13.1   &    13.1   &    13.1   & CVPR’22 \\
    GLIP-T (B)~\cite{li2022grounded} &       &   32.0   &    30.1   &   30.1    &    30.1   &    30.1   &    13.8   &    12.7   &   12.7    &    12.7   &   12.7  & CVPR’22 \\
    Region CLIP (Res50) ~\cite{zhong2022regionclip}&       &    13.0   &    6.5   &    6.5   &   6.5   &    6.5   &   5.7    &    3.5   &    3.5   &    3.5   &   3.5    & CVPR’22 \\
    YOLO-World (Mid)~\cite{cheng2024yolo} &       &33.2      &33.3       &33.0       &32.5       &30.8       &14.1       &14.2       &14.5       &14.5       &14       & CVPR’24 \\
    YOLO-World (Large)~\cite{cheng2024yolo} &       & 33.3  & 32.1    &   32.1    &   32.1    &    32.1   &   14.5    &   14.0    &     14.1  &   14.1    &  14.1      & CVPR’24 \\
    GroundingDino (Swin-T)~\cite{liu2023grounding} &  &  51.4     &   50.9    &   50.9    &  50.9     &  50.9     &    22.7   &   22.7    &   22.7    &  22.7     &  22.7     & ECCV’24 \\
    OVDINO (B) ~\cite{wang2024ov} & &34.2&38.8&40.0&40.0&39.9&15.9&18.4&18.8&18.8&18.8& Arxiv'24 \\
    \bottomrule
    \end{tabular}%
}
\end{table*}

%% file: tables/det_robust.tex
\begin{table*}[htbp]

\centering
\caption{Robustness and Noise Resistance comparision~\cite{michaelis2019benchmarking} on VOC-C~\cite{Pascal_VOC}, COCO-C~\cite{lin2014microsoft}, and Cityscapes-C~\cite{cordts2016cityscapes}.} 
\label{tab:det_robust}
\scriptsize
\resizebox{1.0\textwidth}{!}{
    \begin{tabular}{l|ccc|ccc|ccc|l}
    \toprule
    

    \multirow{2}{*}{Method} & \multicolumn{3}{c|}{VOC-C} & \multicolumn{3}{c|}{COCO-C} & \multicolumn{3}{c|}{Cityscapes-C} & \multirow{2}{*}{Published} \\ \cline{2-10} 
        & P$_{clean}$ & mPC & rPC & P$_{clean}$ & mPC & rPC & P$_{clean}$ & mPC & rPC \\
    \midrule \midrule
    PB-OVD ~\cite{gao2022open}  &  24.5   &   12.2  &  49.7     & 16.2    &  7.9   & 49.0    &  7.6  & 4.7    & 62.6   & ECCV’22\\
    GLIP-T (A) ~\cite{li2022grounded} &  57.6   &   36.1  &   62.6  &  43.0   &  25.3   &  58.7   & 28.4    &   18.9  &  66.7   & CVPR’22 \\
    GLIP-T (B) ~\cite{li2022grounded} & 62.2   &  40.2   &  64.6   &   44.9  &   27.8  &  62.0   &   25.2  &  17.6   &   70.0  & CVPR’22 \\
    RegionClip (Res50) ~\cite{zhong2022regionclip} &   17.9  &   7.8  & 43.5    &    13.4 &  5.6   & 42.2    & 3.2    &  2.3   & 72.4    & CVPR’22 \\
    YOLO-World (Mid) ~\cite{cheng2024yolo}  &65.1     &46.1     &70.9     &42.2     &27.4     &63.3     &22.3     &16.7     &75.0     & CVPR’24 \\
    YOLO-World (Large) ~\cite{cheng2024yolo} &    65.6 &  48.6   &  74.1   &    45.7 &  28.5   &  62.3   &  25.9   &  18.8   & 72.6    & CVPR’24 \\
    GroundingDino (Swin-T) ~\cite{liu2023grounding}  &   61.5  &   44.4  &  72.3   &   48.5  &  32.3   &  66.7   &    30.6 &  21.1   &   68.9  & ECCV’24\\
    OVDINO (A) ~\cite{wang2024ov} &55.8     &34.7     &62.1     &52.4     &34.6     &66.0     &32.5     &22.5     &69.4     & Arxiv'24 \\
    OVDINO (B) ~\cite{wang2024ov}  &56.6     & 35.5    &62.7     & 53.5    & 35.8    &67.0     & 34.9    &24.8     &70.9     & Arxiv'24 \\
    \bottomrule
    \end{tabular}%
}
\end{table*}

%% file: tables/det_fine_grained.tex
\begin{table*}[htbp]

\centering
\caption{Fine-grained performance on Stanford Dogs~\cite{khosla2011novel}, and CUB-200-2011~\cite{WahCUB_200_2011}.} 
\label{tab:det_fine_grained}
\scriptsize
    \begin{tabular}{p{3.5cm}|>{\centering\arraybackslash}p{2cm}|>{\centering\arraybackslash}p{1.2cm} >{\centering\arraybackslash}p{1.2cm} >{\centering\arraybackslash}p{1.2cm}| >{\centering\arraybackslash}p{1.2cm}>{\centering\arraybackslash}p{1.2cm} >{\centering\arraybackslash}p{1.2cm}|>{\centering\arraybackslash}p{1.cm}}
    \toprule
    \multirow{2}{*}{Method} & \multirow{2}{*}{Fine-tuning Ways} & \multicolumn{3}{c|}{Stanford Dogs} & \multicolumn{3}{c|}{CUB-200-2011}     & \multirow{2}{*}{Published} \\
\cline{3-8}          &  & AP$_{50}$ & AP$_{75}$ & AP    & AP$_{50}$ & AP$_{75}$ & AP    &  \\
   \midrule
   \midrule
    GroundingDino (swin-T) ~\cite{liu2023grounding} & \multirow{12}[2]{*}{Zero Prediction} &0.2&0.2&0.2&0.1&0.1&0.1& ECCV’24 \\
    GroundingDino (swin-B) ~\cite{liu2023grounding} &  &1.5&1.1&1.1&0.9&0.9&0.9& ECCV’24 \\
    GLIP-T (A) ~\cite{li2022grounded} &       &    0.0   &   0.0    &    0.0   &    0.0   &     0.0     &       0.0       & CVPR’22 \\
    GLIP-T (B) ~\cite{li2022grounded} &       &   0.0    &   0.0    &   0.0    &      0.0    &    0.0   &     0.0        & CVPR’22 \\
    GLIP-T (C) ~\cite{li2022grounded} &  &    0.2    &   0.2    &   0.2    &    0.1   &     0.0     &    0.0   &CVPR’22 \\
    Region CLIP (Res50) ~\cite{zhong2022regionclip}  &       &   6.9   &    0.7    &    2.0   &       5.9        &   0.5    &   1.8,    & CVPR’22 \\
    Region CLIP (Res50 x 4) ~\cite{zhong2022regionclip}  &       &       13.2&          0.7&              3.5&       13.0&       0.9&       3.8& CVPR’22 \\
    OVDINO (A) ~\cite{wang2024ov} &       &          0.1&       0.1&       0.1&       0.2&       0.2&       0.2&Arxiv'24 \\ 
    OVDINO (B) ~\cite{wang2024ov}&        &          1.4&       1.3&       1.3&       1.3&       1.3&       1.2& Arxiv'24 \\
    YOLO-World (Mid) ~\cite{cheng2024yolo} &       &2.6       &2.5       &2.3         &0.3             &0.3       &0.3       & CVPR’24 \\
    YOLO-World (Large) ~\cite{cheng2024yolo} & &   2.1    &   2.1    &     2.0   &   1.9      &     1.6         &  1.7     &        CVPR’24 \\
    PB-OVD ~\cite{gao2022open} &       &   0.7    &    0.5   &   0.5    &    1.2   &     0.9    &    0.8     & ECCV’22 \\
    \hline
    GroundingDino (swin-T) ~\cite{liu2023grounding} & \multirow{6}[2]{*}{Visual Fine-tuing} &66.4&65.0&61.9&52.3&50.7&47.6& ECCV’24 \\
    GLIP-T (A) ~\cite{li2022grounded} &       &    54.2   &    52.4   &    49.6   &    18.8   &      18.1    &     16.7   & CVPR’22 \\
    GLIP-T (B) ~\cite{li2022grounded} &    &   48.1  &   46.9    &  44.7    &    17.8   &     17.2     &   16.1    & CVPR’22 \\
    Region CLIP (Res50) ~\cite{zhong2022regionclip}  &       &    27.7    &    4.4   &  10.1     &      59.1      &    26.9      &    30.6   & CVPR’22 \\
    YOLO-World (Large) ~\cite{cheng2024yolo} &       &    76.7   &    75.0       &         71.4   &   68.0           &    67.1    &62.4& CVPR’24 \\
    OVDINO (B) ~\cite{wang2024ov} &       &          53.5&       52.1&       48.7&       53.5&       52.1&       48.7&Arxiv'24 \\ 
    PB-OVD ~\cite{gao2022open} &       &    64.4   &   52.8    &    45.7   &    65.3   &    57.1   &    47.8      & ECCV’22 \\
    \hline
    GroundingDino (swin-T) ~\cite{liu2023grounding} & \multirow{6}[2]{*}{Text Prompt} &0.2&   0.2    &       0.2       &   0.1    &  0.1     &   0.1    & ECCV’24 \\
    GLIP-T (A) ~\cite{li2022grounded} &       &    0.1   &    0.1   &    0.1   &   0.0    &     0.0     &      0.0        & CVPR’22 \\
    GLIP-T (B) ~\cite{li2022grounded} &       &    0.0   &    0.0   &    0.0   &      0.0    &    0.0   &       0.0      & CVPR’22 \\
    Region CLIP (Res50) ~\cite{zhong2022regionclip}  &       &    0.7    &   0.1    &    0.2   & 0.7           &   0.1   &  0.2     & CVPR’22 \\
    YOLO-World (Large) ~\cite{cheng2024yolo} &       &    2.0   & 1.9         & 1.8      &   1.9    & 1.8             & 1.6      & CVPR’24 \\
    OVDINO (B) ~\cite{wang2024ov} &       &          11.7&       11.5&       10.9&       16.5&       15.8&       14.6&Arxiv'24 \\ 
    PB-OVD ~\cite{gao2022open} &       &   8.8    &   4.3    &   4.6    &   9.9    &   5.2    &   5.4    & ECCV’22 \\
    \bottomrule
    \end{tabular}%
\end{table*}

%% file: tables/fg_ovd.tex
\begin{table*}[htbp]
    \centering
    \caption{mAP evaluation results on the FG-OVD \cite{fgovd} dataset, including results for the difficulty-based benchmark (N=5) and the attribute-based benchmark (N=2).}
    \label{tab:fgovd}
    \scriptsize
    \resizebox{1.0\textwidth}{!}{%
    \begin{tabular}{l|cccccccc|l}
        \toprule
        Method & Hard & Medium & Easy & Trivial & Color & Material & Pattern & Transp & Published \\
        \midrule \midrule
        Detic \cite{zhou2022detecting} & 11.5 & 18.6 & 18.8 & 69.7 & 21.6 & 38.8 & 30.3 & 24.8 & ECCV'22 \\
        ViLD \cite{gu2021open} & 22.1 & 36.1 & 40.0 & 56.6 & 43.1 & 34.8 & 24.9 & 30.6 & ICLR'22 \\
        OWL-ViT (B/16) \cite{minderer2022simple} & 26.4 & 40.4 & 38.9 & 55.4 & 45.5 & 37.4 & 26.8 & 34.1 & ECCV'22 \\
        OWL-ViT (L/14) \cite{minderer2022simple} & 26.6 & 39.8 & 44.5 & 67.0 & 44.0 & 45.0 & 36.2 & 29.2 & ECCV'22 \\
        CORA \cite{wu2023cora} & 14.7 & 22.1 & 24.3 & 35.2 & 24.7 & 18.7 & 20.1 & 27.0 & CVPR'23 \\
        OWLv2 (B/16) \cite{minderer2024scaling} & 25.4 & 39.0 & 40.5 & 54.4 & 45.2 & 33.6 & 19.3 & 28.5 & NeurIPS'23 \\
        OWLv2 (L/14) \cite{minderer2024scaling} & 25.6 & 41.8 & 43.3 & 65.0 & 53.4 & 37.0 & 23.4 & 12.2 & NeurIPS'23 \\
        GroundingDino \cite{liu2023grounding} & 17.2 & 28.3 & 30.9 & 62.9 & 41.6 & 30.4 & 31.3 & 26.9 & ECCV'24 \\
        \bottomrule
    \end{tabular}
    }
\end{table*}

%% file: tables/dense_det.tex
\begin{table}[!htbp]

\centering
\caption{Comparison results (\%) on CrowdHuman, WiderPerson, and OCHuman.} 
\label{tab:dense_det}
\fontsize{5.6}{7}\selectfont
\begin{tabular}{p{2.4cm}|p{1.4cm}|>{\centering\arraybackslash}p{1.1cm}|>{\centering\arraybackslash}p{0.9cm}|>{\centering\arraybackslash}p{0.6cm}}
\toprule
\multirow{2}{*}{Method}& \multirow{2}{*}{Finetuning Ways}& {CrowdHuman} &  {WiderPerson} &  {OCHuman} \\ \cline{3-5}
 & & AP  & AP    & AP   \\ \midrule \midrule 
GroundingDino (swin-T) ~\cite{liu2023grounding} &  \multirow{9}{*}{Zero Prediction}  & 26.6 & 29.6 &34.5  \\
GLIP-T (C) ~\cite{li2022grounded}   &  & 19.7 & 22.5 & 37.1  \\
RegionClip (Res50) ~\cite{zhong2022regionclip}  &   & 5.3 & 5.5 & 1.5  \\
RegionClip (Res50 $\times$ 4) ~\cite{zhong2022regionclip}   &    & 12.3& 21.3&  6.7 \\
OVDINO (A) ~\cite{wang2024ov}  &    & 2.2& 2.6&  42.2 \\
OVDINO (B) ~\cite{wang2024ov}   &  &  17.3& 16.9&   42.0 \\
YOLO-World (Large) ~\cite{cheng2024yolo}  & & 17.7 & 12.7 & 43.6   \\
YOLO-World (Mid) ~\cite{cheng2024yolo}   & &16.8  &15.0  &41.9    \\
PB-OVD ~\cite{gao2022open}   &   &1.92   &0.7   & 14.9   \\
\hline
GroundingDino (swin-T) ~\cite{liu2023grounding} &  \multirow{6}{*}{Visual Fine-tuning}  & 53.6 & 69.8 & -  \\
GLIP-T (C) ~\cite{li2022grounded}   &  & 50.3 & 68.0 & -  \\
RegionClip (Res50) ~\cite{zhong2022regionclip}  &    & 66.4 & 85.6 & - \\
YOLO-World (Large) ~\cite{cheng2024yolo}  &  & 45.9 & 63.2 & -   \\
OVDINO (B) ~\cite{wang2024ov}  &    & 53.0& 69.4& -  \\
PB-OVD ~\cite{gao2022open}   &   & 36.9  & 57.2  &  -  \\
\hline
GroundingDino (swin-T) ~\cite{liu2023grounding} &  \multirow{6}{*}{Text Prompt}  &31.9&31.8& -  \\
GLIP-T (C) ~\cite{li2022grounded}   &  & 20.6 & 24.2 & -  \\
RegionClip (Res50) ~\cite{zhong2022regionclip}  &    & 21.3 & 30.7 &-  \\
YOLO-World (Large) ~\cite{cheng2024yolo}  &  & 17.9 &13.2 & -    \\
OVDINO (B) ~\cite{wang2024ov}  &    & 30.1& 29.6&  - \\
PB-OVD ~\cite{gao2022open}   &   &  1.92 &  0.7 &  -  \\
\bottomrule
\end{tabular}
\end{table}

%% file: tables/mess_full.tex
\begin{table*}[h]
    \begin{center}
    \caption{Zero-shot semantic segmentation results of quantitative evaluation on MESS. mIoU (\%) metric is used in every experiment.}
    \label{tab:mess}
    \resizebox{\textwidth}{!}{
    \begin{tabular}{l|ccccccc|cccccc|ccccc|ccccc|cccc|c}
    \toprule
 & \multicolumn{7}{c|}{General} & \multicolumn{6}{c|}{Earth Monitoring} & \multicolumn{5}{c|}{Medical Sciences} & \multicolumn{5}{c|}{Engineering} & \multicolumn{4}{c|}{Agri. and Biology} & \\
 & \rotatebox[origin=l]{90}{BDD100K} & \rotatebox[origin=l]{90}{Dark Zurich} & \rotatebox[origin=l]{90}{MHP v1} & \rotatebox[origin=l]{90}{FoodSeg103} & \rotatebox[origin=l]{90}{ATLANTIS} & \rotatebox[origin=l]{90}{DRAM} &  \rotatebox[origin=l]{90}{Mean} & \rotatebox[origin=l]{90}{iSAID} & \rotatebox[origin=l]{90}{ISPRS Pots.} & \rotatebox[origin=l]{90}{WorldFloods} & \rotatebox[origin=l]{90}{FloodNet} & \rotatebox[origin=l]{90}{UAVid} &  \rotatebox[origin=l]{90}{Mean} & \rotatebox[origin=l]{90}{Kvasir-Inst.} & \rotatebox[origin=l]{90}{CHASE DB1} & \rotatebox[origin=l]{90}{CryoNuSeg} & \rotatebox[origin=l]{90}{PAXRay-4} & \rotatebox[origin=l]{90}{Mean} & \rotatebox[origin=l]{90}{Corrosion CS} & \rotatebox[origin=l]{90}{DeepCrack} & \rotatebox[origin=l]{90}{PST900} & \rotatebox[origin=l]{90}{ZeroWaste-f} & \rotatebox[origin=l]{90}{Mean} & \rotatebox[origin=l]{90}{SUIM} & \rotatebox[origin=l]{90}{CUB-200} & \rotatebox[origin=l]{90}{CWFID} & \rotatebox[origin=l]{90}{Mean} & \rotatebox[origin=l]{90}{Mean} \\
\midrule\midrule
\textit{Best sup. } & \textit{44.8}\phantom{0} & \textit{63.9}\phantom{0} & \textit{50.0}\phantom{0} & \textit{45.1}\phantom{0} & \textit{42.2} & \textit{45.7} & \textit{48.6} & \textit{65.3}\phantom{0} & \textit{87.6} & \textit{92.7} & \textit{82.2} & \textit{67.8}\phantom{0} & \textit{79.1} & \textit{93.7}\phantom{0} & \textit{97.1} & \textit{73.5} & \textit{93.8} & \textit{89.5} & \textit{49.9} & \textit{85.9}\phantom{0} & \textit{82.3}\phantom{0} & \textit{52.5}\phantom{0} & \textit{67.7} & \textit{74.0}\phantom{0} & \textit{84.6}\phantom{0} & \textit{87.2} & \textit{81.9} & \textit{80.0} \\
\midrule

Cat-Seg~\cite{cho2024cat} & 46.7  & 28.9  & 23.7  & 26.7  & 40.3  & 65.8  & 38.7  & 19.3  & 45.4  & 35.7  & 37.6  & 41.6  & 35.9  & 48.2  & 17.0  & 15.7  & 31.5  & 28.1  & 12.3  & 31.7  & 19.9  & 17.5  & 20.4  & 44.7  & 10.2  & 42.8  & 32.6  & 32.0 \\
SAN~\cite{xu2023side} & 37.4  & 24.4  & 8.9  & 19.3  & 36.5  & 46.7  & 28.9  & 4.8  & 37.6  & 31.8  & 37.4  & 41.7  & 30.7  & 69.9  & 17.9  & 12.0  & 19.7  & 29.9  & 3.1  & 50.3  & 19.7  & 21.3  & 23.6  & 22.6  & 16.9  & 5.7  & 15.1  & 26.7 \\
SimpleBase~\cite{xu2022simple}  & 32.4  & 16.9  & 7.0  & 8.1  & 22.1  & 33.1  & 19.9  & 3.8  & 11.6  & 23.2  & 21.0  & 30.3  & 18.0  & 46.9  & 37.0  & 38.7  & 44.7  & 41.8  & 3.1  & 35.4  & 18.8  & 8.8  & 16.5  & 30.2  & 4.4  & 32.5  & 22.4  & 22.7\\
ZegFormer~\cite{ding2022decoupling}  &  14.1  & 4.5  & 4.3  & 10.0  & 19.0  & 29.5  & 13.6  & 2.7  & 14.0  & 25.9  & 22.7  & 20.8  & 17.2  & 27.4  & 12.5  & 11.9  & 18.1  & 17.5  & 4.8  & 29.8  & 19.6  & 17.5  & 17.9  & 28.3  & 16.8  & 32.3  & 25.8  & 17.6 \\
ZegCLIP~\cite{zhou2023zegclip} &31.0  & 16.1  & 6.8  & 3.9  & 20.7  & 59.1  & 22.9  & 5.0  & 25.7  & 32.9  & 14.9  & 21.4  & 20.0  & 53.8  & 46.7  & 27.2  & 37.3  & 41.3  & 4.6  & 40.0  & 20.7  & 16.0  & 20.3  & 26.7  & 1.0  & 38.1  & 21.9  & 25.0\\
CascadeCLIP~\cite{li2024cascade} &33.8  & 19.4  & 8.6  & 5.0  & 23.7  & 54.7  & 24.2  & 5.5  & 25.5  & 29.2 & 21.7  & 26.2  & 21.6 & 56.8  & 26.9  & 15.1 & 43.5 & 35.6 & 7.6 & 31.0 & 20.3 & 21.7 & 20.2  & 31.2  & 1.5  & 32.9  & 21.9 & 24.6\\
MaskCLIP~\cite{ding2022open} &32.2  & 20.7  & 9.0  & 14.1  & 28.7  & 31.0  & 22.6  & 14.6  & 24.2  & 22.8  & 13.2  & 24.9  & 19.9  & 45.4  & 46.7  & 38.1  & 35.8  & 41.5  & 26.2  & 47.8  & 19.8  & 16.8  & 27.7  & 26.8  & - & 17.8  & 22.3  & 26.5\\
DeOP~\cite{han2023open} & 34.4  & 18.2  & 2.5  & 12.4  & 25.8  & 43.4  & 22.8  & 11.5  & 29.0  & 20.5  & 21.7  & 31.8  & 22.9  & 45.7  & 46.5  & 38.0  & 35.0  & 41.3  & 9.8  & 52.9  & 20.7  & 4.2  & 21.9  & 33.6  & 3.6  & 45.0  & 27.4  & 26.6 \\
FC-CLIP~\cite{yu2024convolutions}& 44.5  & 22.4  & 5.1  & 7.0  & 27.3  & 19.0  & 20.9  & 3.7  & 33.9  & 36.7  & 24.2  & 38.2  & 27.3  & 56.7  & 4.0  & 11.9  & 14.2  & 21.7  & 5.3  & 13.7  & 8.1  & 19.4  & 11.6  & 25.5  & 12.5  & 1.9  & 13.3  & 19.8\\
MAFT~\cite{jiao2023learning} &45.4  & 28.4  & 12.1  & 15.0  & 37.3  & 50.8  & 31.5  & 5.5  & 40.9  & 36.2  & 31.6  & 38.9  & 30.6  & 63.6  & 20.6  & 13.4  & 36.2  & 33.5  & 7.1  & 42.1  & 15.3  & 20.9  & 21.4  & 30.8  & 16.5  & 18.5  & 21.9  & 28.5\\
MAFT+~\cite{jiao2025collaborative} &  45.7  & 25.9  & 6.8  & 20.3 & 37.5  & 44.8  & 30.2  & 9.6  & 42.5  & 37.6  & 37.8  & 39.8  &  33.5 & 72.1  & 14.0  & 11.6 & 40.0 & 34.4 & 9.3 & 19.6 & 25.3  &  19.0 & 18.3 &  52.3 & 27.9  & 33.2 & 37.8 & 30.6\\
SED~\cite{xie2024sed} & 41.6  & 28.7  & 21.8  & 24.3  & 40.0  & 54.6  & 35.2  & 12.4  & 47.0  & 35.0  & 29.2  & 43.3  & 33.4  & 60.8  & 29.0  & 13.3  & 38.5  & 35.4  & 2.5  & 36.5  & 22.5  & 26.4  & 22.0  & 37.9  & 17.1  & 50.0  & 35.0  & 32.4 \\
SCAN~\cite{liu2024open} &49.2  & 23.6  & 8.0  & 17.4  & 34.0  & 59.6  & 32.0  & 9.4  & 45.6  & 43.8  & 30.2  & 44.7  & 34.7  & 63.3  & 18.9  & 18.2  & 28.6  & 32.3  & 11.9  & 23.6  & 21.8  & 15.8  & 18.3  & 50.0  & 25.7  & 35.4  & 37.0  & 30.9 \\
EBSeg~\cite{shan2024open}&42.9  & 27.4  & 8.4  & 19.5  & 36.7  & 49.5  & 30.7  & 5.5  & 47.3  & 43.0  & 33.3  & 42.1  & 34.2  & 46.4  & 21.1  & 12.0  & 28.7  & 27.1  & 7.0  & 43.9  & 19.9  & 14.6  & 21.4  & 29.3  & 10.3  & 28.8  & 22.8  & 28.1\\
SegCLIP~\cite{luo2023segclip} & 12.8  & 5.4  & 7.6  & 4.3  & 17.1  & 38.3  & 14.3  & 5.4  & 20.2  & 27.9  & 10.0  & 16.4  & 16.0  & 37.2  & 25.4  & 17.2  & 37.5  & 29.3  & 8.4  & 37.8  & 16.4  & 10.4  & 18.3  & 22.0  & 1.4  & 25.7  & 16.4  & 18.4 \\
TCL~\cite{cha2023learning} &18.4  & 14.4  & 11.1  & 17.6  & 21.5  & 29.9  & 18.8  & 3.5  & 29.7  & 37.3  & 13.8  & 24.5  & 21.8  & 41.6  & 23.5  & 20.8  & 38.5  & 31.1  & 5.4  & 66.8  & 21.3  & 13.2  & 26.7  & 26.4  & 6.2  & 6.5  & 13.0  & 22.4 \\
MaskCLIP~\cite{zhou2022extract} &16.9  & 14.0  & 10.0  & 5.6  & 17.7  & 23.4  & 14.6  & 2.7  & 30.9  & 25.0  & 1.6  & 37.9  & 19.6  & 43.0  & 40.5  & 40.6  & 48.9  & 43.3  & 16.5  & 45.5  & 16.5  & 19.7  & 24.6  & 21.0  & 2.6  & 39.2  & 20.9  & 23.6\\
CLIPtrase~\cite{shao2024explore}& 20.5 & 14.4 & 10.9 & 28.1 & 17.8 & 30.7 & 20.4  & 4.9 & 36.8 & 44.8 & 27.8 & 30.2 & 28.9  & 46.8 & 20.3 & 12.6 & 34.7 & 28.6  & 3.1 & 54.6 & 18.5 & 21.1 & 24.3  & 25.6 & 6.7 & 28.0 & 20.1  & 24.5\\
        \bottomrule
    \end{tabular}
    }
    \end{center}
\end{table*}

%% file: tables/ovss.tex
\begin{table*}[htbp]
\centering
\caption{Comparison results for open vocabulary semantic segmentation. mIoU (\%) metric is used in every experiment.}
\label{tab:ovss}
\scriptsize 
\resizebox{1.0\textwidth}{!}{%
\begin{tabular}{p{2.cm}|>{\centering\arraybackslash}p{2cm}|>{\centering\arraybackslash}p{3.cm}|>{\centering\arraybackslash}p{1.cm}>{\centering\arraybackslash}p{1.cm}>{\centering\arraybackslash}p{1.cm}>{\centering\arraybackslash}p{1.cm}>{\centering\arraybackslash}p{1.cm}>{\centering\arraybackslash}p{1cm}}
\toprule
Method & VLM & Training Dataset & A-150 & A-847 & PC-59 & PC-459 & VOC20 & Avg.\\ 
\midrule\midrule
\textit{Best sup. } & - & - & 63.0 &- & 71.0 &- &-&-\\
\midrule
SegCLIP~\cite{luo2023segclip} & CLIP VIT-B/16 & CC3M + COCO Captions& 8.7&3.0 &25.6 & 5.7&76.8 & 24.0\\ 
TCL~\cite{cha2023learning} &CLIP VIT-B/16 & CC3M + CC12M &17.1 & 6.2 & 33.9 & 8.9& 83.2& 29.9\\
MaskCLIP~\cite{zhou2022extract} &CLIP VIT-B/16 & - & 14.2&3.3 &29.8 &7.5 &63.5 & 23.7\\ 
CLIPtrase~\cite{shao2024explore}&CLIP VIT-B/16 & - & 17.5&5.6 &35.7 &10.1 & 80.6& 30.0\\
ZegFormer~\cite{ding2022decoupling} &CLIP VIT-B/16& COCO-Stuff-156 &16.9 &4.9 &42.8 &9.1 &86.2 & 32.0\\
SimpleBase~\cite{xu2022simple}& CLIP VIT-B/16 & COCO-stuff-156 & 20.5 & 7.0 & 47.3 & 8.4 & 87.2 & 34.1\\ 
DeOP~\cite{han2023open} & CLIP VIT-B/16 & COCO-Stuff-156 & 22.9 & 7.1 & 48.8 & 9.4 & 91.7 & 36.0\\ 
SAN~\cite{xu2023side} & CLIP VIT-B/16 & COCO-Stuff-171 & 27.6 & 10.2 & 54.1 & 16.7 & 93.9 & 40.5\\
MAFT~\cite{jiao2023learning}& CLIP VIT-B/16 & COCO-Stuff-156& 29.1& 10.1&53.5 &12.8 & 90.0& 39.1\\ 
SCAN~\cite{liu2024open}& CLIP VIT-B/16 & COCO-Stuff-171 & 30.8 & 10.8 & 58.4 & 13.2 & 97.0 & 42.0 \\
EBSeg~\cite{shan2024open}& CLIP VIT-B/16 & COCO-Stuff-171 & 30.0 & 11.1 & 56.7 & 17.3 & 94.6 & 41.9\\ 
Cat-Seg~\cite{cho2024cat}& CLIP VIT-B/16& COCO-Stuff-171&31.8 & 12.0 & 57.5 & 19.0 & 94.6 & 43.0\\ 
SED~\cite{xie2024sed} & ConvNeXt-B & COCO-Stuff-171 & 31.6 & 11.4 & 57.3 & 18.6 & 94.4 & 42.7\\
MAFT+~\cite{jiao2025collaborative} & CLIP ConvNeXt-B & COCO-Stuff-171 & 33.6 & 13.2 & 55.9 & 14.2 & 93.9& 42.2 \\
ZegCLIP~\cite{zhou2023zegclip} & CLIP VIT-B/16  & COCO-Stuff-156&19.0 &5.0 &41.2 &8.9 & 93.6& 33.5\\
CascadeCLIP~\cite{li2024cascade} & CLIP ViT-B & COCO-Stuff-156 & 20.7 & 5.7 &47.7 & 9.0 & 94.0 & 35.4\\
\midrule
MaskCLIP~\cite{ding2022open}&CLIP VIT-L/14& COCO-Stuff-171& 23.7&8.2 &45.9 &10.0 &83.6 & 34.3\\ 
FC-CLIP~\cite{yu2024convolutions} &CLIP ConvNeXt-L&COCO panoptic&34.1&14.8 &58.4 &18.2 &95.4 & 44.2\\ 
MAFT~\cite{jiao2023learning} & CLIP ConvNeXt-L & COCO-Stuff-156&34.4 & 13.1 &57.5 & 17.0 & 93.0 & 43 \\
EBSeg~\cite{shan2024open}& CLIP VIT-L/14 & COCO-Stuff-171 & 32.8 & 13.7 & 60.2 & 21.0 & 96.4 & 44.8\\ 
SCAN~\cite{liu2024open} & CLIP VIT-L/14 & COCO-Stuff-171 & 33.5 &14.0&59.3&16.7 &97.2 & 44.1\\
Cat-Seg~\cite{cho2024cat}& CLIP VIT-L/14& COCO-Stuff-171&37.9 & 16.0 & 63.3 & 23.8 & 97.0 & 47.6\\ 
SED~\cite{xie2024sed} & CLIP ConvNeXt-L & COCO-Stuff-171& 35.2 & 13.9 & 60.6 & 22.6 & 96.1 &45.7\\
MAFT+~\cite{jiao2025collaborative} & CLIP ConvNeXt-L & COCO-Stuff-171&36.1 & 15.1 &59.4 & 21.6 & 96.5 & 45.7 \\
\bottomrule
\end{tabular}
}
\end{table*}

%% file: tables/fine_grained_segmentation.tex
\begin{table}[!th]
    \centering
    \caption{Comparison of different methods on PASCAL-Part and ADE20k-Part-234 in terms of mIoU. All methods are trained on seen split and evaluated on both split.}
\setlength{\tabcolsep}{2mm}
    \label{finegrained-seg}
    \begin{tabular}{c|cc|cc|c}
        \toprule
         \multirow{2}{*}{Method} & \multicolumn{2}{c|}{Pascal-Part} & \multicolumn{2}{c}{ADE20k-Part-234}  & \multirow{2}{*}{Published} \\
        
        \cmidrule(lr){2-3}\cmidrule(lr){4-5} 
        & seen & unseen & seen & unseen & \\
        \midrule
         \midrule
         Cat-Seg \cite{cho2024cat} & 44.0 & 26.1 & 31.4 & 25.8 & CVPR'24  \\
         SimpleBase \cite{xu2022simple}  & 36.3 & 19.7 & 26.5 & 18.0 & ECCV'22 \\
         MaskCLIP \cite{ding2022open}  & 39.4 & 19.6 & 32.0 & 20.7 
 & ICML'23 \\
         FC-CLIP \cite{yu2024convolutions}  & 55.6 & 24.5 & 44.3 & 26.8 & NeurIPS'23 \\
         MAFT \cite{jiao2023learning} & 34.3 & 18.0 & 28.6 & 19.1  & NeurIPS'23 \\
         SED \cite{xie2024sed}  & 48.6 & 27.3 & 39.5 & 27.7 & CVPR'24 \\
         SCAN \cite{liu2024open}  & 49.4 & 12.6 & 42.1 & 26.9 & CVPR'24 \\
         EBSeg \cite{shan2024open}  & 46.2 & 22.0 & 38.9 & 26.6 & CVPR'24 \\
         ZegCLIP \cite{zhou2023zegclip}  & 43.2 & 24.3 & 31.4 & 22.1 & CVPR'23 \\
        MAFT+ \cite{jiao2025collaborative_maft+}  & 47.3 & 18.4 & 39.4 & 27.5 & ECCV'24 \\
         Cascade-CLIP \cite{li2024cascade} & 45.9 & 24.8 & 33.6 & 25.3  & ICML'24 \\        
        \bottomrule
    \end{tabular} 
\end{table}

%% file: tables/fewshot-semantic-segmentation-pascal.tex
\begin{table*}[!th]
    \centering
    \caption{Comparison of different methods on Pascal 5$^i$ in terms of mIoU and FB-IoU. Pascal 5$^i$ consists of four subsets in total. Two types of scenarios, 1-shot and 5-shot, are tested.}
\setlength{\tabcolsep}{2.5mm}
    \label{fewshot-seg1}
    \begin{tabular}{c|cccccc|cccccc|c}
        \toprule
         \multirow{2}{*}{Method} &  \multicolumn{6}{c}{1-shot} & \multicolumn{6}{c}{5-shot} & \multirow{2}{*}{Published} \\
        \cmidrule(lr){2-7}\cmidrule(lr){8-13}
        & $5^0$ & $5^1$ & $5^2$ & $5^3$ & \textbf{mIoU} & \textbf{FB-IoU} & $5^0$ & $5^1$ & $5^2$ & $5^3$ & \textbf{mIoU} & \textbf{FB-IoU} & \\
        \midrule
        \midrule
         \underline{SegGPT} \cite{wang2023seggpt}  & - & - & - & - & 83.2 & - & - & - & - & - & 89.8 & - & ICCV'23 \\
         SAN \cite{xu2023side}  & 72.9 & 79.4 & 66.5 & 73.2 & 74.6 & 82.4 & 73.2 & 80.0 & 66.8 & 73.7 & 75.3 & 83.1 & CVPR'23 \\
         SimpleBase \cite{xu2022simple}  & 63.5 & 69.2 & 57.8 & 60.1 & 58.2 & 67.2 & 64.3 & 70.1 & 58.2 & 60.9 & 59.0 & 67.6 & ECCV'22 \\
         MaskCLIP \cite{ding2022open} & 67.4 & 74.3 & 61.6 & 64.2 & 63.9 & 73.8 & 67.9 & 74.8 & 62.1 & 64.6 & 64.4 & 74.3 & ICML'23  \\
         DeOP \cite{han2023open}  & 16.8 & 32.7 & 14.5 & 14.4 & 23.6 & 29.3 & 17.3 & 33.6 & 15.7 & 15.4 & 24.5 & 31.1 & ICCV'23\\
         FC-CLIP \cite{yu2024convolutions} &  75.1 & 83.3 & 68.8 & 75.7 & 76.3 & 84.5 & 77.6 & 84.3 & 72.4 & 75.1 & 75.9 & 84.4  & NeurIPS'23 \\
         MAFT \cite{jiao2023learning} & 69.3 & 76.5 & 63.4 & 66.1 & 66.2 & 76.4 & 69.2 & 78.6 & 61.4 & 69.5 & 70.4 & 78.9 & 
NeurIPS'23  \\
         SED \cite{xie2024sed} & 70.3 & 76.8 & 63.8 & 66.5 & 67.1 & 76.8 & 70.1 & 76.6 & 63.8 & 66.7 & 67.5 & 77.1 & CVPR'24 \\
         SCAN \cite{liu2024open} & 73.9 & 82.4 & 67.9 & 74.6 & 75.5 & 83.2 & 74.1 & 82.5 & 68.1 & 74.8 & 75.8 & 83.6 & CVPR'24 \\
         EBSeg \cite{shan2024open} & 70.6 & 77.4 & 64.5 & 67.3 & 67.5 & 77.6 & 70.5 & 79.9 & 62.5 & 70.7 & 71.5 & 80.2 & CVPR'24 \\
         ZegCLIP \cite{zhou2023zegclip}  & 69.3 & 75.6 & 63.2 & 65.5 & 66.1 & 75.2 & 70.7 & 77.1 & 64.6 & 66.7 & 67.5 & 77.8 & CVPR'23 \\
         TCL \cite{cha2023learning} & 68.8 & 74.5 & 61.9 & 64.8 & 65.3 & 75.0 & 69.6 & 76.3 & 62.8 & 66.1 & 67.1 & 76.7 & CVPR'23 \\
        MAFT+ \cite{jiao2025collaborative_maft+}  & 72.6 & 78.8 & 66.7 & 72.8 & 74.3 & 81.7 & 73.3 & 80.2 & 67.3 & 74.0 & 75.5 & 83.4 & ECCV'24 \\
         Cascade-CLIP \cite{li2024cascade}  & 69.7 & 76.3 & 62.9 & 66.0 & 66.8 & 76.3 & 71.2 & 77.4 & 64.3 & 67.2 & 68.4 & 80.4 & ICML'24 \\
        \bottomrule
    \end{tabular} 
\end{table*}

\begin{table*}[!th]
    \centering
    \caption{Comparison of different methods on COCO-20$^i$ in terms of mIoU and FB-IoU. COCO-20$^i$ consists of four subsets in total. Two types of scenarios, 1-shot and 5-shot, are tested.}
\setlength{\tabcolsep}{2.5mm}
    \label{fewshot-seg2}
    \begin{tabular}{c|cccccc|cccccc|c}
        \toprule
         \multirow{2}{*}{Method} & \multicolumn{6}{c}{1-shot} & \multicolumn{6}{c}{5-shot} & \multirow{2}{*}{Published} \\
        \cmidrule(lr){2-7}\cmidrule(lr){8-13}
        & $20^0$ & $20^1$ & $20^2$ & $20^3$ & \textbf{mIoU} & \textbf{FB-IoU} & $20^0$ & $20^1$ & $20^2$ & $20^3$ & \textbf{mIoU} & \textbf{FB-IoU} & \\
        \midrule
        \midrule
         \underline{PGMA-Net} \cite{chen2024visual_pgmanet}  & 55.2 & 62.7 & 60.3 & 59.4 & 59.4 & 78.5 & 55.9 & 65.9 & 63.4 & 61.9 & 61.8 & 79.4 & TMM'24 \\
         SAN \cite{xu2023side}  & 42.6 & 47.8 & 38.7 & 40.6 & 44.3 & 48.1 & 44.3 & 49.7 & 40.4 & 42.5 & 46.1 & 50.4 & CVPR'23 \\
         SimpleBase \cite{xu2022simple}  & 35.7 & 38.9 & 30.8 & 35.6 & 36.5 & 38.8 & 36.4 & 39.7 & 31.6 & 36.8 & 37.2 & 39.7 & ECCV'22 \\
         MaskCLIP \cite{ding2022open} & 36.6 & 40.3 & 32.1 & 37.2  & 38.3 & 41.5 & 37.3 & 41.8 & 32.9 & 38.0 & 39.1 & 43.1 & 
ICML'23   \\
         DeOP \cite{han2023open}& 9.4 & 12.3 & 8.7 & 9.8 & 9.6 & 12.1 & 10.3 & 12.8 & 9.7 & 10.6 & 10.7 & 12.8  & ICCV'23  \\
         FC-CLIP \cite{yu2024convolutions}  & 54.4 & 62.9 & 49.4 & 57.4 & 56.7 & 68.2 & 55.4 & 63.2 & 52.8 & 58.1 & 56.9 & 66.8  & NeurIPS'23 \\
         MAFT \cite{jiao2023learning} & 39.2 & 41.7 & 34.3 & 39.2 & 40.1 & 42.2 & 40.5 & 42.9 & 35.1 & 40.6 & 41.5 & 43.6 & NeurIPS'23 \\
         SED \cite{xie2024sed} & 53.1 & 61.9 & 46.1 & 54.8 & 53.5 & 65.1 & 55.2 & 62.2 & 48.9 & 55.7 & 55.1 & 66.0  & CVPR'24  \\
         SCAN \cite{liu2024open} & 53.9 & 62.7 & 46.6 & 55.7 & 54.3 & 65.8 & 55.7 & 62.9 & 49.6 & 56.2 & 55.8 & 66.7 & CVPR'24  \\
         EBSeg \cite{shan2024open} & 51.7 & 60.2 & 45.1 & 53.9 & 52.6 & 64.1 & 53.2 & 61.9 & 46.2 & 55.2 & 54.4 & 65.5  & CVPR'24 \\
         ZegCLIP \cite{zhou2023zegclip}  & 48.6 & 56.4 & 41.9 & 49.2 & 47.8 & 61.4 & 50.2 & 57.8 & 43.2 & 50.7 & 49.1 & 63.9 & CVPR'23 \\
         TCL \cite{cha2023learning}  & 32.5 & 36.2 & 28.7 & 33.1 & 32.8 & 36.8 & 34.6 & 38.4 & 30.4 & 34.8 & 34.2 & 38.5 & CVPR'23 \\
         MAFT+ \cite{jiao2025collaborative_maft+}  & 47.5 & 52.7 & 40.6 & 47.2 & 46.8 & 58.9 & 49.7 & 56.1 & 44.0 & 49.6 & 48.7 & 61.7 & ECCV'24 \\
        \bottomrule
    \end{tabular} 
\end{table*}

\begin{table}[h]
    \centering
    \caption{Comparison of different methods on FSS-1000 in terms of mIoU. Two types of scenarios, 1-shot and 5-shot, are tested.}
\setlength{\tabcolsep}{4mm}
    \label{fewshot-3}
    \begin{tabular}{c|cc|c}
        \toprule
         \multirow{2}{*}{Method}  & \multicolumn{2}{c}{mIoU} & \multirow{2}{*}{Published}  \\
        \cmidrule(lr){2-3}
        & \textbf{1-shot} & \textbf{5-shot} &  \\
       \midrule
       \midrule
         \underline{DACM} \cite{xiong2022doubly_dacm}   & 90.8 & 91.7 & ECCV'22\\
         SAN \cite{xu2023side} & 84.1 & 86.9 & CVPR'23 \\
         SimpleBase \cite{xu2022simple}  & 68.4 & 71.2 & ECCV'22 \\
         MaskCLIP \cite{ding2022open}  & 70.5 & 71.2 & ICML'23 \\
         DeOP \cite{han2023open}  & 47.8 & 48.6 & ICCV'23 \\
         FC-CLIP \cite{yu2024convolutions}  & 85.6 & 88.2 & NeurIPS'23 \\
         MAFT \cite{jiao2023learning}  & 69.3 & 70.6 & NeurIPS'23\\
         SED \cite{xie2024sed}  & 74.8 & 77.2 & CVPR'24\\
         SCAN \cite{liu2024open} & 85.2 & 88.7  & CVPR'24\\
         EBSeg \cite{shan2024open}  & 75.4 & 77.5 & CVPR'24\\
         ZegCLIP \cite{zhou2023zegclip}  & 73.2 & 75.5 & CVPR'23\\
         TCL \cite{cha2023learning}  & 75.7 & 78.1 & CVPR'23\\
         MAFT+ \cite{jiao2025collaborative_maft+}  & 83.7 & 86.0 & ECCV'24 \\
         Cascade-CLIP \cite{li2024cascade} & 74.1 & 76.5  & ICML'24 \\
        \bottomrule
    \end{tabular} 
\end{table}

%% file: tables/robust_seg.tex
\begin{table*}[htbp]
\centering
\caption{Robustness and Noise Resistance comparision ~\cite{michaelis2019benchmarking} for Segmentation} 
\label{tab:seg_robust}
\resizebox{1.0\textwidth}{!}{
    \begin{tabular}{c|ccc|ccc|ccc|ccc|c}
    \toprule
    \multirow{2}{*}{Method} & \multicolumn{3}{c|}{ADE150} & \multicolumn{3}{c|}{ADE849} & \multicolumn{3}{c|}{PC59} & \multicolumn{3}{c|}{PC459} & \multirow{2}{*}{Published} \\
        & P$_{clean}$ & mPC & rPC & P$_{clean}$ & mPC & rPC & P$_{clean}$ & mPC & rPC & P$_{clean}$ & mPC & rPC & \\
    \midrule\midrule
        SimpleBase \cite{xu2022simple} & 20.4 & 12.3 & 60.3 & 7.4 & 4.8 & 64.9 & 47.2 & 31.3 & 66.3 & 8.4 & 4.8 & 57.1 & ECCV'22 \\
        MaskCLIP \cite{ding2022open} & 23.7 & 17.4 & 73.4 & 8.2 & 5.7 & 69.5 & 45.9 & 35.5 & 77.3 & 10.0 & 6.5 & 65.0 & ECCV'22 \\
        MAFT \cite{jiao2023learning} & 29.1 & 19.7 & 67.7 & 10.2 & 7.2 & 70.6 & 53.3 & 38.5 & 72.2 & 12.8 & 7.1 & 55.5 & CVPR'23 \\
        SAN \cite{xu2023side} & 27.6 & 21.0 & 76.1 & 10.2 & 7.5 & 73.5 & 53.8 & 42.8 & 79.6 & 16.7 & 10.5 & 62.9 & CVPR'23 \\
        DeOP \cite{han2023open} & 23.0 & 14.8 & 64.3 & 7.0 & 4.8 & 68.6 & 48.9 & 33.8 & 69.1 & 9.8 & 5.4 & 55.1 & ICCV'23 \\
        FC-CLIP \cite{yu2024convolutions} & 34.1 & 24.6 & 72.1 & 14.8 & 10.6 & 71.6 & 58.4 & 43.6 & 74.7 & 18.2 & 11.8 & 64.8 & NeurlPS'23 \\
        SCAN \cite{liu2024open} & 30.8 & 20.3 & 65.9 & 10.8 & 7.1 & 65.7 & 58.4 & 43.8 & 75.0 & 13.2 & 9.6 & 72.7 & CVPR'24 \\
        EBSeg \cite{shan2024open} & 30.0 & 22.1 & 73.7 & 11.1 & 8.3 & 74.8 & 56.7 & 43.8 & 77.2 & 17.3 & 10.8 & 62.4 & CVPR'24 \\
        Cat-Seg \cite{cho2024cat} & 31.8 & 24.1 & 75.8 & 12.0 & 9.2 & 76.7 & 57.5 & 45.0 & 78.3 & 19.0 & 11.9 & 62.6 & CVPR'24\\
        SED \cite{xie2024sed} & 31.8 & 21.6 & 67.9 & 11.2 & 8.0 & 71.4 & 57.7 & 39.6 & 68.6 & 18.6 & 9.7 & 52.2 & CVPR'24 \\
        MAFT-Plus \cite{jiao2025collaborative} & 33.6 & 24.3 & 72.3 & 13.2 & 10.0 & 75.8 & 55.9 & 40.5 & 72.5 & 14.2 & 7.7 & 54.2 & ECCV'24  \\
        Cascade-CLIP \cite{li2024cascade} & 22.1 & 16.0 & 72.1 & 6.26 & 4.8 & 76.7 & 51.7 & 38.1 & 73.7 & 9.8 & 6.1 & 62.2 & ICML'24 \\
    \bottomrule
    \end{tabular}
}
\end{table*}

%% file: tables/zeroshot_seg.tex
\begin{table*}[t]
    \centering
    \vskip -0.1in
    \caption{Comparison with the state-of-the-art zero-shot segmentation methods on COCO-Stuff 164K, and PASCAL VOC 2012 datasets.
    R denotes ResNet~\cite{he2016deep}.
    %
    }
    \label{tab:zeroshotseg}
    \begin{center}
    \begin{small}
    \setlength{\tabcolsep}{2mm}{
    \begin{tabular}{l|c|c|ccc|ccc|c}
    \toprule
    \multirow{2}{*}{Methods} & \multirow{2}{*}{Backbone} & \multirow{2}{*}{Segmentor} & \multicolumn{3}{c|}{COCO-Stuff 164K~(171)} & \multicolumn{3}{c|}{PASCAL VOC 2012~(20)} & \multirow{2}{*}{Published} \\  
    \cmidrule(lr){4-6}  \cmidrule(lr){7-9}
    & &  & mIoU$^S$$\uparrow$ & mIoU$^U$$\uparrow$ & hIoU$\uparrow$ & mIoU$^S$$\uparrow$ & mIoU$^U$$\uparrow$ & hIoU$\uparrow$ & \\
    \midrule
    \midrule
    SPNet-C~\cite{xian2019semantic} & R101& W2V\&FT& 35.2& 8.7 & 14.0& 78.0& 15.6& 26.1 & CVPR'19 \\
    ZS3Net~\cite{bucher2019zero}  & R101& W2V & 34.7& 9.5 & 15.0& 77.3& 17.7& 28.7 & NeurIPS’19\\
    CaGNet~\cite{gu2020context}   & R101& W2V\&FT& 33.5& 12.2& 18.2& 78.4& 26.6& 39.7 & ACM MM‘2020\\
    SIGN~\cite{cheng2021sign}   & R101& W2V\&FT     & 32.3& 15.5& 20.9& 75.4& 28.9& 41.7 & ICCV'2021\\
    ZegFormer~\cite{ding2022decoupling}&R101\&CLIP-B& MaskFormer &36.6& 33.2& 34.8& 86.4& 63.6& 73.3 & CVPR’22\\
    Zsseg~\cite{xu2022simple}&R101\&CLIP-B& MaskFormer &39.3& 36.3& 37.8& 83.5& 72.5& 77.5 & ECCV’22\\
    DeOP~\cite{han2023open}&R101\&CLIP-B& MaskFormer&38.0&38.4&38.2&88.2&74.6&80.8 & ICCV’23\\
    Zsseg+MAFT~\cite{jiao2023learning}&R101\&CLIP-B& MaskFormer& 40.6& 40.1&40.3&88.4&66.2&75.7 & NeurIPS’23\\
    ZegCLIP~\cite{zhou2023zegclip}&CLIP-B &SegViT &40.2& 41.4& 40.8& 91.9& 77.8& 84.3 & CVPR’23\\
    CascadeCLIP\cite{li2024cascade} & CLIP-B &SegViT&41.1&43.4&42.2&92.7&83.1&87.7 & ICML’24\\
    \midrule
    \end{tabular}}
    \end{small}
    \end{center}
    \vskip -0.1in
\end{table*}

%% file: tables/seg_dense_obj.tex
\begin{table*}[!th]
    \centering
    \caption{Dense Evaluation Results for Open Vocabulary Segmentation}
\setlength{\tabcolsep}{3mm}
\resizebox{1\linewidth}{!}{
    \label{tab:seg_dense}
    \begin{tabular}{c|ccc|ccc|ccc|c}
        \toprule
         \multirow{2}{*}{Method} & \multicolumn{3}{c}{COCO-OCC} & \multicolumn{3}{c}{CIS} & \multicolumn{3}{c|}{OCHuman} & \multirow{2}{*}{Published} \\
        
        \cmidrule(lr){2-4} \cmidrule(lr){5-7} \cmidrule(lr){8-10} 
        & AP & AP50 & AP75 & AP & AP50 & AP75 &  AP & AP50 & AP75 & \\
        \midrule\midrule
        
MaskCLIP~\cite{ding2022open} & 13.87 & 20.86 & 14.54 & 25.98 & 35.97 & 29.55 & 12.00 & 20.78 & 12.36 & ECCV'22 \\ 
FreeSeg~\cite{qin2023freeseg} & 18.32 & 27.73 & 19.91 & 46.27 & 74.98 & 49.43& 13.92 & 25.02 & 13.99 & CVPR'23 \\ 
MAFT-Plus \cite{jiao2025collaborative} & 36.15 & 57.29 & 38.83 & 52.29 & 78.09 & 58.09 & 28.39 & 46.04 & 31.39 & ECCV'24 \\
ODISE~\cite{xu2023open} & \color{black}{44.38} & 67.62 & \color{black}{48.42} & 62.47 & 86.67 & 69.14 & 30.95 & 44.25 & 34.42 & CVPR'23 \\ 
FC-CLIP~\cite{yu2024convolutions} & 44.05 & \color{black}{67.95} & 47.39 & \color{black}{63.51} & \color{black}{90.08} & \color{black}{70.65} & \color{black}{34.21} & \color{black}{50.62} & \color{black}{38.62} & NeurIPS'23 \\ 

        \bottomrule
    \end{tabular} 
}
\end{table*}

%% file: tables/seg_small_object_city_camvid.tex
\begin{table*}[htbp]
    \centering
    \caption{Small Object Semantic Segmentation Comparison results (\%) on Cityscapes, CamVid.} 
\scriptsize 
\resizebox{1.0\textwidth}{!}{
    \begin{tabular}{l|cccc|cccc|l}
   
    \toprule
    \multirow{2}{*}{Method}   & \multicolumn{4}{c|}{Cityscapes} & \multicolumn{4}{c|}{CamVid}&\multirow{2}{*}{Published} \\

        \cline{2-9}
        & mIoU$\uparrow$ & mACC$\uparrow$ & fwIoU$\uparrow$ & pACC$\uparrow$ & mIoU $\uparrow$& mACC $\uparrow$& fwIoU$\uparrow$ & pACC $\uparrow$\\
\hline\hline
\textit{Best sup.} & \textit{78.3 }\phantom{0}& \textit{ -}\phantom{0}& \textit{- }\phantom{0}& \textit{ -}\phantom{0}& \textit{81.7 }\phantom{0}& \textit{- }\phantom{0}& \textit{- }\phantom{0}& \textit{ -}\phantom{0} & \textit{-}\phantom{0} \\
\hline
Simple Baseline~\cite{xu2022simple}  &  34.45 & 49.35  &  47.76 & 62.54  & 36.75  & 50.55 & 50.30 & 63.69    & ECCV'22 \\
SAN~\cite{xu2023side}  & 38.12  &  51.75 & 73.55  & 82.53  & 51.20  &  61.35 &  78.68 &  87.45 & CVPR'23    \\
MAFT~\cite{jiao2023learning}  & 45.23 & 56.37 & 79.05 & 82.53 & 55.53 & 66.48 & 80.00 & 88.15 & NeurIPS'23  \\
SegCLIP~\cite{luo2023segclip}  &  11.00  &  22.26  &  - &  29.75  &  7.38  &  20.26  &  -  &  26.20 & ICML'24 \\
CLIPtrase~\cite{shao2024explore} & 21.06 & 36.92 & 49.15 & 63.23  & 25.49  & 36.76  & 44.41 & 60.14    & ECCV'24\\
Cascade-CLIP~\cite{li2024cascade}& 39.79 & 56.45 & - & 76.15  & 51.46  & 60.64  & - & 88.15 & CVPR'24   \\
SED~\cite{xie2024sed}  & 41.45  & 52.07  & 72.45  & 83.63  & 55.39  & 65.66  &  79.04  & 88.10 & CVPR'24 \\
Cat-Seg~\cite{cho2024cat} & 43.98 & 55.26 & 78.78 & 87.53  & 55.04  & 63.94  & 81.19 & 89.23  & CVPR'24  \\
EBSeg~\cite{shan2024open}  &  44.56 & 57.71 & 75.55 & 84.00  & 49.72 & 61.88  & 74.53 & 83.60  & CVPR'24  \\
SCAN~\cite{liu2024open} & 49.70  &  60.15 & 81.71 & 89.51 &  57.68 & 65.82 & 82.52  & 90.18   & CVPR'24 \\
FC-CLIP~\cite{yu2024convolutions} & 55.46 & 69.19 & 81.70 & 89.01 & 51.01 & 67.77 & 71.89 & 80.22 & NeurIPS'23\\
MAFT+~\cite{jiao2025collaborative}   & 53.36 & 64.24 & 82.95 & 90.31 & 56.63 & 71.15 & 82.75 & 90.26 & ECCV'24\\

    \bottomrule
    \end{tabular}}
    \label{tab:seg_small_1}
\end{table*}

%% file: tables/seg_small_object_uavid_udd6.tex
\begin{table*}[htbp]
\centering
\caption{Small Object Semantic Segmentation Comparison results (\%) on UAVid, UDD6} 
\scriptsize 
\resizebox{1.0\textwidth}{!}{%
\begin{tabular}{l|cccc|cccc|l}
\toprule
\multirow{2}{*}{Method}  &\multicolumn{4}{c|}{UAVid} & \multicolumn{4}{c|}{UDD6} & \multirow{2}{*}{Published}\\ 
\cline{2-9}
&  mIoU$\uparrow$ & mACC$\uparrow$ & fwIoU$\uparrow$ & pACC$\uparrow$ & mIoU$\uparrow$ & mACC$\uparrow$ & fwIoU $\uparrow$& pACC$\uparrow$ \\ 
\hline\hline
\textit{Best sup.}  & \textit{69.5} & \textit{-} & \textit{-} & \textit{-} & \textit{79.7} & \textit{-} & \textit{-} & \textit{-}& \textit{-}\phantom{0} \\
\hline
Simple Baseline~\cite{xu2022simple}   & 19.19  & 31.05  & 29.81  & 42.69 & 24.03 & 35.03 & 29.35 & 44.80 & ECCV'22 \\
SAN~\cite{xu2023side}  & 25.05  & 45.53 & 36.34 & 57.65 & 40.83  & 56.86  & 47.05 & 64.64  & CVPR'23  \\
MAFT~\cite{jiao2023learning} & 31.03 & 50.28 & 43.73 & 64.84 & 41.19 & 57.16 & 49.49 & 64.65 & NeurIPS'23  \\
SegCLIP~\cite{luo2023segclip} & 11.71  & 23.48   & -  & 35.93  & 17.63  & 29.84  & -  & 37.47  & ICML'24 \\
CLIPtrase~\cite{shao2024explore} & 10.29 & 17.98 & 21.46 & 30.21 & 38.81  & 25.99  & 42.56 & 61.02 & ECCV'24   \\
Cascade-CLIP~\cite{li2024cascade} & 16.65 & 26.58 & - & 41.56 & 40.75  & 59.31  & - & 61.50  & CVPR'24  \\
SED~\cite{xie2024sed} & 26.46  & 47.91  & 37.04  & 60.51  & 43.03  & 60.83  & 50.00 & 67.02  & CVPR'24   \\
Cat-Seg~\cite{cho2024cat}  & 26.77  & 47.91 & 39.21 & 63.55 & 50.61  & 68.15 & 56.97 & 71.62  & CVPR'24  \\
EBSeg~\cite{shan2024open}  & 27.23 & 48.14 & 39.66 & 61.85 & 40.97 & 57.45 & 46.91 & 64.16   & CVPR'24  \\
SCAN~\cite{liu2024open} & 29.56 & 48.14 & 39.66 & 61.85 & 39.74 & 54.68 & 44.93 & 60.78  & CVPR'24  \\
FC-CLIP~\cite{yu2024convolutions} & 27.92 & 40.17 & 49.70 & 63.45 & 62.19 & 75.71 & 71.52 & 82.84 & NeurIPS'23  \\
MAFT+~\cite{jiao2025collaborative}  & 27.93 & 50.22 & 41.42 & 65.24 & 68.07 & 71.56 & 79.21 & 83.18 & ECCV'24  \\
\bottomrule
\end{tabular}}
\label{tab:seg_small_2}
\end{table*}